\documentclass[10pt,journal, twocolumn]{IEEEtran}
\ifCLASSOPTIONcompsoc
\else
\fi

\ifCLASSINFOpdf
\else
\fi

\usepackage[table]{xcolor}
\usepackage{times}
\usepackage{epsfig}
\usepackage{graphicx}
\usepackage{amsmath}
\usepackage{amssymb}
\usepackage{epsfig}
\usepackage{graphicx}
\usepackage{amsmath}
\usepackage{amssymb}
\usepackage{bbm}

\usepackage{algpseudocode,algorithm,algorithmicx} 

\usepackage{array,setspace, amsfonts,url,bm, cellspace}
\usepackage{tikz}
\usepackage{epstopdf}
\usetikzlibrary{positioning}
\usetikzlibrary{arrows}
\usepackage{pifont}
\usepackage{multirow, hhline}

\newcolumntype{M}[1]{>{\centering\arraybackslash}m{#1}}

\usepackage[pagebackref=true,breaklinks=true,letterpaper=true,colorlinks,bookmarks=false]{hyperref}

\newcommand{\etal}{\textit{et al}.}

\begin{document}

\title{A Quadruplet Loss for Enforcing Semantically Coherent Embeddings in Multi-output Classification Problems}

\author{Hugo Proen\c{c}a,~\IEEEmembership{Senior Member,~IEEE}, Ehsan Yaghoubi and Pendar Alirezazadeh
\IEEEcompsocitemizethanks{Authors are with the IT: Instituto de Telecomunica\c{c}\~{o}es, Department of Computer Science, University of Beira Interior, Covilh\~{a}, Portugal, E-mail: hugomcp@di.ubi.pt, \{D2401, D2389\}@di.ubi.pt.  }
\thanks{Manuscript received: January, 2020.}}

% The paper headers
\markboth{IEEE TRANSACTIONS ON INFORMATION FORENSICS AND SECURITY,~Vol.~??, No.~??, ??~2018}%
{Shell \MakeLowercase{\textit{et al.}}: Bare Demo of IEEEtran.cls for Computer Society Journals}

\IEEEtitleabstractindextext{%
\begin{abstract}
This paper describes one objective function for learning semantically coherent feature embeddings in multi-output classification problems, i.e., when the response variables have dimension higher than one. In particular, we consider the problems of  identity retrieval and soft biometrics labelling in visual surveillance environments, which have been attracting growing interests. Inspired by the triplet loss~\cite{Schroff2015} function, we propose a generalization that: 1) defines a metric that considers the number of agreeing labels between pairs of elements; and 2) disregards the notion of \emph{anchor}, replacing $d(A_1,A_2) < d(A_1,B)$ by $d(A,B) < d(C,D),~\forall~A, B, C, D$ distance constraints, according to the number of agreeing labels between pairs. As the triplet loss formulation, our proposal also privileges small distances  between \emph{positive} pairs, but at the same time explicitly enforces that the distance between  other pairs corresponds directly to their similarity in terms of agreeing labels. This yields feature embeddings with a strong correspondence between the classes centroids and their semantic descriptions, i.e., where elements are closer to others that share some of their labels than to elements with fully disjoint labels membership. As practical effect, the proposed loss can be seen as particularly suitable for performing joint coarse (soft label) + fine (ID) inference, based on simple rules as \emph{k-neighbours}, which is a novelty with respect to previous related loss functions.  Also, in opposition to its triplet counterpart, the proposed loss is  agnostic with regard to any demanding criteria for mining learning instances (such as the \emph{semi-hard} pairs).  Our experiments were carried out in five different datasets (BIODI, LFW, IJB-A, Megaface and PETA) and validate our assumptions, showing highly promising results.  
\end{abstract}

}

\maketitle

\IEEEdisplaynontitleabstractindextext

\IEEEpeerreviewmaketitle

\section{Introduction}

\IEEEPARstart{C}haracterizing pedestrians in crowds has been attracting growing attention, with soft labels  as \emph{gender}, \emph{ethnicity} or \emph{age} being particularly important to determine the identities in a scene. This kind of labels is closely related to human perception and describes the visual appearance of subjects, with applications in identity retrieval~\cite{Wang2016}\cite{Shi2015}  and person re-identification~\cite{Halstead2018}\cite{Li2019} problems. 

Deep learning frameworks have been repeatedly improving the state-of-the-art in many computer vision tasks, such as object detection and classification~\cite{Liu2016}\cite{Wang2017}, action recognition~\cite{Ji2013}\cite{Choutas2018}, semantic segmentation~\cite{Lateef2019}\cite{Zhou2014} and soft biometrics labelling~\cite{Samangouei2016}.  In this context, the triplet loss~\cite{Schroff2015} is an extremely popular concept, where three learning elements are considered at a time, two of them of the same class and a third one belonging to a different class. By imposing larger distances between the elements of the \emph{negative} than of the \emph{positive} pair, the intra-class compactness and inter-class discrepancy on the destiny space are enforced. This learning paradigm was successfully applied to various problems, upon the mining of the \emph{semi-hard} negative input pairs, i.e., cases where the negative element is not closer to the anchor than the positive, but still provides a positive loss due to an imposed margin.

 \begin{figure}[ht!]
\begin{center}
\begin{tikzpicture}

\def\deltaX{0}
\draw (0+\deltaX,0) node(n1)  {\includegraphics[width=1.9 cm]{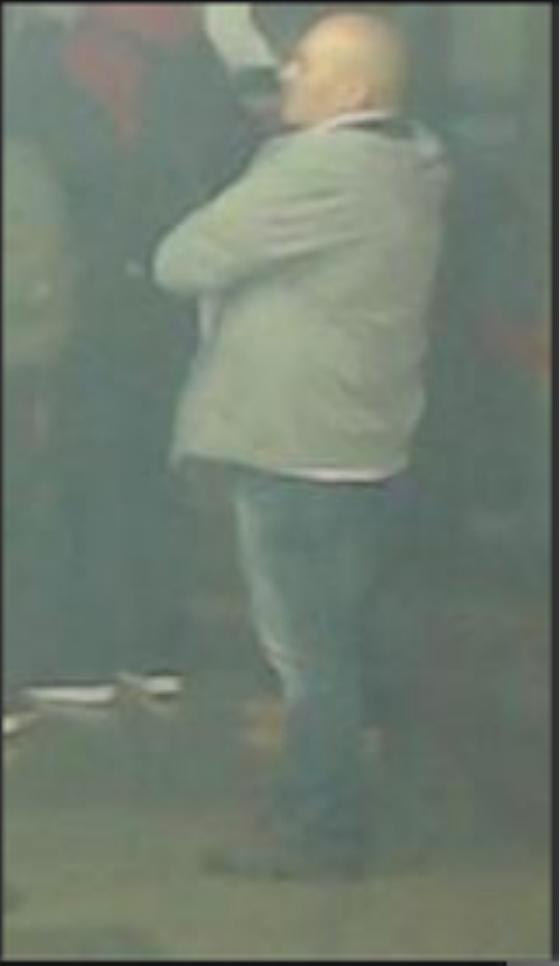}};       
\draw [ thick, rounded corners, red ] (-0.975+\deltaX, -1.675) rectangle (0.975+\deltaX,1.675);     

\draw [fill, red](-0.7+\deltaX, -1.95) circle(0.19);
\draw [white] (-0.7+\deltaX,-1.95) node {\scriptsize{\textbf{A$_1$}}};       

\draw (0.1+\deltaX,-1.95) node {\scriptsize{''\emph{ID: A}''}};     
\draw (0.1+\deltaX,-2.25) node {\scriptsize{''\emph{male}''}};     
\draw (0.1+\deltaX,-2.55) node {\scriptsize{''\emph{adult}''}};    
\draw (0.1+\deltaX,-2.85) node {\scriptsize{''\emph{bald}''}};    

\def\deltaX{2}
\draw (0+\deltaX,0) node(n1)  {\includegraphics[width=1.9 cm]{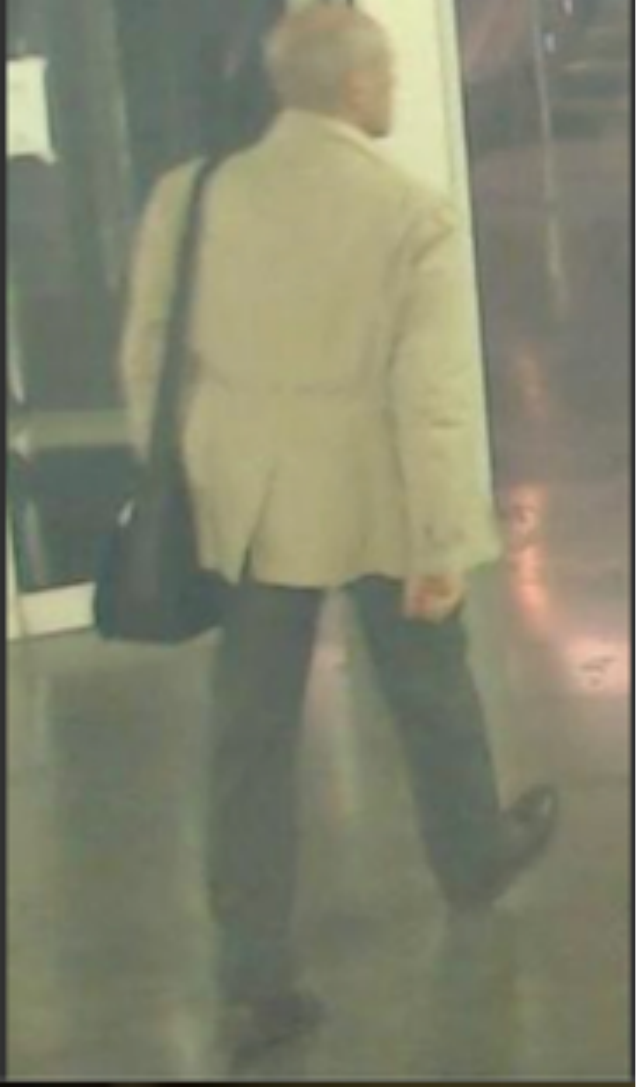}};       
\draw  [ thick, rounded corners, red ] (-0.975+\deltaX, -1.675) rectangle (0.975+\deltaX,1.675);       

\draw [fill, red](-0.7+\deltaX, -1.95) circle(0.19);
\draw [white] (-0.7+\deltaX,-1.95) node {\scriptsize{\textbf{A$_2$}}};       

\draw (0.1+\deltaX,-1.95) node {\scriptsize{''\emph{ID: A}''}};     
\draw (0.1+\deltaX,-2.25) node {\scriptsize{''\emph{male}''}};     
\draw (0.1+\deltaX,-2.55) node {\scriptsize{''\emph{adult}''}};    
\draw (0.1+\deltaX,-2.85) node {\scriptsize{''\emph{bald}''}};    

\def\deltaX{4}
\draw (0+\deltaX,0) node(n1)  {\includegraphics[width=1.9 cm]{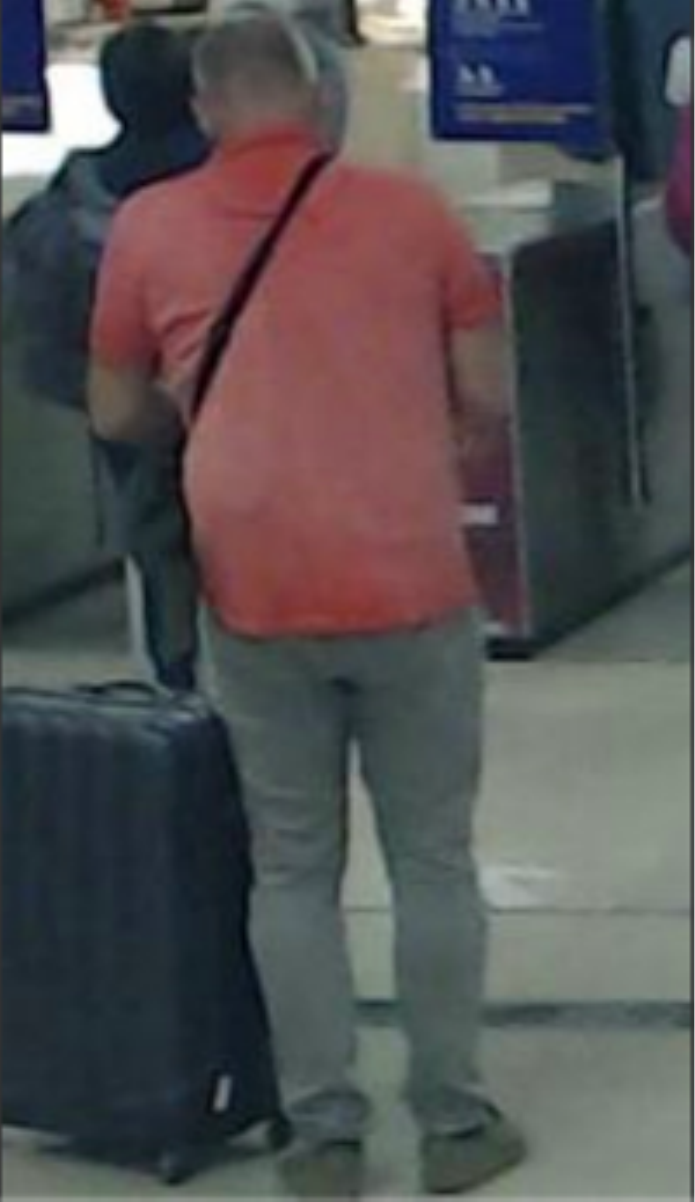}};       
\draw [ thick, rounded corners, blue ] (-0.975+\deltaX, -1.675) rectangle (0.975+\deltaX,1.675);      

\draw [fill, blue](-0.7+\deltaX, -1.95) circle(0.19);
\draw [white] (-0.7+\deltaX,-1.95) node {\scriptsize{\textbf{B}}};       

\draw (0.1+\deltaX,-1.95) node {\scriptsize{''\emph{ID: B}''}};     
\draw (0.1+\deltaX,-2.25) node {\scriptsize{''\emph{male}''}};     
\draw (0.1+\deltaX,-2.55) node {\scriptsize{''\emph{adult}''}};    
\draw (0.1+\deltaX,-2.85) node {\scriptsize{''\emph{bald}''}};    

\def\deltaX{6}
\draw (0+\deltaX,0) node(n1)  {\includegraphics[width=1.9 cm]{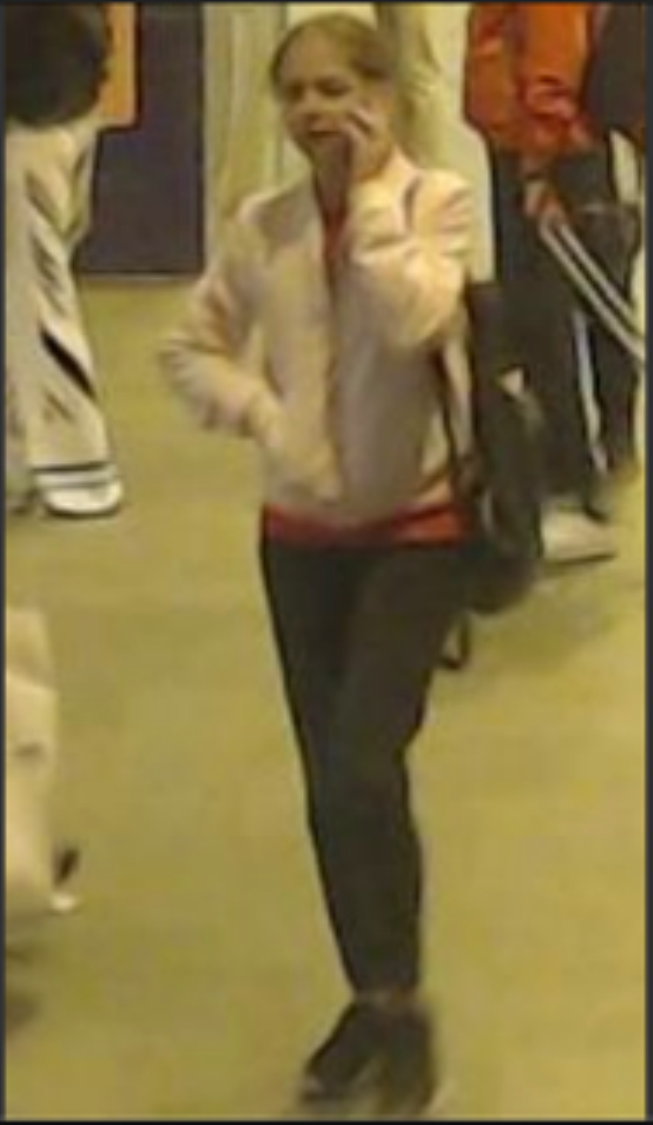}};       
\draw [ thick, rounded corners, black ] (-0.975+\deltaX, -1.675) rectangle (0.975+\deltaX,1.675);      

\draw [fill, black](-0.7+\deltaX, -1.95) circle(0.19);
\draw [white] (-0.7+\deltaX,-1.95) node {\scriptsize{\textbf{C}}};       

\draw (0.1+\deltaX,-1.95) node {\scriptsize{''\emph{ID: C}''}};     
\draw (0.1+\deltaX,-2.25) node {\scriptsize{''\emph{female}''}};     
\draw (0.1+\deltaX,-2.55) node {\scriptsize{''\emph{young}''}};    
\draw (0.1+\deltaX,-2.85) node {\scriptsize{''\emph{blond}''}};    

\def\deltaY{-1.2}

\draw [thick, dashed, rounded corners ] (-1, -2.0+\deltaY) rectangle (7, -3.25+\deltaY);      

\draw (3.0,-2.9+\deltaY) node {\small{\textbf{Learning}}};       

\draw [thick, ->] (2.5, -2.5+\deltaY) .. controls (3, -2.4+\deltaY) .. (3.5, -2.5+\deltaY);

\draw [] (-0.5, -2.5+\deltaY) -- (0.55, -2.9+\deltaY); 
\draw [] (-0.5, -2.5+\deltaY) -- (0.75, -2.25+\deltaY); 
\draw [] (-0.5, -2.5+\deltaY) -- (1.5, -2.85+\deltaY); 
\draw [] (1.5, -2.85+\deltaY) -- (0.55, -2.9+\deltaY); 

\draw [fill, red](-0.5, -2.5+\deltaY) circle(0.1);
\draw [fill, black](0.55, -2.9+\deltaY) circle(0.1);
\draw [fill, red](0.75, -2.25+\deltaY) circle(0.1);
\draw [fill, blue](1.5, -2.85+\deltaY) circle(0.1);

\def\deltaX{4.75}

\draw [] (-0.5+\deltaX, -2.5+\deltaY) -- (1.9+\deltaX, -2.4+\deltaY); 
\draw [] (-0.5+\deltaX, -2.5+\deltaY) -- (-0.05+\deltaX, -2.35+\deltaY); 
\draw [] (-0.5+\deltaX, -2.5+\deltaY) -- (0.20+\deltaX, -2.85+\deltaY); 
\draw [] (1.9+\deltaX, -2.4+\deltaY) -- (0.275+\deltaX, -2.85+\deltaY); 

\draw [fill, red](-0.5+\deltaX, -2.5+\deltaY) circle(0.1);
\draw [fill, black](1.9+\deltaX, -2.4+\deltaY) circle(0.1);
\draw [fill, red](-0.05+\deltaX, -2.35+\deltaY) circle(0.1);
\draw [fill, blue](0.275+\deltaX, -2.85+\deltaY) circle(0.1);

\end{tikzpicture}
    \caption{Likewise the triplet loss~\cite{Schroff2015}, the proposed \textbf{quadruplet} loss also seeks to minimize the distances between the elements of \emph{positive} pairs $\{A_1$, $A_2\}$, but simultaneously considers the relative similarity between the different classes ($A$, $B$ and $C$), yielding embeddings that are particularly suitable for identity retrieval. In this example, the proposed loss will privilege projections into the destiny space such that $d(A_1,A_2) < d(A_i,B) < d(A_i,C)$.}
        \label{fig:idea_1}
    \end{center}
\end{figure}
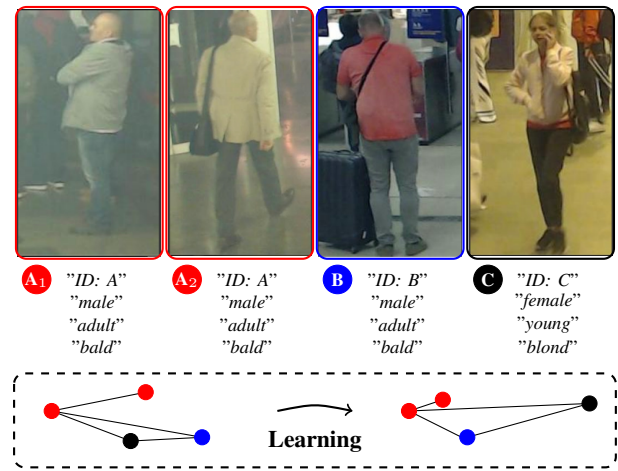

Based on the concept of triplet loss, this paper describes one objective function that can be regarded as a generalization of its predecessor. Instead of the binary division of the learning pairs into \emph{positive}/\emph{negative}, we  define a metric that analyzes the similarity between any two classes (identities). In learning time, four elements of arbitrary classes are considered at a time and the soft margins between the pairwise distances yield from the number of common labels in each pair (Fig.~\ref{fig:idea_1}).  Under this formulation, different identities that are semantically close to each other (e.g., two ''\emph{young, black, bald, male}'' subjects) are projected into adjacent regions of the destiny space. Also, as this objective function imposes different margins between two \emph{negative} pairs, it leverages the difficulties in mining appropriate learning instances, which is one on the main difficulties in the triplet loss formulation. 

The proposed loss function is particularly suitable for \emph{coarse-to-fine} classification problems, where some of the labels are easier to infer than others and the global problem can be decomposed into more tractable sub-components.This hierarchical paradigm is an efficient way of organizing object recognition, not only to accommodate a large number of  hypotheses, but also to systematically exploit any shared attributes. The identity retrieval problem is of particular interest, with the finest labels (IDs) being seen leaves of hierarchical structures with roots such as the ''gender'' or ''ethnicity'' features. \\

The remainder of this paper is organized as follows: Section~\ref{sec:Related} summarizes the most relevant research in the scope of our work. Section~\ref{sec:ProposedMethod} provides a detailed description of the proposed objective function. In Section~\ref{sec:Results} we discuss the obtained results and the conclusions are given in Section~\ref{sec:Conclusions}.

\section{Related Work}
\label{sec:Related}

Deep learning methods for biometrics can be roughly divided into two major groups: 1) methods that directly learn multi-class classifiers used in identity retrieval and soft biometrics inference; and 2) methods that learn low-dimensional feature embeddings, where identification yields from nearest neighbour search. 

\subsection{Soft Biometrics and Identity Retrieval}

Bekele \etal~\cite{Bekele2017}  proposed a residual network for multi-output inference that  handles classes-imbalance directly in the cost function, without depending of data augmentation techniques.  Almudhahka \etal~\cite{Almudhahka2017} explored the concept of comparative soft biometrics and assessed the impact of automatic estimations  on face retrieval performance. Guo \etal~\cite{Guo2018} studied the influence of distance in the effectiveness of body and facial soft biometrics, introducing a joint density distribution based rank-score fusion strategy~\cite{Guo2018b}. Vera-Rodriguez \etal~\cite{Rodriguez2017} used hand-crafted features extracted from the distances between key points in body  silhouettes. Martinho-Corbishley \etal~\cite{Martinho2019} introduced the idea of \emph{super-fine} soft attributes, describing multiple concepts of one trait as multi-dimensional perceptual coordinates. Also, using joint attribute regression and a deep residual CNN, they observed substantially better ranked retrieval performance  in comparison to conventional labels. Schumann and Specker used an ensemble of classifiers for robust attribute inference~\cite{Schumann2018}, extended to full body search by combining it with a human silhouette detector. He \etal~\cite{He2017} proposed a weighted multi-task CNN with a loss term that dynamically updates the weight for each task during the learning phase. 

Several works relied regarded semantic segmentation as a tool to support labels inference: Galiyawala \etal~\cite{Galiyawala2018} described a deep learning framework  for person retrieval using the height, clothes' color, and gender labels, with a semantic segmentation module used to remove clutter. Similarly, Cipcigan and Nixon~\cite{Cipcigan2018} obtained semantically segmented regions of the body, that subsequently fed two CNN-based feature extraction and inference modules. 

Finally, specifically designed for handheld devices, Samangouei and Chellappa~\cite{Samangouei2016}  extracted facial soft biometric information from mobile phones, while Neal and Woodard~\cite{Neal2019} developed a human retrieval scheme based on thirteen demographic and behavioural attributes from mobile phones data, such as calling, SMS and application data, having authors positively concluded about the feasibility of this kind of recognition. 

A comprehensive summary of the most relevant research in soft biometrics is given in~\cite{Sosa2018}.

\subsection{Feature Embeddings and Loss Functions}

Triplet loss functions were motivated by the concept of \emph{contrastive} loss~\cite{Hadsell2006}, where the rationale is to penalize distances between \emph{positive} pairs, while favouring  distances  between \emph{negative} pairs. Kang \etal~\cite{Kang2017} used a deep ensemble of multi-scale CNNs, each one based on triplet loss functions.  Song \etal~\cite{Song2016}  learned semantic feature embeddings that lift the vector of pairwise distances within the batch to the matrix of pairwise distances, and described a structured loss on the lifted problem. Liu and Huan~\cite{Liu2017}  proposed a triplet loss learning architecture composed of four CNNs, each one learning features from different body parts that are fused at the score level.  

A posterior concept was the  \emph{center} loss~\cite{Wen2016}, which finds a center for elements of each class and penalizes the distances between the projections and their corresponding class centers. Jian \etal~\cite{Jiang2018} combined additive margin \emph{softmax} with center loss to increase inter-class distances and avoid overconfidence on  classifications.  Ranjan \etal~\cite{Ranjan2015} described the \emph{crystal} loss, that restricts the features to lie on a hypersphere of a fixed radius, by adding  a constraint on the features projections such that their $\ell_2$-norm remains constant. Chen \etal~\cite{Chen2015} used deep representations to feed a joint Bayesian metrics learning module that maximizes the log-likelihood ratio between intra- and inter-classes distances. Based on the concept of \emph{Sphereface}, Deng \etal~\cite{Deng2019}  proposed an additive angular margin loss, with a clear geometric interpretation due to the correspondence to the geodesic distance on the hypersphere.

Observing that CNN-based methods tend to overfit in person re-identification tasks, Shi \etal~\cite{Shi2015} used siamese architectures to provide a joint description to a metric learning module,  regularizing the learning process and improving the generalization ability. Also, to cope with large intra-class variations, they suggested the idea of \emph{moderate positive mining}, again to prevent overfitting. Motivated by the difficulties in generate learning instances for triplet loss frameworks, Su \etal~\cite{Su2017} performed adaptive CNN fine-tuning, along with an adaptive loss function that relates the maximum distance among positive pairs to the margin demanded for separate the \emph{positive} from the \emph{negative} pairs. Hu \etal~\cite{Hu2017} proposed an objective function that generalizes the Maximum Mean Discrepancy~\cite{Schlkopf2008} metric, with a weighting scheme that favours good quality data. Duan \etal~\cite{Duan2019} proposed the \emph{uniform} loss to learn deep equi-distributed representations for face recognition. Finally,  observing the typical unbalance between positive and negative pairs, Wang \etal~\cite{Wang2017} described an adaptive margin list-wise loss, in which learning data are provided with a set of negative pairs divided into three classes (\emph{easy}, \emph{moderate}, and \emph{hard}), depending of the distance rank with respect to the query.

Finally, we note the differences between the work proposed in this paper and the (also \emph{quadruplet}) loss described by Chen \etal~\cite{Chen2017}. These authors attempt to augment the inter-class margins and the intra-class compactness without explicitly using any semantical constraint, in the sense that -- as in the original triplet loss formulation -- there is nothing that explicitly enforces to project similar classes (i.e., different identities that share most of the remaining labels) to neighbour regions of the latent space. In opposition, our method  concerns essentially about this semantical coherence of assuring the projection \emph{similar} classes into adjacent regions and not in obtain larger margins between the different classes. Also, even the idea behind the loss formulation is radically different in both methods, in the sense that ~\cite{Chen2017}  still considers the concept of \emph{anchor} (as the original triplet-loss formulation), which is - again - in opposition to our proposal.

 \section{Proposed Method}
 \label{sec:ProposedMethod}

\begin{figure*}[ht!]
\begin{center}
  \begin{tikzpicture}
  
\draw (0,0.0) node {\scriptsize{\textbf{ID}}};          
\draw (0.95,0.0) node {\scriptsize{\textbf{Gender}}};        
\draw (2.15,-0.02) node {\scriptsize{\textbf{Ethnicity}}};        

\draw (4.15,-0.02) node {\scriptsize{\textbf{Elements}}};       

\draw [dashed, thick] (0.35, 0.1) -- (0.35, -2.3); 
\draw [dashed, thick] (1.5, 0.1) -- (1.5, -2.3); 
\draw [dashed, thick] (2.75, 0.1) -- (2.75, -2.3);

\draw [dashed, very thick] (5.5, 0.1) -- (5.5, -2.5); 
\draw [dashed, very thick] (-0.5, -2.6) -- (5.5, -2.6); 

\draw (0,-0.5) node {\scriptsize{x} - $\bullet$};          
\draw (0,-1.25) node {\scriptsize{y - $\circ$}};  
\draw (0,-2.0) node {\scriptsize{z - $\diamond$}};  

\draw (0.95,-0.5) node {\scriptsize{Female}};          
\draw (0.95,-1.25) node {\scriptsize{Female}};  
\draw (0.95,-2.0) node {\scriptsize{Male}};  

\draw (2.15,-0.5) node {\scriptsize{Black}};          
\draw (2.15,-1.25) node {\scriptsize{White}};  
\draw (2.15,-2.0) node {\scriptsize{White}};

\draw (3.5,-0.5) node(n1)  {\includegraphics[width=0.6 cm]{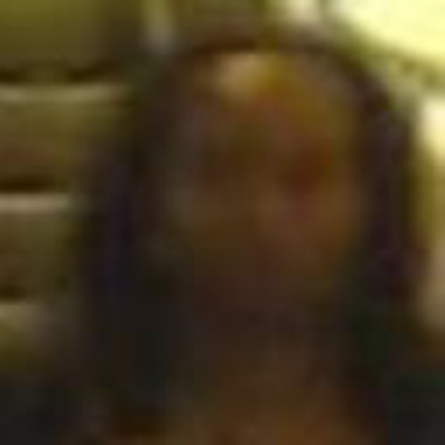}};       
\draw (4.6,-0.5) node(n1)  {\includegraphics[width=0.6 cm]{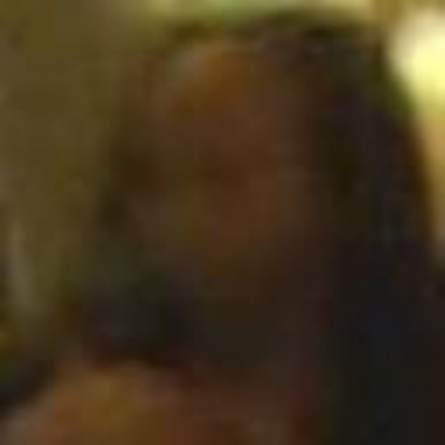}};       
\draw (4.0,-0.5) node {\scriptsize{$\bm{x}_1$}};          
\draw (5.1,-0.5) node {\scriptsize{$\bm{x}_2$}};

\draw (3.5,-1.25) node(n1)  {\includegraphics[width=0.6 cm]{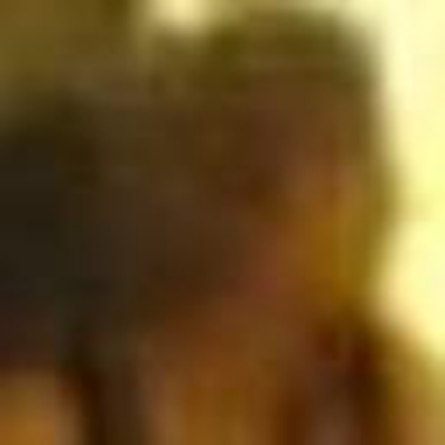}};       
\draw (4.6,-1.25) node(n1)  {\includegraphics[width=0.6 cm]{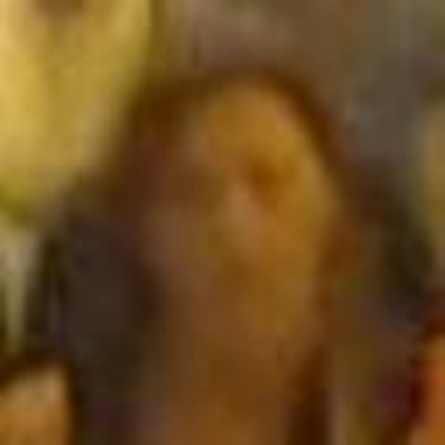}};      
\draw (4.0,-1.25) node {\scriptsize{$\bm{y}_1$}};          
\draw (5.1,-1.25) node {\scriptsize{$\bm{y}_2$}};       

\draw (3.5,-2.0) node(n1)  {\includegraphics[width=0.6 cm]{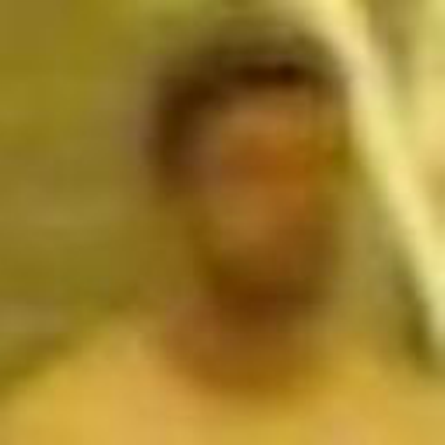}};       
\draw (4.6,-2.0) node(n1)  {\includegraphics[width=0.6 cm]{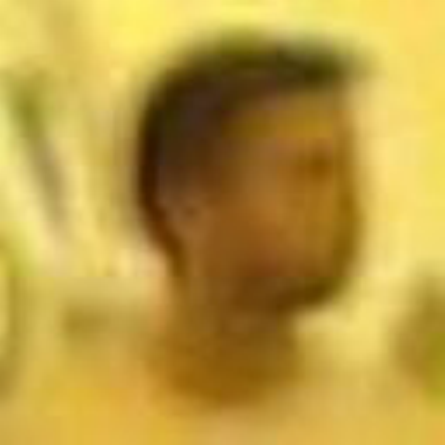}};      
\draw (4.0,-2.0) node {\scriptsize{$\bm{z}_1$}};          
\draw (5.1,-2.0) node {\scriptsize{$\bm{z}_2$}};

%%%%%%%%%%%%%%%%%%%%%%%%%%%%%%%%%%%%%%%%%%%
%% BAD PROJECTION  1    
%%%%%%%%%%%%%%%%%%%%%%%%%%%%%%%%%%%%%%%%%%%

\def\deltaX{0}
\def\deltaY{-0.75}

\draw (1.15+\deltaX, -2.75+\deltaY) node {$\Psi_1$};    
\draw [thick, dashed, rounded corners, fill=red!10 ] (-0.5+\deltaX, -3.0+\deltaY) rectangle (2.8+\deltaX, -4.6+\deltaY);      

\draw (0+\deltaX,-3.75+\deltaY) node {$\bullet$};    
\draw (0.15+\deltaX,-3.6+\deltaY) node {\scriptsize{$\bm{x}_1$}};     

\draw (0.2+\deltaX,-3.95+\deltaY) node {$\bullet$};    
\draw (0.35+\deltaX,-3.8+\deltaY) node {\scriptsize{$\bm{x}_2$}};     
  
\draw (1.15+\deltaX,-3.75+\deltaY) node {$\diamond$};    
\draw (1.35+\deltaX,-3.6+\deltaY) node {\scriptsize{$\bm{z}_1$}};     
\draw (1.2+\deltaX,-3.95+\deltaY) node {$\diamond$};        
\draw (1.45+\deltaX,-4.05+\deltaY) node {\scriptsize{$\bm{z}_2$}};     

\draw (2.15+\deltaX,-4.0+\deltaY) node {$\circ$};    
\draw (2.35+\deltaX,-4.15+\deltaY) node {\scriptsize{$\bm{y}_1$}};     
\draw (2.2+\deltaX,-3.75+\deltaY) node {$\circ$};         
\draw (2.4+\deltaX,-3.6+\deltaY) node {\scriptsize{$\bm{y}_2$}};            

%%%%%%%%%%%%%%%%%%%%%%%%%%%%%%%%%%%%%%%%%%%
%% BAD PROJECTION  2   
%%%%%%%%%%%%%%%%%%%%%%%%%%%%%%%%%%%%%%%%%%%

\def\deltaX{3.75}

\draw (1.15+\deltaX, -2.75+\deltaY) node {$\Psi_2$};    
\draw [thick, dashed, rounded corners, fill=red!10 ] (-0.5+\deltaX, -3.0+\deltaY) rectangle (2.8+\deltaX, -4.6+\deltaY);      

\draw (0+\deltaX,-3.75+\deltaY) node {$\circ$};    
\draw (0.15+\deltaX,-3.6+\deltaY) node {\scriptsize{$\bm{y}_1$}};     

\draw (0.2+\deltaX,-3.95+\deltaY) node {$\circ$};    
\draw (0.35+\deltaX,-3.8+\deltaY) node {\scriptsize{$\bm{y}_2$}};     
  
\draw (1.15+\deltaX,-3.75+\deltaY) node {$\bullet$};    
\draw (1.35+\deltaX,-3.6+\deltaY) node {\scriptsize{$\bm{x}_1$}};     
\draw (1.2+\deltaX,-3.95+\deltaY) node {$\bullet$};        
\draw (1.45+\deltaX,-4.05+\deltaY) node {\scriptsize{$\bm{x}_2$}};     

\draw (2.15+\deltaX,-4.0+\deltaY) node {$\diamond$};    
\draw (2.35+\deltaX,-4.15+\deltaY) node {\scriptsize{$\bm{z}_1$}};     
\draw (2.2+\deltaX,-3.75+\deltaY) node {$\diamond$};         
\draw (2.4+\deltaX,-3.6+\deltaY) node {\scriptsize{$\bm{z}_2$}};

%%%%%%%%%%%%%%%%%%%%%%%%%%%%%%%%%%%%%%%%%%%
%% GOOD PROJECTION      
%%%%%%%%%%%%%%%%%%%%%%%%%%%%%%%%%%%%%%%%%%%

\def\deltaX{7.5}

\draw (1.15+\deltaX, -2.75+\deltaY) node {$\Psi_3$};    
\draw [thick, dashed, rounded corners, fill=green!10 ] (-0.5+\deltaX, -3.0+\deltaY) rectangle (2.8+\deltaX, -4.6+\deltaY);      

\draw (0+\deltaX,-3.75+\deltaY) node {$\bullet$};    
\draw (0.15+\deltaX,-3.6+\deltaY) node {\scriptsize{$\bm{x}_1$}};     

\draw (0.2+\deltaX,-3.95+\deltaY) node {$\bullet$};    
\draw (0.35+\deltaX,-3.8+\deltaY) node {\scriptsize{$\bm{x}_2$}};     
  
\draw (1.15+\deltaX,-3.75+\deltaY) node {$\circ$};    
\draw (1.35+\deltaX,-3.6+\deltaY) node {\scriptsize{$\bm{y}_1$}};     
\draw (1.2+\deltaX,-3.95+\deltaY) node {$\circ$};        
\draw (1.45+\deltaX,-4.05+\deltaY) node {\scriptsize{$\bm{y}_2$}};     

\draw (2.15+\deltaX,-4.0+\deltaY) node {$\diamond$};    
\draw (2.35+\deltaX,-4.15+\deltaY) node {\scriptsize{$\bm{z}_1$}};     
\draw (2.2+\deltaX,-3.75+\deltaY) node {$\diamond$};         
\draw (2.4+\deltaX,-3.6+\deltaY) node {\scriptsize{$\bm{z}_2$}};

\draw (8.5,-0.0) node {\textbf{Triplet Loss}};     

\draw (8.5,-2.4) node {\rotatebox{270}{$\Rightarrow$}};     

 \fill [rounded corners, black] (5.6, -2.65) rectangle (10.90, -3.05);    
\draw  [white] (8.25, -2.85) node {\small{Embeddings $\Psi_1, \Psi_2, \Psi_3$ are possible}};

\draw (8.25,-0.75) node {\scriptsize{$\| f(\bm{x}_1) - f(\bm{x}_2)\|_2^2 + \alpha < \| f(\bm{x}_i) - f(\bm{y}_j)\|_2^2 $}};       

\draw (8.25,-1.25) node {\scriptsize{$\forall i, j \in \{1,2\} $}};       

\draw (6.75,-1.25) node {\scriptsize{$\vdots$}};  
\draw (6.75,-2.0) node {\scriptsize{\textbf{Positive pairs}}};  
% \draw [->] (7.35, -1.85) .. controls (7.5, -1.75) .. (7.25, -1.1);

\draw (8.35,-2.0) node {\scriptsize{\textbf{ $<$}}};  

\draw (9.75,-1.25) node {\scriptsize{$\vdots$}};  
\draw (9.75,-2.0) node {\scriptsize{\textbf{Negative pairs}}};  

%\draw (8.25,-1.25) node {\scriptsize{$\| f(\textbf{y}_1) - f(\textbf{y}_2)\|_2^2 + \alpha < \| f(\textbf{y}_i) - f(\textbf{z}_j)\|_2^2 $}};   

%\draw (8.25,-2.0) node {\scriptsize{$\| f(\textbf{z}_1) - f(\textbf{z}_2)\|_2^2 + \alpha < \| f(\textbf{z}_i) - f(\textbf{x}_j)\|_2^2 $}};   

\draw [dashed, very thick] (11.0, 0.1) -- (11.0, -5.5); 

\draw [fill, black](11.3, -0.5) circle(0.175);
\draw [white] (11.3,-0.5) node {\scriptsize{\textbf{1}}};       
    
\draw [fill, black](11.3, -1.25) circle(0.175);
\draw [white] (11.3,-1.25) node {\scriptsize{\textbf{2}}};       

\draw [fill, black](11.3, -2.0) circle(0.175);
\draw [white] (11.3,-2.0) node {\scriptsize{\textbf{3}}};       
 
\draw (14.25,-0.5) node {\scriptsize{$\| f(\bm{y}_i) - f(\bm{y}_j)\|_2^2 + \alpha < \| f(\bm{y}_k) - f(\bm{z}_l)\|_2^2 $}};  

\draw (14.25,-1.25) node {\scriptsize{$\| f(\bm{z}_i) - f(\bm{z}_j)\|_2^2 + \alpha < \| f(\bm{x}_k) - f(\bm{y}_l)\|_2^2 $}};  

\draw (14.25,-2.0) node {\scriptsize{$\| f(\bm{x}_i) - f(\bm{y}_j)\|_2^2 + \alpha < \| f(\bm{x}_k) - f(\bm{z}_l)\|_2^2 $}};  

\draw (14.5,0.0) node {\textbf{Proposed Quadruplet Loss}};     

\def\deltaY{-2.5}
\draw [fill, black](11.4, -0.5+\deltaY) circle(0.175);
\draw [white] (11.4,-0.5+\deltaY) node {\scriptsize{\textbf{1}}};       
\draw (14.0,-0.5+\deltaY) node {\scriptsize{\textbf{Positive pairs $<$ Negative pairs}}};

\draw [fill, black](11.4, -1.0+\deltaY) circle(0.175);
\draw [white] (11.4,-1.0+\deltaY) node {\scriptsize{\textbf{2}}};       
\draw (14.0,-1.0+\deltaY) node {\scriptsize{\textbf{Positive pairs $\ll$ \emph{More} Negative pairs}}};  

\draw [fill, black](11.4, -1.5+\deltaY) circle(0.175);
\draw [white] (11.4,-1.5+\deltaY) node {\scriptsize{\textbf{3}}};     
\draw (14.0,-1.5+\deltaY) node {\scriptsize{\textbf{Negative pairs $<$ \emph{More} Negative pairs}}};

\fill [rounded corners, black] (12.00, -4.8) rectangle (16.00, -5.2);    
\draw  [white] (14.0, -5.0) node {\small{Embedding $\Psi_3$ is enforced}};    

\draw (14.0,-4.5) node {\rotatebox{270}{$\Rightarrow$}};     
      
\end{tikzpicture}
    \caption{Illustration of the key difference between the triplet loss~\cite{Schroff2015} and the solution proposed in this paper. Using a loss function that analyzes the semantic similarity (in terms of soft biometrics) between the different identities, we enforce embeddings ($\Psi_3$) that are semantically coherent, i.e., where: 1) elements of the same class appear near each other; and also 2) elements of relatively similar classes appear closer to each other than elements with no labels in common. This is in opposition to the original formulation of the triplet loss, that relies exclusively in the elements appearance to define the geometry of the destiny space, which might result - in case of noisy image features - in semantically incoherent embeddings (e.g., in $\Psi_1$ and $\Psi_2$, classes are compact and discriminative, but the $\bm{x}/\bm{z}$ centroids are too close to each other).}
        \label{fig:Ocular}
    \end{center}
\end{figure*}
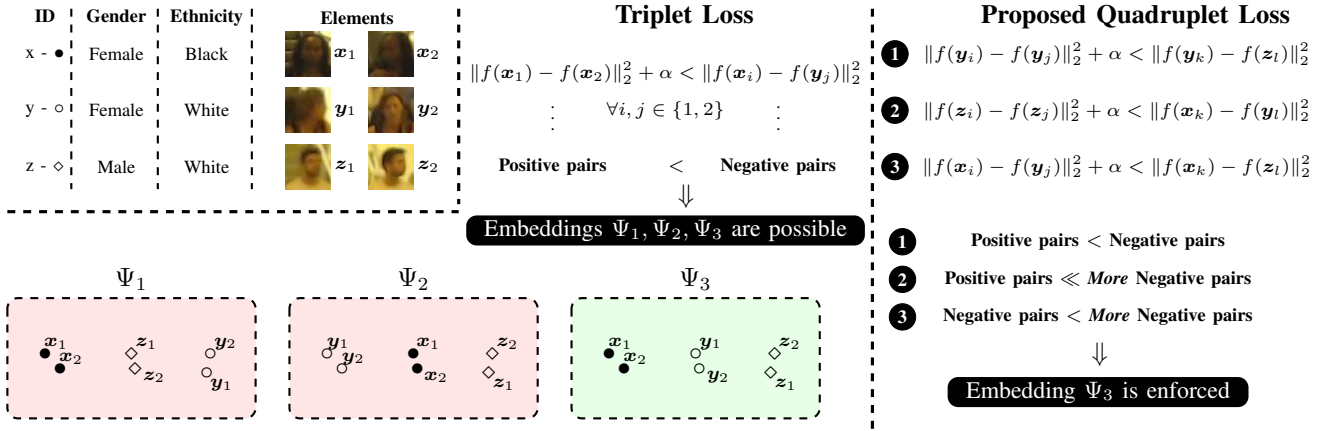
 
 \subsection{Quadruplet Loss: Definition}
 
Consider a supervised classification problem, where $t$ is the dimensionality of the response variable $\bm{y}_i$ associated to the input element $\bm{x}_i  \in [0, 255]^n$. Let $f(.)$ be one embedding function that maps $\bm{x}_i$ into a d-dimensional space $\Psi$, with $\bm{f}_i=f(\bm{x}_i) \in \Psi$ being the projected vector. Let $\{\bm{x}_1,\ldots, \bm{x}_b\}$ be a batch of $b$ images from the learning set.  We define $\phi(\bm{y}_i, \bm{y}_j) \in \mathbb{N}, ~\forall i, j \in \{1,\ldots,b\}$ as the function that measures the semantic similarity between $\bm{x}_i$ and $\bm{x}_j$:
  
\begin{align}
\phi(\bm{y}_i, \bm{y}_j)  = || \bm{y}_i - \bm{y}_j ||_0, 
 \label{eq:Loss1}
 \end{align}
with $||.||_0$ being the $\ell_0$-norm operator. 
 
In practice, $\phi(., .)$ counts the number of disagreeing labels between the $\{\bm{x}_i, \bm{x}_j\}$ pair, i.e., $\phi(\bm{y}_i, \bm{y}_j) = t$  when the i$^{th}$ and j$^{th}$ elements have fully disjoint classes membership (e.g., one "\emph{black, adult, male}'' and another ''\emph{white, young, female}'' subjects), while  $\phi(\bm{y}_1, \bm{y}_2) = 0$ when they have the exact same label (class) across all dimensions, i.e., when they constitute a \emph{positive} pair. 

Let $\{i, j, p, q\}$ be the indices of four images in the batch. The corresponding quadruplet loss value $\ell_{i, j, p, q}$  is given by:
 
  \begin{align}
 \ell_{i,j,p,q} &= \emph{sgn}\Big( \phi(\bm{y}_i, \bm{y}_j) -   \phi(\bm{y}_p, \bm{y}_q) \Big) \nonumber \\
 &  \Big[ \big(   \| \bm{f}_p - \bm{f}_q)\|_2^2 - \| \bm{f}_i - \bm{f}_j\|_2^2   \big) + \alpha \Big],
 \label{eq:Loss2}
 \end{align}
 where $\emph{sgn}()$ is the sign function and $\alpha$ is the desired margin ($\alpha = 0.1$ was used in all our experiments). Evidently, $\ell_.$ will be zero when both image pairs have the same number of agreeing labels (as $\emph{sgn}(0)=0$ in these cases). In all other cases, the sign function will determine the pairs for which the distances in the embedding should be minimized, i.e., if the elements of the $(p,q)$ pair are semantically closer to each other than the elements of the  $(i,j)$ pair \Big(i.e.$, \phi(\bm{y}_p, \bm{y}_q) <  \phi(\bm{y}_i, \bm{y}_j)$\Big), we want to ensure that $  \| \bm{f}_p - \bm{f}_q)\|_2^2 < \| \bm{f}_i - \bm{f}_j\|_2^2$.
 
 The accumulated loss in the batch is given by the truncated mean of a sample (of size $s$) randomly taken from the subset of the ${b}\choose{4}$ individual loss values where $\phi(\bm{y}_i, \bm{y}_j) \neq \phi(\bm{y}_p, \bm{y}_q)$:
 
 \begin{align}
 \mathcal{L} = \frac{1}{{s}}~\sum_{\bm{z}=1}^{s} \Big[\ell_{\bm{z}}\Big]_+, 
 \label{eq:Loss3}
 \end{align}
 where $\bm{z} \in \{1, \dots, s\}^4$ denotes the z$^{th}$ composition of four elements in the batch and $[.]_+$ is the $\max(., 0)$ function. Even considering that a large fraction of the combinations in the batch will be invalid (i.e., with $\phi()=0$), large values of $b$ will result in an intractable number of combinations at each iteration. In practical terms, after filtering out those invalid  combinations, we randomly sample a subset of the remaining instances, which is designated as the mini-batch.

\subsection{Quadruplet Loss: Inference}

Consider four indices $\{i, j, p, q\}$ of elements in the mini-batch, with $\phi(\bm{y}_i, \bm{y}_j) > \phi(\bm{y}_p, \bm{y}_q)$.  Let $\Delta_{\phi}$ denote the difference between the number of disagreeing labels of the $\{i,j\}$ and $\{p,q\}$ pairs:

\begin{align}
\Delta_{\phi}=\phi(\bm{y}_{i}, \bm{y}_{j}) -  \phi(\bm{y}_{p}, \bm{y}_{q}). 
\label{eq:implementation0}
\end{align}

Also, let $\Delta_{\bm{f}}$ be the distance between the elements of the most alike pair minus the distance between the elements of the least alike pair in the destiny space (plus the margin):

\begin{align}
\Delta_{\bm{f}}=\| \bm{f}_{p} - \bm{f}_{q})\|_2^2 - \| \bm{f}_{i} - \bm{f}_{j}\|_2^2 + \alpha. 
\label{eq:implementation01}
\end{align}

Upon basic algebraic manipulation, the gradients of $\mathcal{L}$ with respect to the quadruplet terms are given by:

 \begin{align}
\frac{\partial \mathcal{L}}{\partial \bm{f}_{i}}= \sum_{\bm{z}} \left\{
 \begin{array}{rl}
 2 (\bm{f}_{j} - \bm{f}_{i})    & \text{, if }   \Delta_{\phi} > 0~\land \Delta_{\bm{f}} \ge 0\\
   0 & \text{, otherwise} 
 \end{array} \right.
 \label{eq:implementation1}
 \end{align}

 \begin{align}
\frac{\partial \mathcal{L}}{\partial \bm{f}_{j}}= \sum_{\bm{z}} \left\{
 \begin{array}{rl}
 2 (\bm{f}_{i} - \bm{f}_{j})    & \text{, if }   \Delta_{\phi} > 0~\land \Delta_{\bm{f}} \ge 0\\
   0 & \text{, otherwise} 
 \end{array} \right.
 \label{eq:implementation2}
 \end{align}
 
  \begin{align}
\frac{\partial \mathcal{L}}{\partial \bm{f}_{p}}= \sum_{\bm{z}} \left\{
 \begin{array}{rl}
 2 (\bm{f}_{p} - \bm{f}_{q})    & \text{, if }   \Delta_{\phi} > 0~\land \Delta_{\bm{f}} \ge 0\\
   0 & \text{, otherwise} 
 \end{array} \right.
 \label{eq:implementation3}
 \end{align}
 
  \begin{align}
\frac{\partial \mathcal{L}}{\partial \bm{f}_{q}}= \sum_{\bm{z}} \left\{
 \begin{array}{rl}
 2 (\bm{f}_{q} - \bm{f}_{p})    & \text{, if }   \Delta_{\phi} > 0~\land \Delta_{\bm{f}} \ge 0\\
   0 & \text{, otherwise} 
 \end{array} \right.
 \label{eq:implementation4}
 \end{align}
 
In practice terms, the model weights will be adjusted only for learning instances where pairs have a different number of agreeing labels (i.e.,  $\Delta_{\phi} > 0$) and when the distance in the destiny space between the elements of the most similar pair is higher or equal than the distance between the elements of the least similar pair (plus the margin, $\Delta_{\bm{f}} \ge 0$). According to this idea, using (\ref{eq:implementation1})-(\ref{eq:implementation4}), the deep learning frameworks supervised by the proposed quadruplet loss are trainable in a way similar to its counterpart triplet loss and can be optimized by the standard Stochastic Gradient Descend (SGD) algorithm, which was done in all our experiments. 

For clarity purposes, Algorithm~\ref{alg:inference} gives a pseudocode description of the learning and batch/mini-batch creation processes during the inference step for the proposed loss function.

\newcommand*\DNA{\textsc{dna}}
\newcommand*\Let[6]{\State #1 $\gets$ #2}  
\algrenewcommand\algorithmicrequire{\textbf{Precondition:}}  
\algrenewcommand\algorithmicensure{\textbf{Postcondition:}}

\newcommand{\round}[1]{\ensuremath{\lfloor#1\rceil}}

\begin{algorithm}  
  \caption{Pseudocode description of the inference and batch/min-batch creation processes.  
    \label{alg:inference}}  
  \begin{algorithmic}
    \Require{$\bm{M}$: CNN, $t_e$: Tot. epochs, $s$: mini-batch size, $b$: batch size, $\bm{I}$: Learning set, $n$ images}  
%    \Statex
    \For{$1 \textrm{ to } t_e$} 
    	\For{$1 \textrm{ to } \round{\frac{n}{s}}$} \\
    		\hspace{1cm} $\bm{b} \gets$ randomly sample $b$ out of $n$ images from $\bm{I}$ \\ 		
    		\hspace{1cm} $\bm{c} \gets$  create ${b}\choose{4}$ quadruplet combinations from $\bm{b}$  \\		
		 \hspace{1cm} $\bm{c}* \gets$  filter out invalid elements from $\bm{c}$  	\\	
		 \hspace{1cm} $\bm{s} \gets$  randomly sample $s$ elements from $\bm{c}^*$  \\				 
		  \hspace{1cm}$\bm{M} \gets$  update weights($\bm{M}, \bm{s}$) (eqs. (\ref{eq:implementation1}-\ref{eq:implementation4}))
	\EndFor	
    \EndFor    \\
    \Return $\bm{M}$
  \end{algorithmic}  
\end{algorithm}

\subsection{Quadruplet Loss: Insight and Example}

Fig.~\ref{fig:Ocular} illustrates our rationale in the proposed quadruplet loss. By defining a metric that analyses the similarity between two classes, we abandon the binary division of pairs into \emph{positive}/\emph{negative} families, and perceive which classes are more/less distinct with respect to others. This enables to explicitly enforce that the \emph{most} negative pairs (e.g., with no common labels) are at the farthest possible distance from each other in the embedding. During the learning phase, we sample image pairs in a stochastic way and enforce that input elements are projected in a way that faithfully resembles the human perception of \emph{semantic similarity}.

\begin{figure}[ht!]
\begin{center}
\begin{tikzpicture}

\draw (0,0) node(n1)  {\includegraphics[width=8.0 cm]{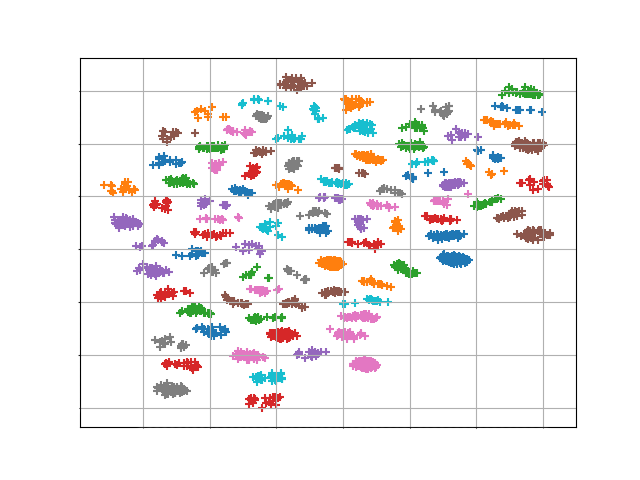}};     

\draw (0, -3.0) node[rectangle] {\small{\textbf{Triplet Loss}}}; 

%%%%%%%%%%%%%%%%%%%%%%%%%%%%%%%%%%%%%%%%%%%%%%%%
%%	TRIPLET
%%%%%%%%%%%%%%%%%%%%%%%%%%%%%%%%%%%%%%%%%%%%%%%%
%Top right triplet
\draw [thick] (2.5, 2.35) -- (2.5, 1.85); 
\draw (2.5,2.35) node(n1)  {\includegraphics[width=0.5 cm]{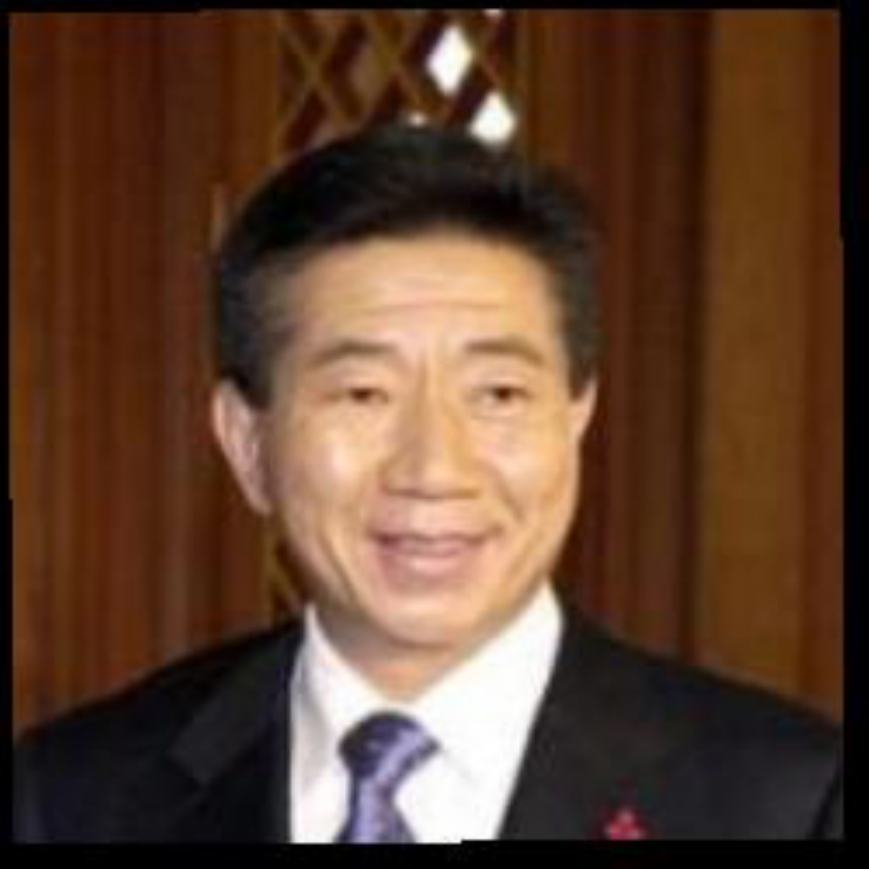}};     

\draw [thick] (3.25, 1.25) -- (2.65, 1.20); 
\draw (3.25,1.25) node(n1)  {\includegraphics[width=0.5 cm]{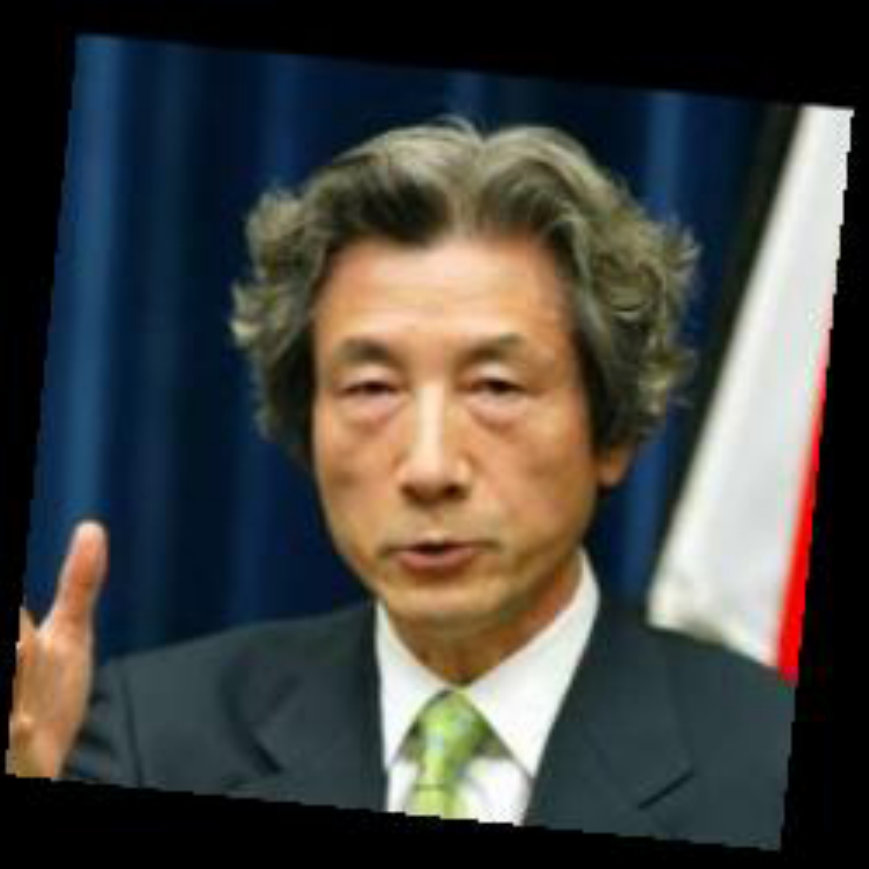}};     

\draw [thick] (3.25, 0.5) -- (2.75, 0.70); 
\draw (3.25,0.5) node(n1)  {\includegraphics[width=0.5 cm]{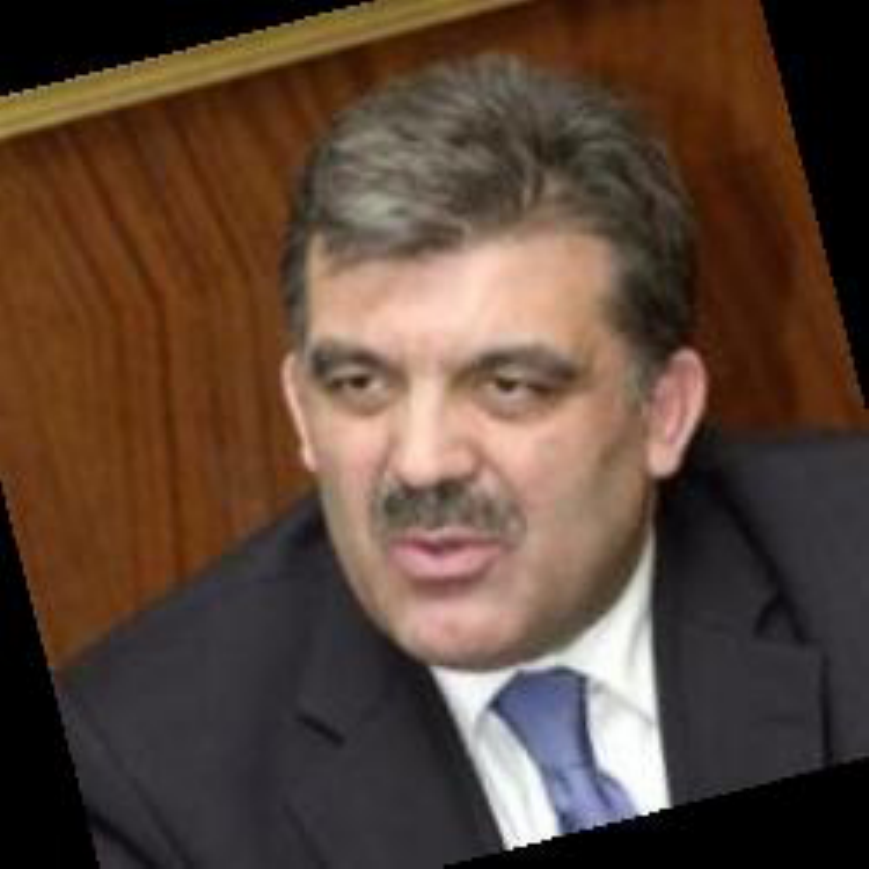}};     

\draw [thick] (3.05, -0.5) -- (2.75, 0.00); 
\draw (3.05,-0.5) node(n1)  {\includegraphics[width=0.5 cm]{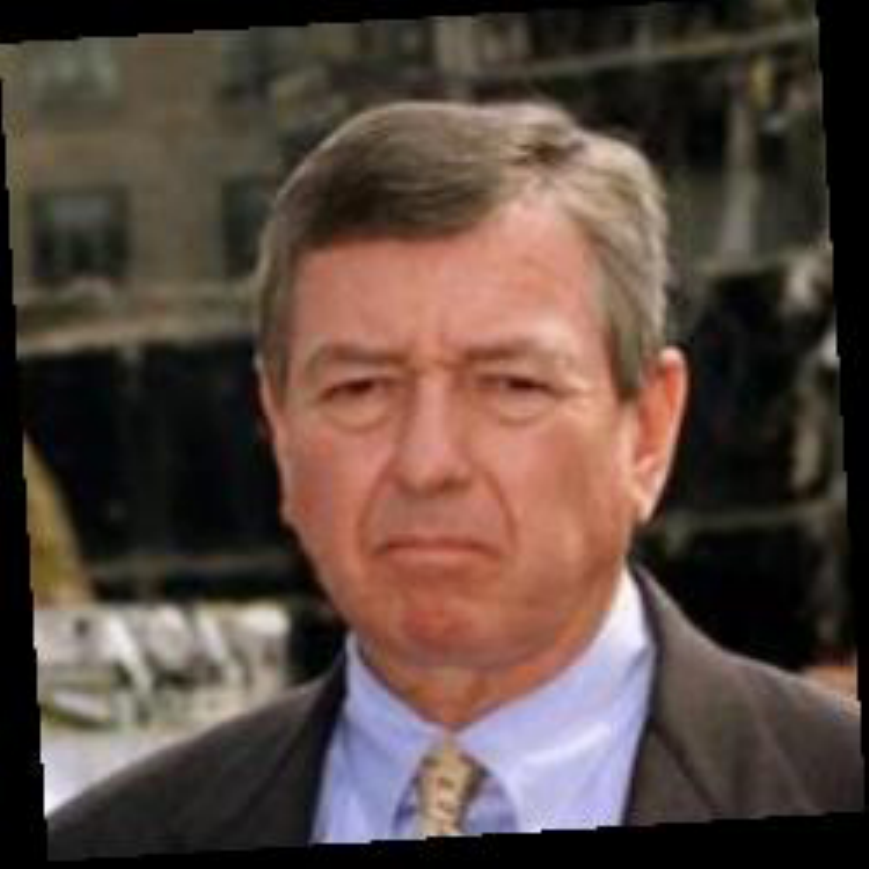}};

%Bottom left triplet
\draw [thick] (-2.65, -2.35) -- (-1.85, -1.90); 
\draw (-2.65,-2.35) node(n1)  {\includegraphics[width=0.5 cm]{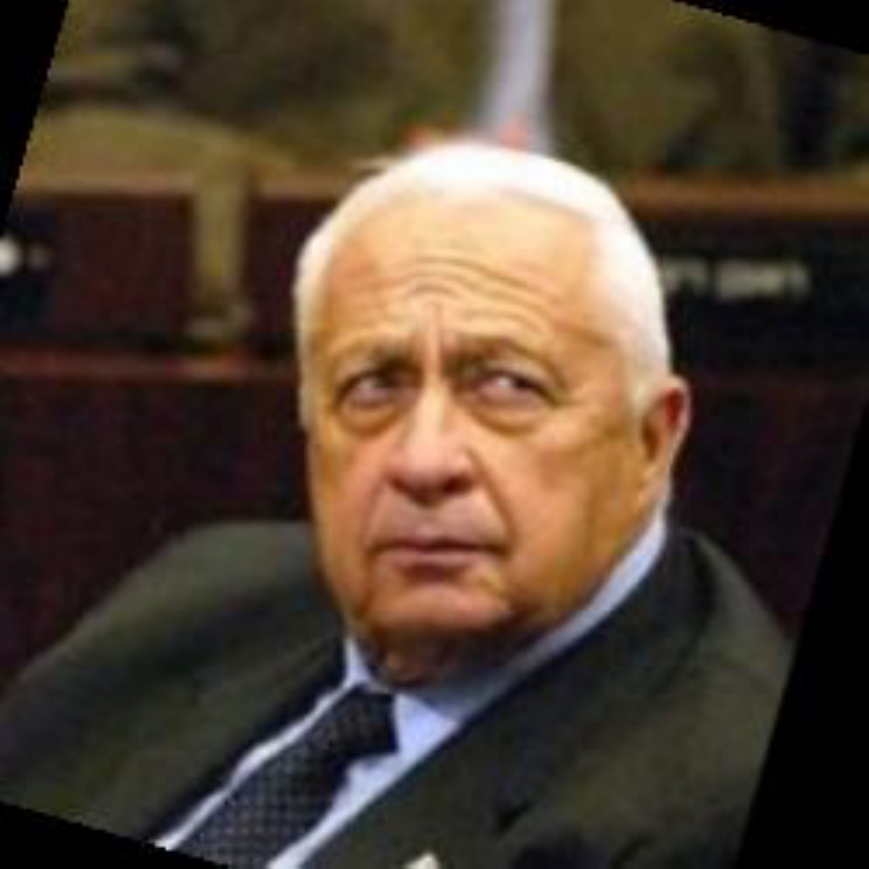}};  
  
\draw [thick] (-2.65, -1.5) -- (-1.65, -1.55);   
\draw (-2.65,-1.5) node(n1)  {\includegraphics[width=0.5 cm]{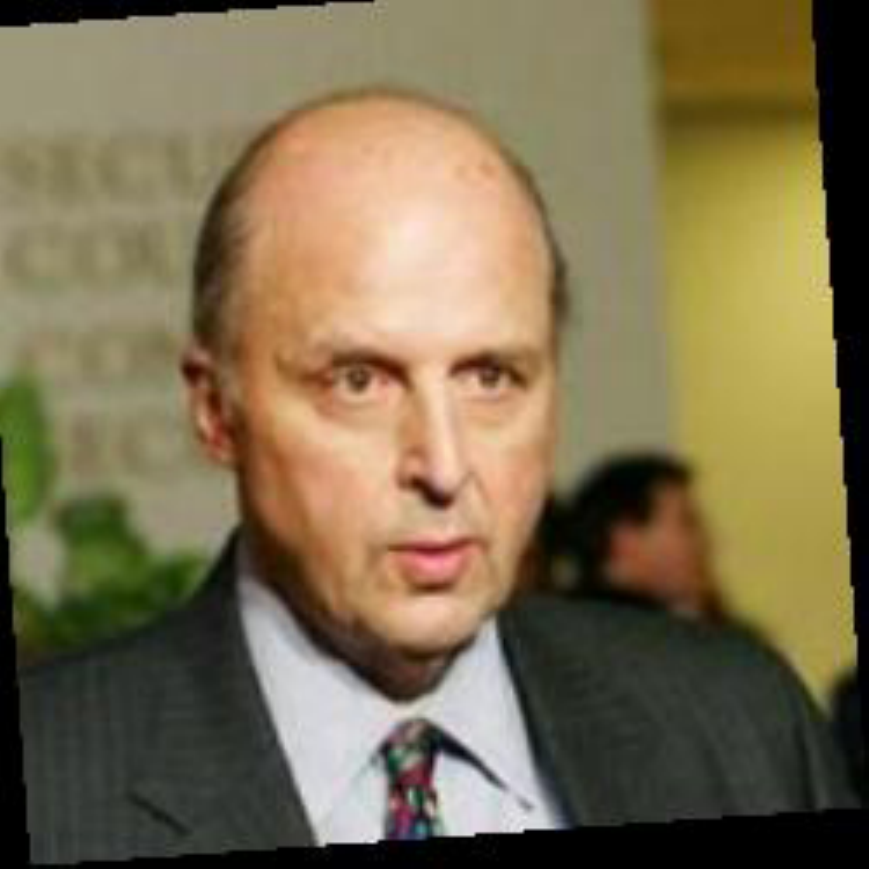}};    

\draw [thick] (-2.75, -0.65) -- (-1.95, -1.20); 
\draw (-2.75,-0.65) node(n1)  {\includegraphics[width=0.5 cm]{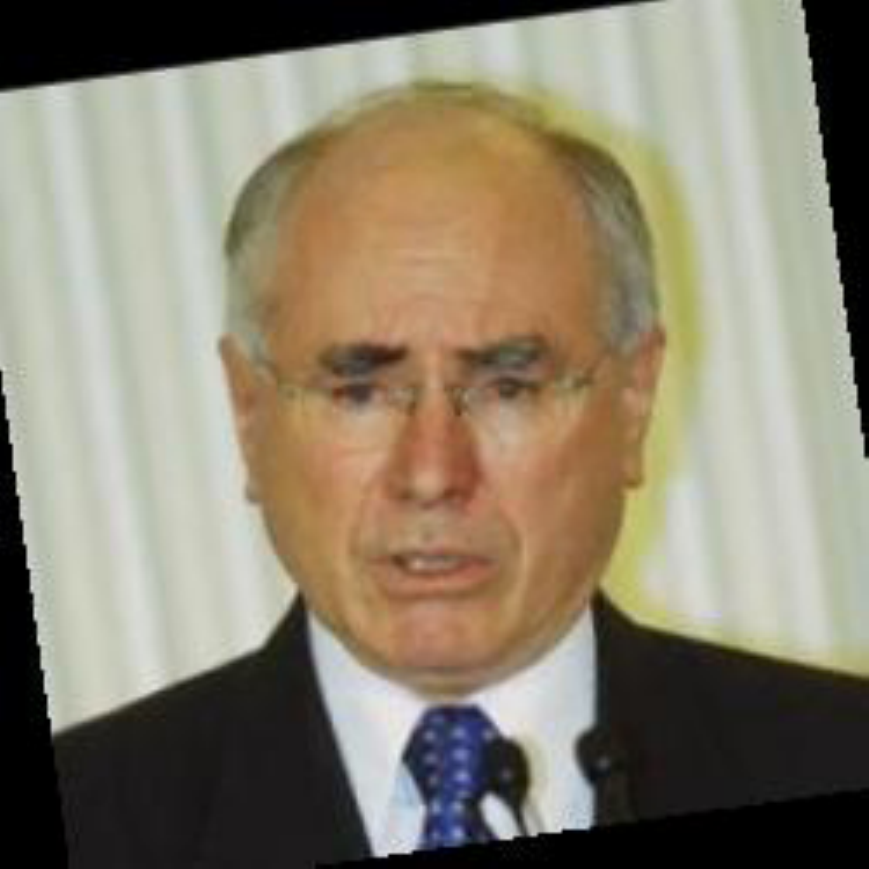}};    

\draw [thick] (-0.25, -2.35) -- (-0.65, -2.00); 
\draw (-0.25,-2.35) node(n1)  {\includegraphics[width=0.5 cm]{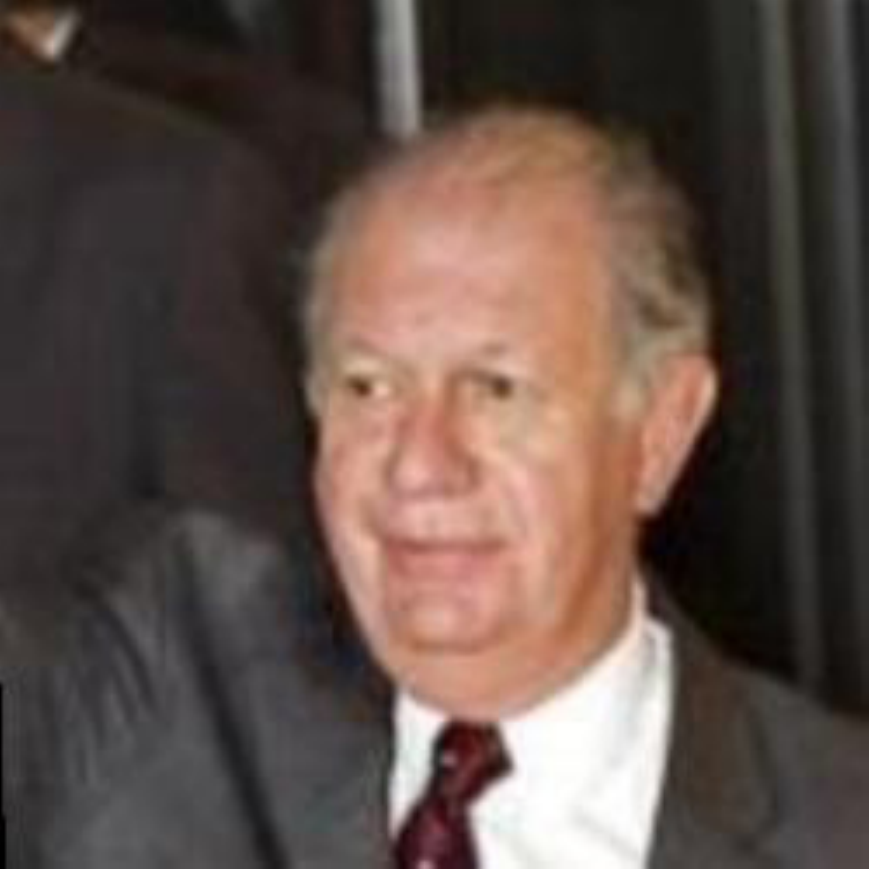}};    

\draw [thick] (1.25, -1.85) -- (0.55, -1.575); 
\draw (1.25,-1.85) node(n1)  {\includegraphics[width=0.5 cm]{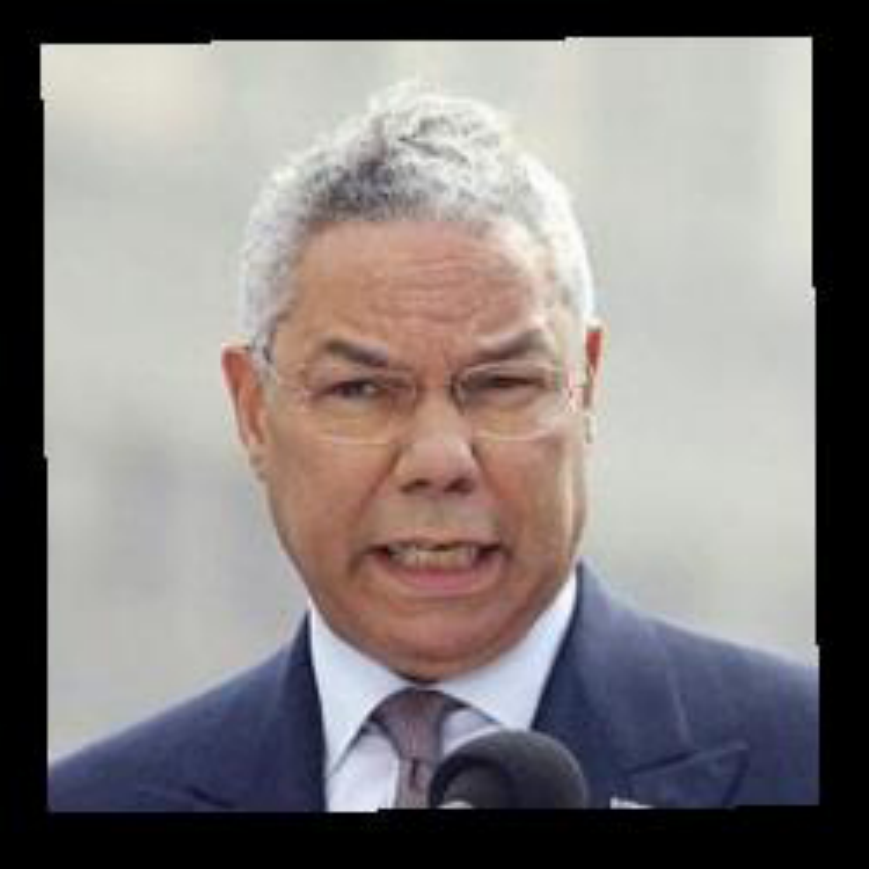}};

%Bottom right corner

\draw [thick] (2.5, -1.75) -- (1.05, -0.375); 
\draw (2.5,-1.75) node(n1)  {\includegraphics[width=0.5 cm]{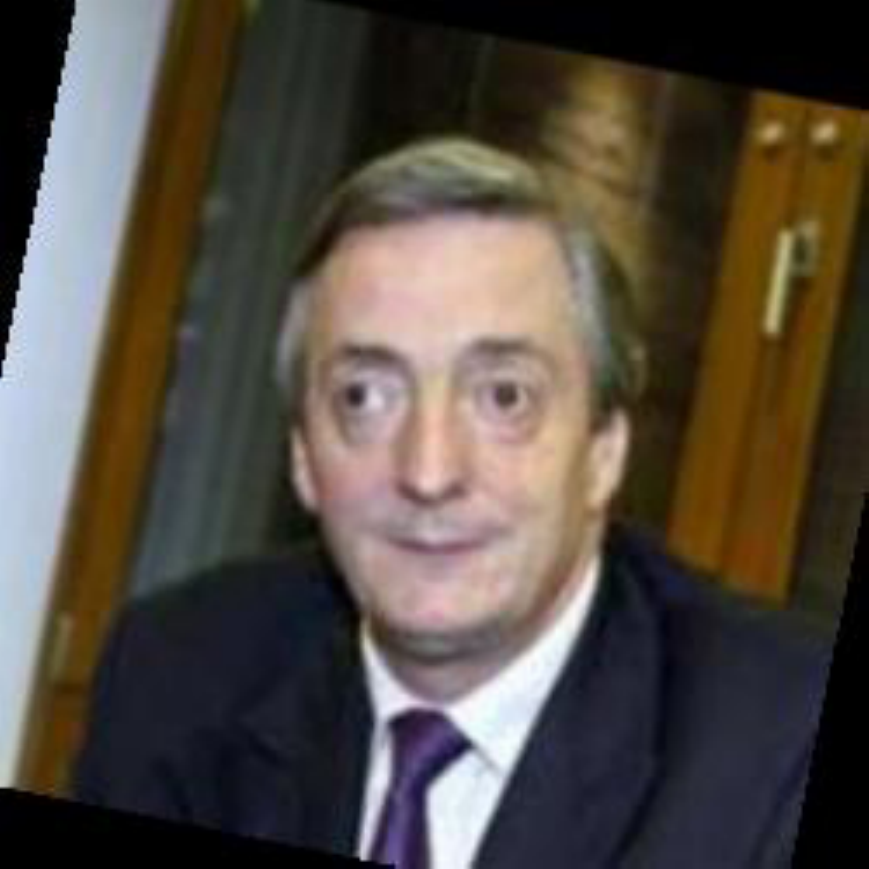}};

\draw [thick] (2.9, -1.2) -- (1.75, -0.275); 
\draw (2.9,-1.2) node(n1)  {\includegraphics[width=0.5 cm]{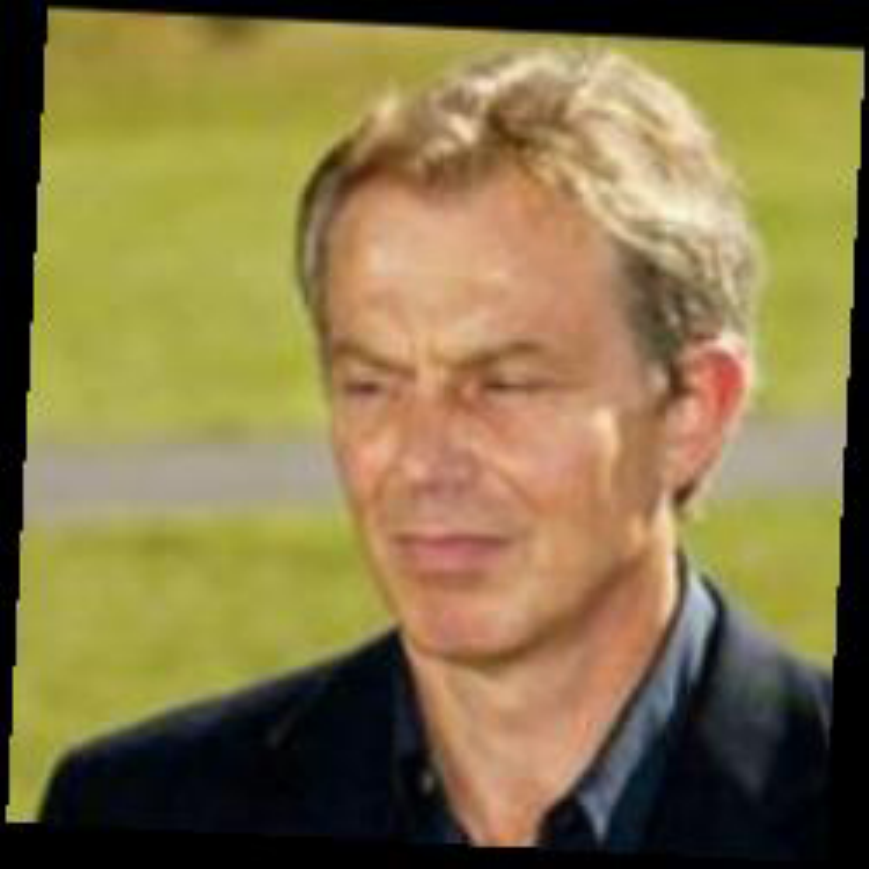}};

\draw [thick] (1.25, -1.2) -- (0.1, -0.275); 
\draw (1.25,-1.2) node(n1)  {\includegraphics[width=0.5 cm]{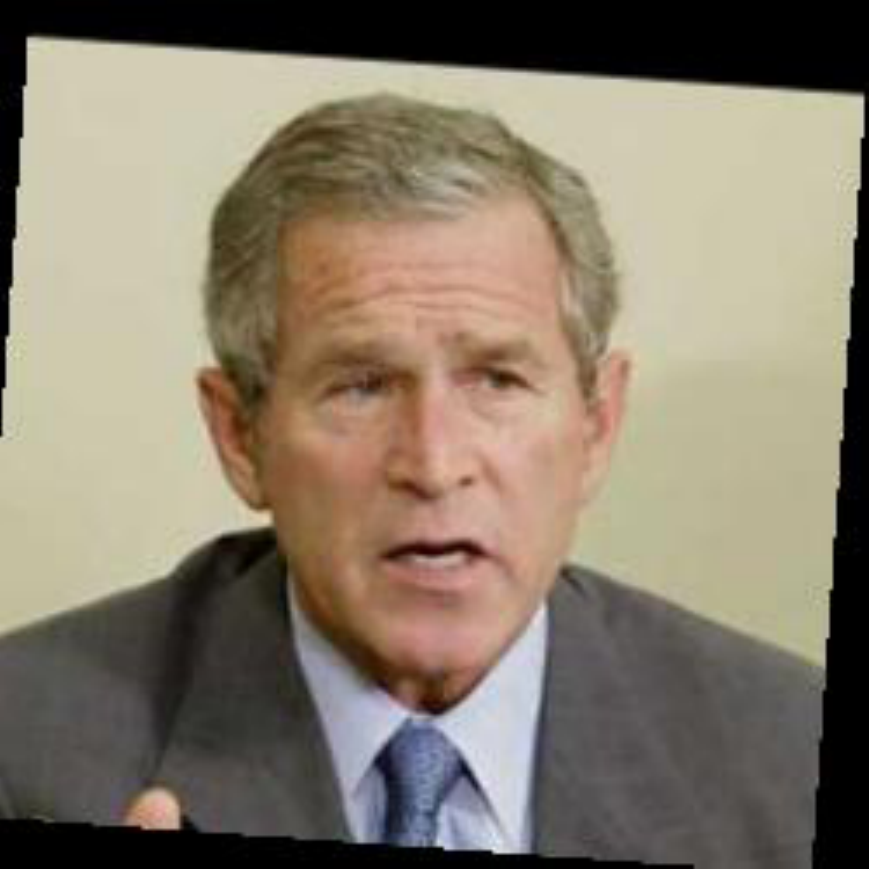}};    

%Top left corner
\draw [thick] (-2.65, 1.85) -- (-1.85, 1.30); 
\draw (-2.65,1.85) node(n1)  {\includegraphics[width=0.5 cm]{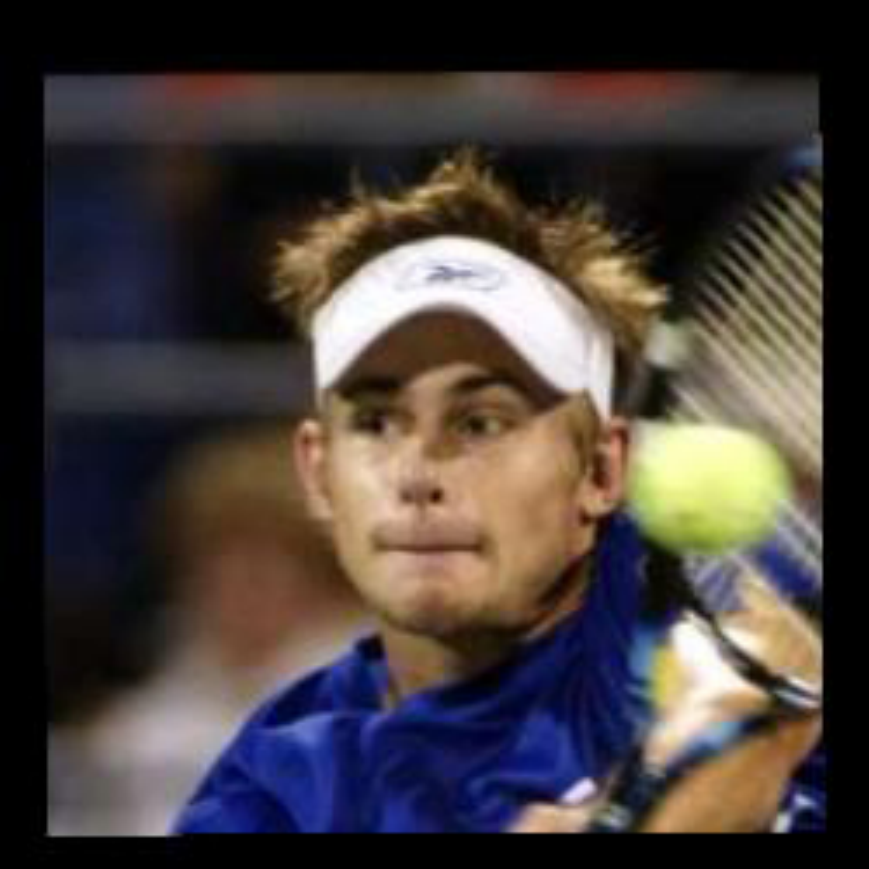}};  

\draw [thick] (-1.85, 2.35) -- (-1.4, 1.60); 
\draw (-1.85,2.35) node(n1)  {\includegraphics[width=0.5 cm]{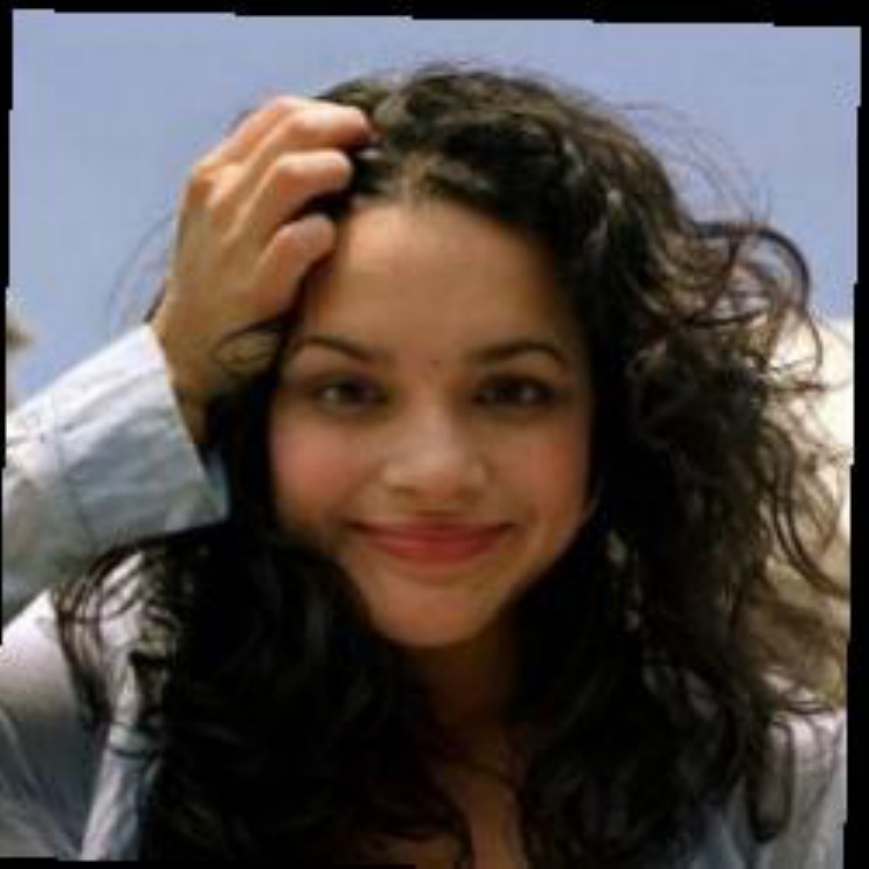}};  

\draw [thick] (-1.25, 2.35) -- (-0.75, 1.80); 
\draw (-1.25,2.35) node(n1)  {\includegraphics[width=0.5 cm]{imgs/4004}};  

\draw [thick] (0.15, 2.5) -- (-0.3, 2.00); 
\draw (0.15,2.5) node(n1)  {\includegraphics[width=0.5 cm]{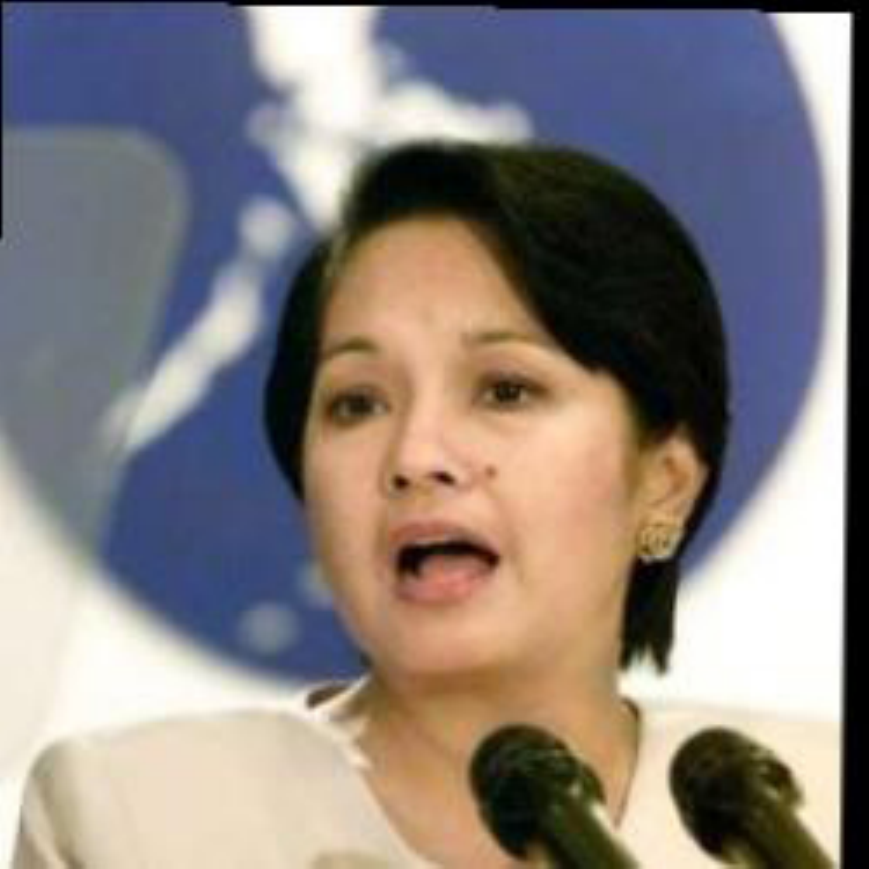}};  

\draw [thick] (-3.25, 1.25) -- (-1.25, 1.15); 
\draw (-3.25,1.25) node(n1)  {\includegraphics[width=0.5 cm]{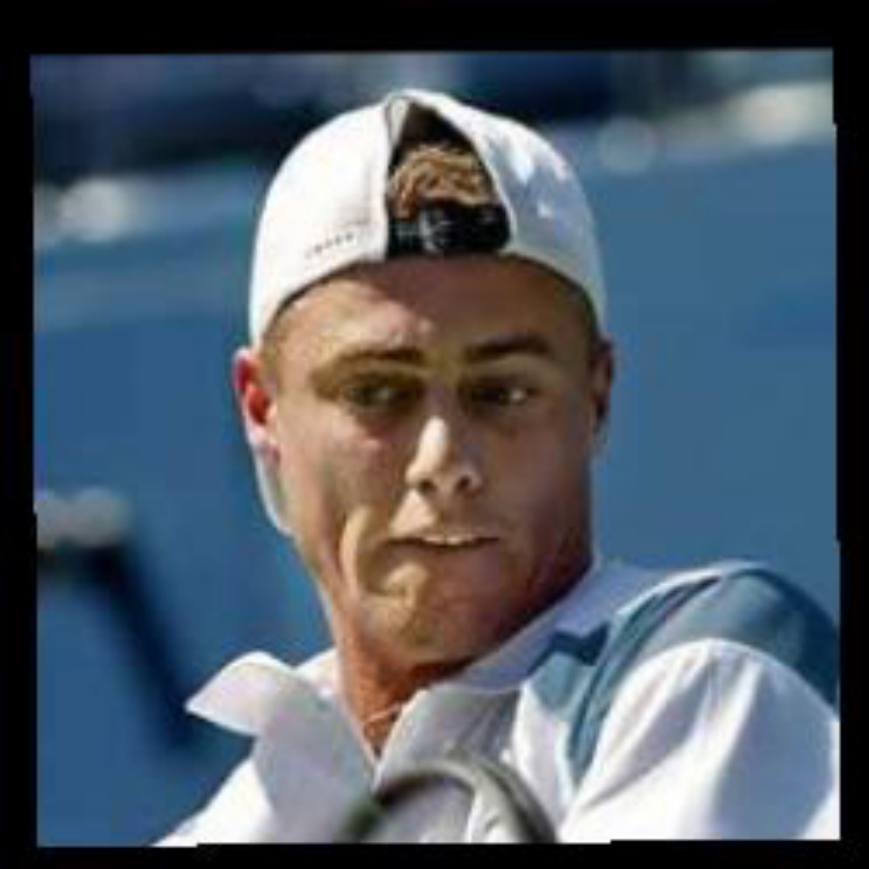}};  

\draw [thick] (-3.25, 0.25) -- (-1.25, 0.95); 
\draw (-3.25,0.25) node(n1)  {\includegraphics[width=0.5 cm]{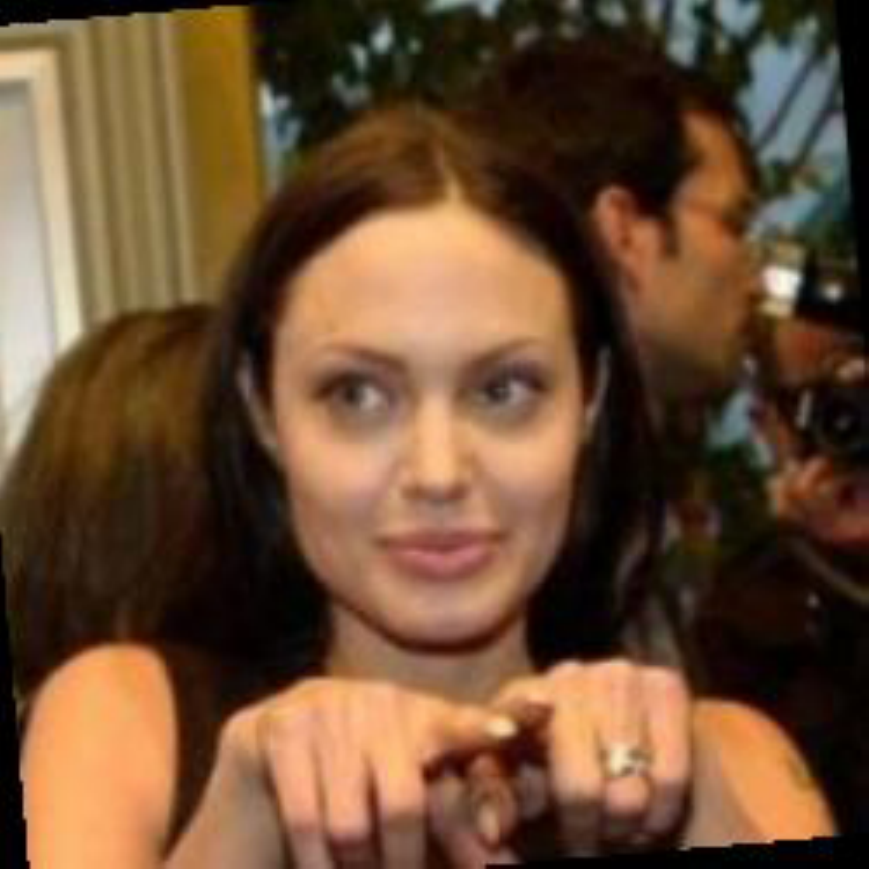}};  

\draw [thick] (0.75, 2.5) -- (-0.65, 1.10); 
\draw (0.75,2.5) node(n1)  {\includegraphics[width=0.5 cm]{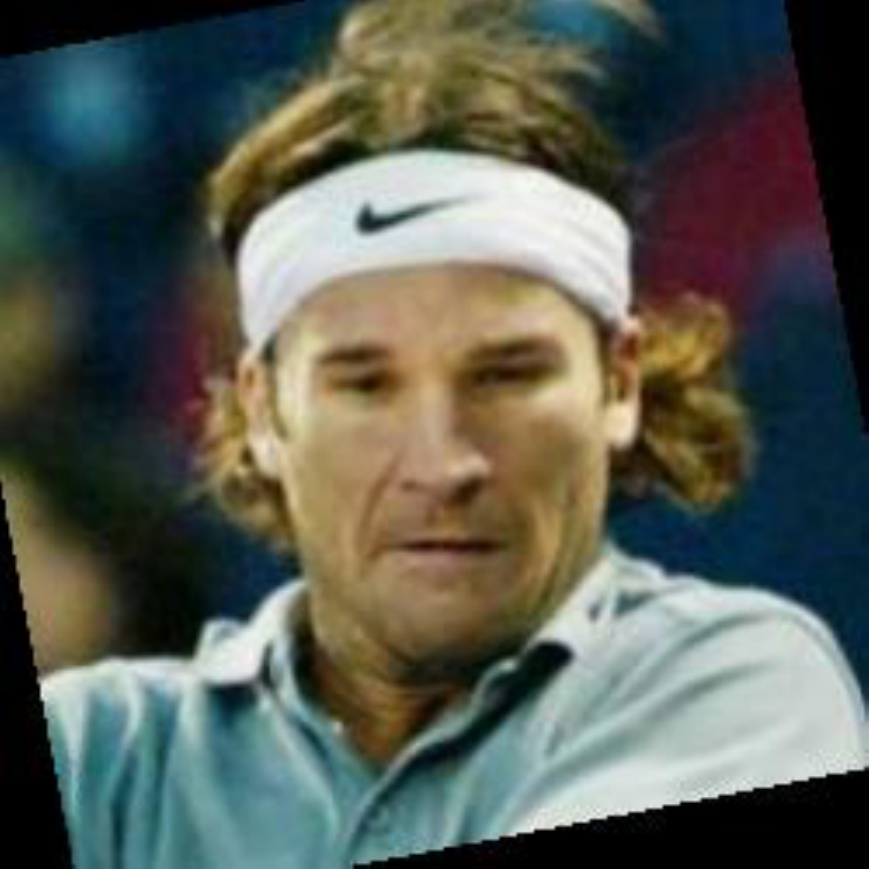}};  

\draw [thick] (-2.75, 2.5) -- (-1.0, 1.30); 
\draw (-2.75,2.5) node(n1)  {\includegraphics[width=0.5 cm]{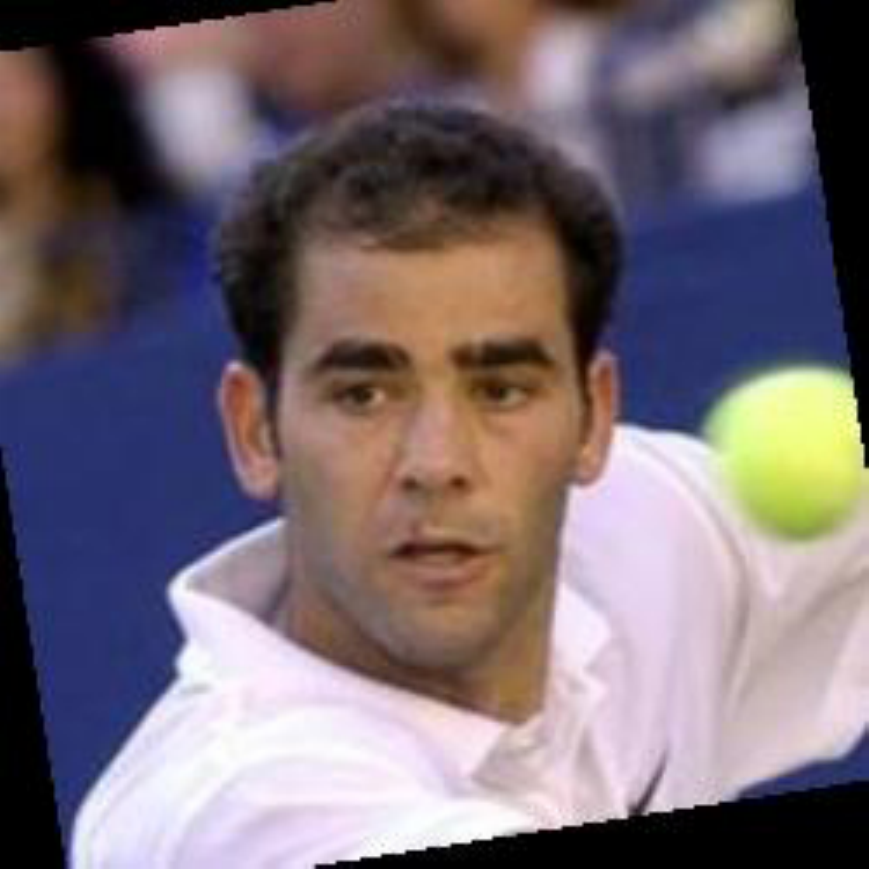}};  

\draw [thick] (1.75, 2.5) -- (-0.83, 0.880); 
\draw (1.75,2.5) node(n1)  {\includegraphics[width=0.5 cm]{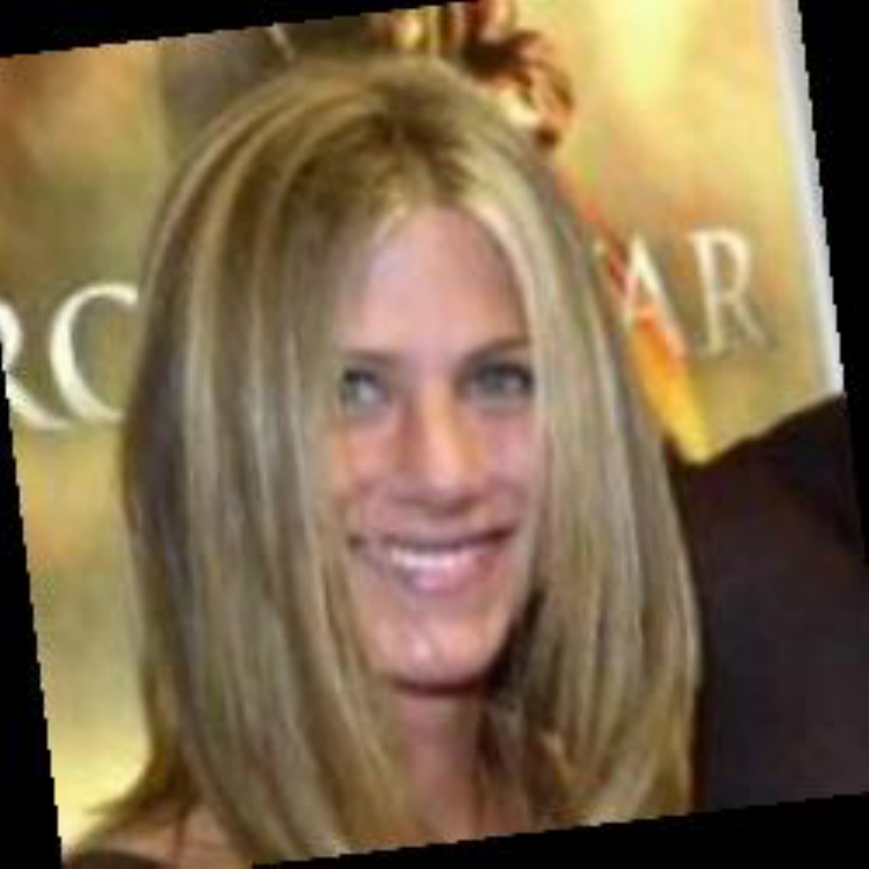}};  

\draw [thick] (1.75, 1.9) -- (-0.33, 0.92); 
\draw (1.75,1.9) node(n1)  {\includegraphics[width=0.5 cm]{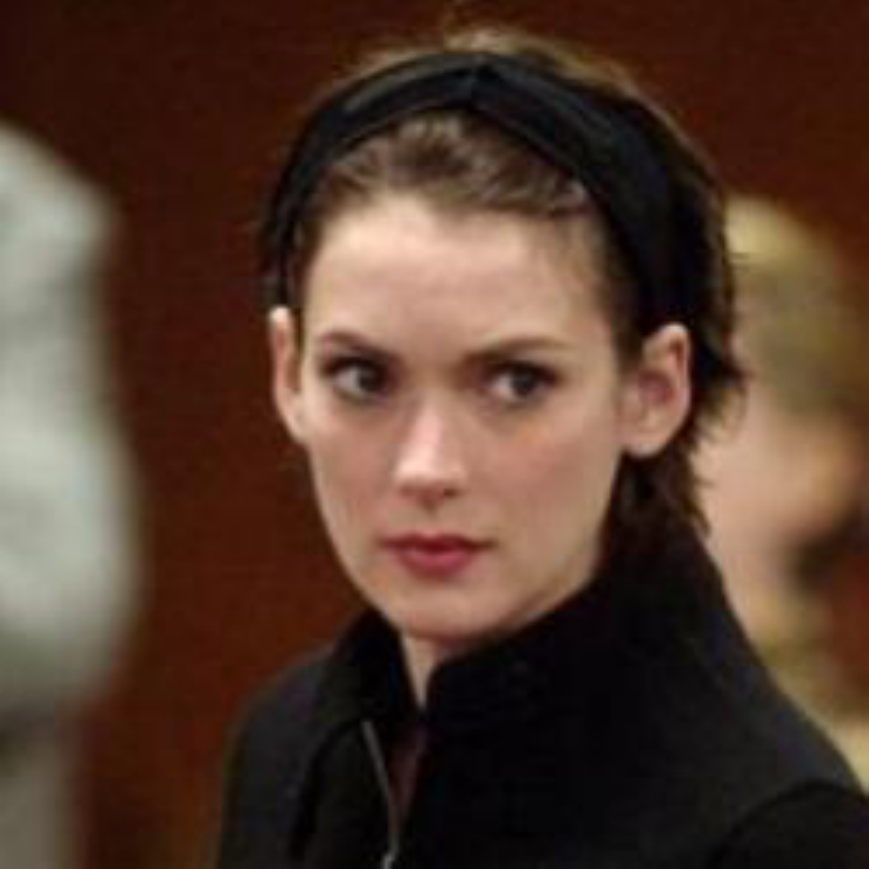}};  

\draw [thick] (-0.5, 2.5) -- (-0.75, 1.50); 
\draw (-0.5,2.5) node(n1)  {\includegraphics[width=0.5 cm]{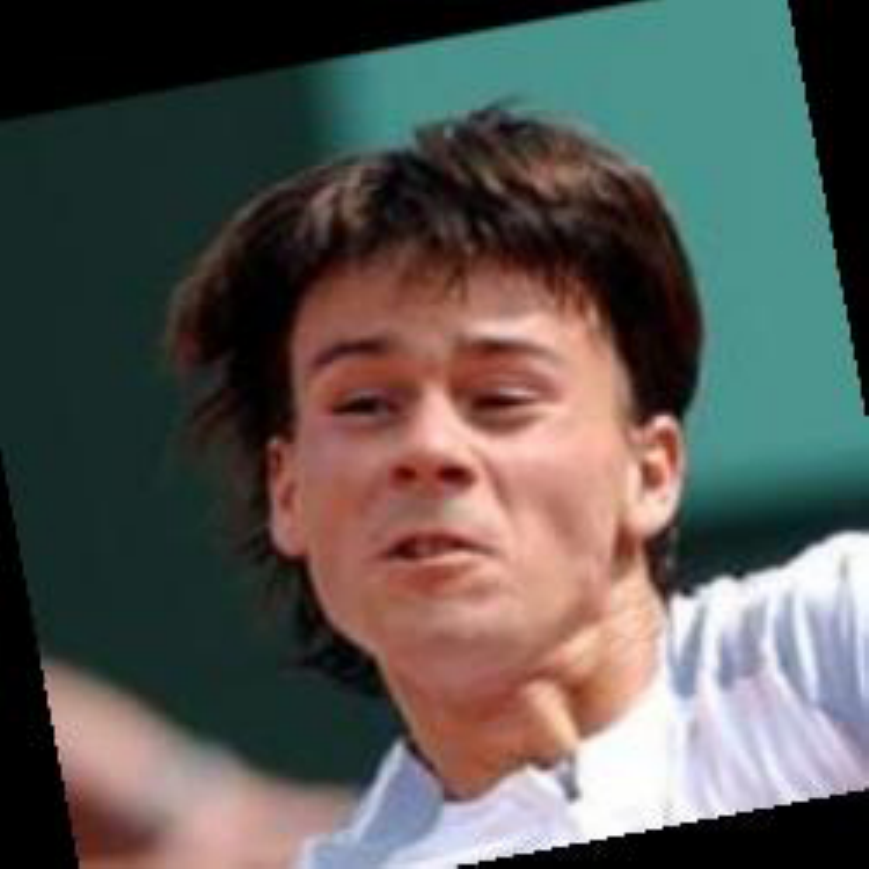}};

%%%%%%%%%%%%%%%%%%%%%%%%%%%%%%%%%%%%%%%%%%%%%%%%
%%	QUADRUPLET
%%%%%%%%%%%%%%%%%%%%%%%%%%%%%%%%%%%%%%%%%%%%%%%%

\def\deltaX{-8.5}
\def\deltaY{-6.5}

\draw (8.5+\deltaX,0+\deltaY) node(n1)  {\includegraphics[width=8.0 cm]{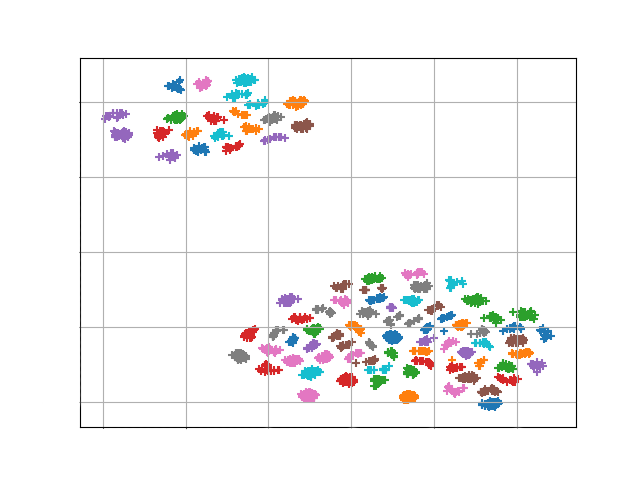}};     
\draw (8.5+\deltaX, -3+\deltaY) node[rectangle] {\small{\textbf{Quadruplet Loss}}};

%Women
\draw [thick] (6.0+\deltaX, 0.5+\deltaY) -- (6.5+\deltaX, 1.3+\deltaY); 
\draw (6.0+\deltaX,0.5+\deltaY) node(n1)  {\includegraphics[width=0.5 cm]{imgs/2507}};    

\draw [thick] (6.75+\deltaX, 2.5+\deltaY) -- (7.05+\deltaX, 1.9+\deltaY); 
\draw (6.75+\deltaX,2.5+\deltaY) node(n1)  {\includegraphics[width=0.5 cm]{imgs/305}};    

\draw [thick] (7.5+\deltaX, 2.5+\deltaY) -- (7.5+\deltaX, 1.55+\deltaY); 
\draw (7.5+\deltaX,2.5+\deltaY) node(n1)  {\includegraphics[width=0.5 cm]{imgs/4117}};    

\draw [thick] (8.3+\deltaX, 2.5+\deltaY) -- (7.7+\deltaX, 1.7+\deltaY); 
\draw (8.3+\deltaX,2.5+\deltaY) node(n1)  {\includegraphics[width=0.5 cm]{imgs/4004}};    

\draw [thick] (8.9+\deltaX, 2.0+\deltaY) -- (8.3+\deltaX, 1.4+\deltaY); 
\draw (8.9+\deltaX,2.0+\deltaY) node(n1)  {\includegraphics[width=0.5 cm]{imgs/1934}};    

\draw [thick] (8.75+\deltaX, 1.0+\deltaY) -- (7.9+\deltaX, 1.5+\deltaY); 
\draw (8.75+\deltaX,1.0+\deltaY) node(n1)  {\includegraphics[width=0.5 cm]{imgs/5660}};

%Men
\draw [thick] (7.25+\deltaX, -2.25+\deltaY) -- (7.85+\deltaX, -1.65+\deltaY); 
\draw (7.25+\deltaX,-2.25+\deltaY) node(n1)  {\includegraphics[width=0.5 cm]{imgs/4573}};    

\draw [thick] (6.25+\deltaX, -1.95+\deltaY -- (7.475+\deltaX, -1.45+\deltaY); 
\draw (6.25+\deltaX,-1.95+\deltaY) node(n1)  {\includegraphics[width=0.5 cm]{imgs/374}};    

\draw [thick] (6.25+\deltaX, -1.05+\deltaY) -- (7.65+\deltaX, -1.2+\deltaY); 
\draw (6.25+\deltaX,-1.05+\deltaY) node(n1)  {\includegraphics[width=0.5 cm]{imgs/2747}};    

\draw [thick] (7.25+\deltaX, -0.15+\deltaY) -- (7.95+\deltaX, -1.2+\deltaY); 
\draw (7.25+\deltaX,-0.15+\deltaY) node(n1)  {\includegraphics[width=0.5 cm]{imgs/2722}};    

\draw [thick] (9.25+\deltaX, -2.5+\deltaY) -- (9.2+\deltaX, -1.775+\deltaY); 
\draw (9.25+\deltaX,-2.5+\deltaY) node(n1)  {\includegraphics[width=0.5 cm]{imgs/4058}};    

\draw [thick] (8.25+\deltaX, -2.4+\deltaY) -- (8.375+\deltaX, -1.95+\deltaY); 
\draw (8.25+\deltaX,-2.4+\deltaY) node(n1)  {\includegraphics[width=0.5 cm]{imgs/1048}};    

\draw [thick] (9.95+\deltaX, -2.5+\deltaY) -- (9.6+\deltaX, -1.95+\deltaY); 
\draw (9.95+\deltaX,-2.5+\deltaY) node(n1)  {\includegraphics[width=0.5 cm]{imgs/1872}};

\draw [thick] (11.35+\deltaX, -2.5+\deltaY) -- (10.65+\deltaX, -2.05+\deltaY); 
\draw (11.35+\deltaX,-2.5+\deltaY) node(n1)  {\includegraphics[width=0.5 cm]{imgs/5459}};    

\draw [thick] (11.9+\deltaX, -2.25+\deltaY) -- (10.6+\deltaX, -1.85+\deltaY); 
\draw (11.9+\deltaX,-2.25) node(n1)  {\includegraphics[width=0.5 cm]{imgs/2942}};    

\draw [thick] (11.9+\deltaX, -1.65+\deltaY) -- (10.9+\deltaX, -1.75+\deltaY); 
\draw (11.9+\deltaX,-1.65+\deltaY) node(n1)  {\includegraphics[width=0.5 cm]{imgs/21}};    

\draw [thick] (10.65+\deltaX, -2.5+\deltaY) -- (10.35+\deltaX, -1.7+\deltaY); 
\draw (10.65+\deltaX,-2.5+\deltaY) node(n1)  {\includegraphics[width=0.5 cm]{imgs/2683}}; 

\draw [thick] (11.9+\deltaX, -0.95+\deltaY) -- (10.8+\deltaX, -1.575+\deltaY); 
\draw (11.9+\deltaX,-0.95+\deltaY) node(n1)  {\includegraphics[width=0.5 cm]{imgs/4774}};     

\draw [thick] (9.2+\deltaX, 0.05+\deltaY) -- (9.175+\deltaX, -0.5+\deltaY); 
\draw (9.2+\deltaX,0.05+\deltaY) node(n1)  {\includegraphics[width=0.5 cm]{imgs/3352}};     

\draw [thick] (9.9+\deltaX, 0.05+\deltaY) -- (9.625+\deltaX, -0.45+\deltaY); 
\draw (9.9+\deltaX,0.05+\deltaY) node(n1)  {\includegraphics[width=0.5 cm]{imgs/4334}};     

\draw [thick] (8.5+\deltaX, 0.05+\deltaY) -- (9.2+\deltaX, -0.625+\deltaY); 
\draw (8.5+\deltaX,0.05+\deltaY) node(n1)  {\includegraphics[width=0.5 cm]{imgs/292}};     

\draw [thick] (10.75+\deltaX, 0.05+\deltaY) -- (9.7+\deltaX, -1.0+\deltaY); 
\draw (10.75+\deltaX,0.05) node(n1)  {\includegraphics[width=0.5 cm]{imgs/1994}};     

\draw [thick] (11.5+\deltaX, 0.05+\deltaY) -- (9.9+\deltaX, -0.875+\deltaY); 
\draw (11.5+\deltaX,0.05+\deltaY) node(n1)  {\includegraphics[width=0.5 cm]{imgs/789}};

\end{tikzpicture}
    \caption{Comparison between the embeddings resulting from the triplet loss~\cite{Schroff2015} (top plot), and from the proposed quadruplet loss (bottom plot). Results are given for $t=2$ features \{''ID'', ''Gender''\} for the LFW identities with at least 15 images (89 elements).}
        \label{fig:comparison_triplet_quadruplet}
    \end{center}
\end{figure}

As an example, Fig.~\ref{fig:comparison_triplet_quadruplet} compares the bidimensional embeddings resulting from the triplet and from the quadruplet losses, in the subset of the LFW identities with more than 15 images in the dataset (using  $t=2:\{\text{''ID''}, \text{''Gender''}\}$ labels). This plot yielded from the projection of a 128-dimensional embedding down to two dimensions, according to the Neighbourhood Component Analysis (NCA)~\cite{Goldberger2005} algorithm.  

It can be seen that the triplet loss provided an embedding where the positions of elements are exclusively determined by their appearance, with pairs of IDS with fully disjoint labels appearing close to each other (this is evident in the upper left corner of the triplet embedding, where ''\emph{female}'' elements appear adjacent to ''\emph{male tennis players}''). In opposition, using the quadruplet loss we obtain a large  separation between the elements of different genders, while keeping the compactness per identity.  This kind of embedding will be interesting in  at least two cases: 1) in identity retrieval, to guarantee that all retrieved elements have the same soft labels of the query; and 2) upon a semantic description of the query (e.g., ''\emph{find adult white males similar to this image}''),  to guarantee that all  retrieved elements  meet the semantic criteria. A third application comprises the direct inference of all the fine + coarse (soft) labels in a simple \emph{k-neighbours} fashion.

\section{Results and Discussion}
 \label{sec:Results}

\subsection{Experimental Setting and Preprocessing}
\label{ssec:Datasets}

Our empirical validation was conducted in one proprietary (BIODI) and four freely available datasets (LFW, PETA, IJB-A and Megaface) well known in the biometrics and re-identification literature. 

The BIODI\footnote{\url{http://di.ubi.pt/~hugomcp/BIODI/}} dataset is proprietary of \emph{Tomiworld}$^{\scriptsize{\textregistered}}$\footnote{\url{https://tomiworld.com/}}, being composed of 849,932 images from 13,876 subjects, taken from 216 indoor/outdoor video surveillance sequences.  All images were manually annotated for 14 labels: gender, age, height, body volume, ethnicity, hair color and style, beard, moustache, glasses and clothing (x4).  The Labeled Faces in the Wild (LFW)~\cite{Miller2016} dataset contains 13,233 images from 5,749 identities, collected from the web, with large variations in pose, expression and lighting conditions. PETA~\cite{Deng2014} is a combination of 10 pedestrian re-identification datasets, composed of 19,000 images from 8,705 subjects, each one annotated with 61 binary and 4 multi-output atributes. The IIJB-A~\cite{Klare2015} dataset contains 5,397 images plus 20,412 video frames from 500 individuals, with large variations in pose and illumination. Finally,  the Megaface~\cite{Kemelmacher2016} set was released to evaluate face recognition performance at the million scale, and consists of a gallery set and a probe set. The gallery set is a subset of Flickr photos from Yahoo (more than 1,000,000 images from 690,000 subjects). The probe dataset includes FaceScrub and FGNet sets. FaceScrub has 100,000 images from 530  individuals and  FGNet contains 1,002 images of 82 identities.

\subsection{Convolutional Neural Networks}

\begin{figure}[ht!]
\begin{center}
\begin{tikzpicture}

\def\posX{-1}
\def\posY{0}
\fill [rounded corners, gray] (\posX-1, \posY-0.15) rectangle (\posX+1, \posY+0.15);    
\draw  [white] (\posX, \posY) node {\scriptsize{VGG-like}};    

\def\posX{-1}
\def\deltaY{-0.75}
\def\posY{-1}

\def\pos{0}
\fill [rounded corners=3pt, yellow!50] (\posX-1, \posY-0.15+\pos*\deltaY) rectangle (\posX+1, \posY+0.15+\pos*\deltaY);    
\draw  [black] (\posX, \posY-\pos*\deltaY) node {\scriptsize{3 $\times$ 3, 64}};    
\draw [->] (\posX, \posY-0.2-\pos*\deltaY) -- (\posX, \posY-0.5-\pos*\deltaY);

\def\pos{1}
\fill [rounded corners=3pt, yellow!50] (\posX-1, \posY-0.15+\pos*\deltaY) rectangle (\posX+1, \posY+0.15+\pos*\deltaY);    
\draw  [black] (\posX, \posY+\pos*\deltaY) node {\scriptsize{3 $\times$ 3, 64}};    
\draw [->] (\posX, \posY-0.2+\pos*\deltaY) -- (\posX, \posY-0.5+\pos*\deltaY);

\def\pos{2}
\fill [rounded corners=3pt, blue!50] (\posX-1, \posY-0.15+\pos*\deltaY) rectangle (\posX+1, \posY+0.15+\pos*\deltaY);    
\draw  [black] (\posX, \posY+\pos*\deltaY) node {\scriptsize{max, 2 $\times$ 2}};    
\draw [->] (\posX, \posY-0.2+\pos*\deltaY) -- (\posX, \posY-0.5+\pos*\deltaY);

\def\pos{3}
\fill [rounded corners=3pt, green!50] (\posX-1, \posY-0.15+\pos*\deltaY) rectangle (\posX+1, \posY+0.15+\pos*\deltaY);    
\draw  [black] (\posX, \posY+\pos*\deltaY) node {\scriptsize{dropout, 0.75}};    
\draw [->] (\posX, \posY-0.2+\pos*\deltaY) -- (\posX, \posY-0.5+\pos*\deltaY);

\def\pos{4}
\fill [rounded corners=3pt, yellow!50] (\posX-1, \posY-0.15+\pos*\deltaY) rectangle (\posX+1, \posY+0.15+\pos*\deltaY);    
\draw  [black] (\posX, \posY+\pos*\deltaY) node {\scriptsize{3 $\times$ 3, 128}};    
\draw [->] (\posX, \posY-0.2+\pos*\deltaY) -- (\posX, \posY-0.5+\pos*\deltaY);

\def\pos{5}
\fill [rounded corners=3pt, yellow!50] (\posX-1, \posY-0.15+\pos*\deltaY) rectangle (\posX+1, \posY+0.15+\pos*\deltaY);    
\draw  [black] (\posX, \posY+\pos*\deltaY) node {\scriptsize{3 $\times$ 3, 128}};    
\draw [->] (\posX, \posY-0.2+\pos*\deltaY) -- (\posX, \posY-0.5+\pos*\deltaY);

\def\pos{6}
\fill [rounded corners=3pt, blue!50] (\posX-1, \posY-0.15+\pos*\deltaY) rectangle (\posX+1, \posY+0.15+\pos*\deltaY);    
\draw  [black] (\posX, \posY+\pos*\deltaY) node {\scriptsize{max, 2 $\times$ 2}};    
\draw [->] (\posX, \posY-0.2+\pos*\deltaY) -- (\posX, \posY-0.5+\pos*\deltaY);

\def\pos{7}
\fill [rounded corners=3pt, yellow!50] (\posX-1, \posY-0.15+\pos*\deltaY) rectangle (\posX+1, \posY+0.15+\pos*\deltaY);    
\draw  [black] (\posX, \posY+\pos*\deltaY) node {\scriptsize{3 $\times$ 3, 256}};    
\draw [->] (\posX, \posY-0.2+\pos*\deltaY) -- (\posX, \posY-0.5+\pos*\deltaY);

\def\pos{8}
\fill [rounded corners=3pt, yellow!50] (\posX-1, \posY-0.15+\pos*\deltaY) rectangle (\posX+1, \posY+0.15+\pos*\deltaY);    
\draw  [black] (\posX, \posY+\pos*\deltaY) node {\scriptsize{3 $\times$ 3, 256}};    
\draw [->] (\posX, \posY-0.2+\pos*\deltaY) -- (\posX, \posY-0.5+\pos*\deltaY);

\def\pos{9}
\fill [rounded corners=3pt, yellow!50] (\posX-1, \posY-0.15+\pos*\deltaY) rectangle (\posX+1, \posY+0.15+\pos*\deltaY);    
\draw  [black] (\posX, \posY+\pos*\deltaY) node {\scriptsize{3 $\times$ 3, 256}};    
\draw [->] (\posX, \posY-0.2+\pos*\deltaY) -- (\posX, \posY-0.5+\pos*\deltaY);

\def\pos{10}
\fill [rounded corners=3pt, yellow!50] (\posX-1, \posY-0.15+\pos*\deltaY) rectangle (\posX+1, \posY+0.15+\pos*\deltaY);    
\draw  [black] (\posX, \posY+\pos*\deltaY) node {\scriptsize{3 $\times$ 3, 256}};    
\draw [->] (\posX, \posY-0.2+\pos*\deltaY) -- (\posX, \posY-0.5+\pos*\deltaY);

\def\pos{11}
\fill [rounded corners=3pt, blue!50] (\posX-1, \posY-0.15+\pos*\deltaY) rectangle (\posX+1, \posY+0.15+\pos*\deltaY);    
\draw  [black] (\posX, \posY+\pos*\deltaY) node {\scriptsize{max, 2 $\times$ 2}};    
\draw [->] (\posX, \posY-0.2+\pos*\deltaY) -- (\posX, \posY-0.5+\pos*\deltaY);

\def\pos{12}
\fill [rounded corners=3pt, green!50] (\posX-1, \posY-0.15+\pos*\deltaY) rectangle (\posX+1, \posY+0.15+\pos*\deltaY);    
\draw  [black] (\posX, \posY+\pos*\deltaY) node {\scriptsize{dropout, 0.75}};    
\draw [->] (\posX, \posY-0.2+\pos*\deltaY) -- (\posX, \posY-0.5+\pos*\deltaY);

\def\pos{13}
\fill [rounded corners=3pt, yellow!50] (\posX-1, \posY-0.15+\pos*\deltaY) rectangle (\posX+1, \posY+0.15+\pos*\deltaY);    
\draw  [black] (\posX, \posY+\pos*\deltaY) node {\scriptsize{3 $\times$ 3, 256}};    
\draw [->] (\posX, \posY-0.2+\pos*\deltaY) -- (\posX, \posY-0.5+\pos*\deltaY);

\def\pos{14}
\fill [rounded corners=3pt, yellow!50] (\posX-1, \posY-0.15+\pos*\deltaY) rectangle (\posX+1, \posY+0.15+\pos*\deltaY);    
\draw  [black] (\posX, \posY+\pos*\deltaY) node {\scriptsize{3 $\times$ 3, 256}};    
\draw [->] (\posX, \posY-0.2+\pos*\deltaY) -- (\posX, \posY-0.5+\pos*\deltaY);

\def\pos{15}
\fill [rounded corners=3pt, yellow!50] (\posX-1, \posY-0.15+\pos*\deltaY) rectangle (\posX+1, \posY+0.15+\pos*\deltaY);    
\draw  [black] (\posX, \posY+\pos*\deltaY) node {\scriptsize{3 $\times$ 3, 256}};    
\draw [->] (\posX, \posY-0.2+\pos*\deltaY) -- (\posX, \posY-0.5+\pos*\deltaY);

\def\pos{16}
\fill [rounded corners=3pt, yellow!50] (\posX-1, \posY-0.15+\pos*\deltaY) rectangle (\posX+1, \posY+0.15+\pos*\deltaY);    
\draw  [black] (\posX, \posY+\pos*\deltaY) node {\scriptsize{3 $\times$ 3, 256}};    
\draw [->] (\posX, \posY-0.2+\pos*\deltaY) -- (\posX, \posY-0.5+\pos*\deltaY);

\def\pos{17}
\fill [rounded corners=3pt, blue!50] (\posX-1, \posY-0.15+\pos*\deltaY) rectangle (\posX+1, \posY+0.15+\pos*\deltaY);    
\draw  [black] (\posX, \posY+\pos*\deltaY) node {\scriptsize{max, 2 $\times$ 2}};    
\draw [->] (\posX, \posY-0.2+\pos*\deltaY) -- (\posX, \posY-0.5+\pos*\deltaY);

\def\pos{18}
\fill [rounded corners=3pt, green!50] (\posX-1, \posY-0.15+\pos*\deltaY) rectangle (\posX+1, \posY+0.15+\pos*\deltaY);    
\draw  [black] (\posX, \posY+\pos*\deltaY) node {\scriptsize{dropout, 0.75}};    
\draw [->] (\posX, \posY-0.2+\pos*\deltaY) -- (\posX, \posY-0.5+\pos*\deltaY);

\def\pos{19}
\fill [rounded corners=3pt, red!50] (\posX-1, \posY-0.15+\pos*\deltaY) rectangle (\posX+1, \posY+0.15+\pos*\deltaY);    
\draw  [black] (\posX, \posY+\pos*\deltaY) node {\scriptsize{4,096}};    
\draw [->] (\posX, \posY-0.2+\pos*\deltaY) -- (\posX, \posY-0.5+\pos*\deltaY);

\def\pos{20}
\fill [rounded corners=3pt, red!50] (\posX-1, \posY-0.15+\pos*\deltaY) rectangle (\posX+1, \posY+0.15+\pos*\deltaY);    
\draw  [black] (\posX, \posY+\pos*\deltaY) node {\scriptsize{4,096}};    
\draw [->] (\posX, \posY-0.2+\pos*\deltaY) -- (\posX, \posY-0.5+\pos*\deltaY);

\def\pos{21}
\fill [rounded corners=3pt, red!50] (\posX-1, \posY-0.15+\pos*\deltaY) rectangle (\posX+1, \posY+0.15+\pos*\deltaY);    
\draw  [black] (\posX, \posY+\pos*\deltaY) node {\scriptsize{$d$}};

%%%%%%%%%%%%%%%%%%%%%%%%%%%%%%%%%%%%%%%%%%%%%%%%%

\def\posX{3}
\def\posY{0}
\fill [rounded corners, gray] (\posX-1, \posY-0.15) rectangle (\posX+1, \posY+0.15);    
\draw  [white] (\posX, \posY) node {\scriptsize{ResNet-like}};    

\def\posX{3}
\def\deltaY{-0.75}
\def\posY{-1}

\def\pos{0}
\fill [rounded corners=3pt, yellow!50] (\posX-1, \posY-0.15+\pos*\deltaY) rectangle (\posX+1, \posY+0.15+\pos*\deltaY);    
\draw  [black] (\posX, \posY-\pos*\deltaY) node {\scriptsize{7 $\times$ 7, 64, /2}};    
\draw [->] (\posX, \posY-0.2-\pos*\deltaY) -- (\posX, \posY-0.5-\pos*\deltaY);

\def\pos{1}
\fill [rounded corners=3pt, blue!50] (\posX-1, \posY-0.15+\pos*\deltaY) rectangle (\posX+1, \posY+0.15+\pos*\deltaY);    
\draw  [black] (\posX, \posY+\pos*\deltaY) node {\scriptsize{max, 2 $\times$ 2}};    
\draw [->] (\posX, \posY-0.2+\pos*\deltaY) -- (\posX, \posY-0.5+\pos*\deltaY);

\def\pos{2}
\fill [rounded corners=3pt, yellow!50] (\posX-1, \posY-0.15+\pos*\deltaY) rectangle (\posX+1, \posY+0.15+\pos*\deltaY);    
\draw  [black] (\posX, \posY+\pos*\deltaY) node {\scriptsize{3 $\times$ 3, 64}};    
\draw [->] (\posX, \posY-0.2+\pos*\deltaY) -- (\posX, \posY-0.5+\pos*\deltaY);

%rectangle
\draw [rounded corners, black, dashed, thick] (\posX-1.75,\posY+0.35+\pos*\deltaY) rectangle (\posX+2, \posY-1.15+\pos*\deltaY);    
\draw (\posX+1.75, \posY+0.55+\pos*\deltaY) node {\textbf{3$\times$}};

\def\pos{3}
\fill [rounded corners=3pt, yellow!50] (\posX-1, \posY-0.15+\pos*\deltaY) rectangle (\posX+1, \posY+0.15+\pos*\deltaY);    
\draw  [black] (\posX, \posY+\pos*\deltaY) node {\scriptsize{3 $\times$ 3, 64}};    
\draw [->] (\posX, \posY-0.2+\pos*\deltaY) -- (\posX, \posY-0.5+\pos*\deltaY);

%skip
\draw [->] (\posX, {\posY-0.2+(\pos-2.1)*\deltaY-0.2}) .. controls (\posX+2.25, {\posY-0.2+(\pos-1.5)*\deltaY})  and (\posX+2.25, {\posY-0.2+(\pos-0.0)*\deltaY})  ..  (\posX, \posY-0.2+\pos*\deltaY-0.2);

\def\pos{4}
\fill [rounded corners=3pt, yellow!50] (\posX-1, \posY-0.15+\pos*\deltaY) rectangle (\posX+1, \posY+0.15+\pos*\deltaY);    
\draw  [black] (\posX, \posY+\pos*\deltaY) node {\scriptsize{3 $\times$ 3, 128, /2}};    
\draw [->] (\posX, \posY-0.2+\pos*\deltaY) -- (\posX, \posY-0.5+\pos*\deltaY);

\def\pos{5}
\fill [rounded corners=3pt, yellow!50] (\posX-1, \posY-0.15+\pos*\deltaY) rectangle (\posX+1, \posY+0.15+\pos*\deltaY);    
\draw  [black] (\posX, \posY+\pos*\deltaY) node {\scriptsize{3 $\times$ 3, 128}};    
\draw [->] (\posX, \posY-0.2+\pos*\deltaY) -- (\posX, \posY-0.5+\pos*\deltaY);

%skip
\draw [->, dashed] (\posX, {\posY-0.2+(\pos-2.1)*\deltaY-0.2}) .. controls (\posX+2.25, {\posY-0.2+(\pos-1.5)*\deltaY})  and (\posX+2.25, {\posY-0.2+(\pos-0.0)*\deltaY})  ..  (\posX, \posY-0.2+\pos*\deltaY-0.2);

\def\pos{6}
\fill [rounded corners=3pt, yellow!50] (\posX-1, \posY-0.15+\pos*\deltaY) rectangle (\posX+1, \posY+0.15+\pos*\deltaY);    
\draw  [black] (\posX, \posY+\pos*\deltaY) node {\scriptsize{3 $\times$ 3, 128}};    
\draw [->] (\posX, \posY-0.2+\pos*\deltaY) -- (\posX, \posY-0.5+\pos*\deltaY);

%rectangle
\draw [rounded corners, black, dashed, thick] (\posX-1.75,\posY+0.35+\pos*\deltaY) rectangle (\posX+2, \posY-1.15+\pos*\deltaY);    
\draw (\posX+1.75, \posY+0.55+\pos*\deltaY) node {\textbf{3$\times$}};

\def\pos{7}
\fill [rounded corners=3pt, yellow!50] (\posX-1, \posY-0.15+\pos*\deltaY) rectangle (\posX+1, \posY+0.15+\pos*\deltaY);    
\draw  [black] (\posX, \posY+\pos*\deltaY) node {\scriptsize{3 $\times$ 3, 128}};    
\draw [->] (\posX, \posY-0.2+\pos*\deltaY) -- (\posX, \posY-0.5+\pos*\deltaY);

%skip
\draw [->] (\posX, {\posY-0.2+(\pos-2.1)*\deltaY-0.2}) .. controls (\posX+2.25, {\posY-0.2+(\pos-1.5)*\deltaY})  and (\posX+2.25, {\posY-0.2+(\pos-0.0)*\deltaY})  ..  (\posX, \posY-0.2+\pos*\deltaY-0.2);

\def\pos{8}
\fill [rounded corners=3pt, yellow!50] (\posX-1, \posY-0.15+\pos*\deltaY) rectangle (\posX+1, \posY+0.15+\pos*\deltaY);    
\draw  [black] (\posX, \posY+\pos*\deltaY) node {\scriptsize{3 $\times$ 3, 256, /2}};    
\draw [->] (\posX, \posY-0.2+\pos*\deltaY) -- (\posX, \posY-0.5+\pos*\deltaY);

\def\pos{9}
\fill [rounded corners=3pt, yellow!50] (\posX-1, \posY-0.15+\pos*\deltaY) rectangle (\posX+1, \posY+0.15+\pos*\deltaY);    
\draw  [black] (\posX, \posY+\pos*\deltaY) node {\scriptsize{3 $\times$ 3, 256}};    
\draw [->] (\posX, \posY-0.2+\pos*\deltaY) -- (\posX, \posY-0.5+\pos*\deltaY);

%skip
\draw [->, dashed] (\posX, {\posY-0.2+(\pos-2.1)*\deltaY-0.2}) .. controls (\posX+2.25, {\posY-0.2+(\pos-1.5)*\deltaY})  and (\posX+2.25, {\posY-0.2+(\pos-0.0)*\deltaY})  ..  (\posX, \posY-0.2+\pos*\deltaY-0.2);

\def\pos{10}
\fill [rounded corners=3pt, yellow!50] (\posX-1, \posY-0.15+\pos*\deltaY) rectangle (\posX+1, \posY+0.15+\pos*\deltaY);    
\draw  [black] (\posX, \posY+\pos*\deltaY) node {\scriptsize{3 $\times$ 3, 256}};    
\draw [->] (\posX, \posY-0.2+\pos*\deltaY) -- (\posX, \posY-0.5+\pos*\deltaY);

%rectangle
\draw [rounded corners, black, dashed, thick] (\posX-1.75,\posY+0.35+\pos*\deltaY) rectangle (\posX+2, \posY-1.15+\pos*\deltaY);    
\draw (\posX+1.75, \posY+0.55+\pos*\deltaY) node {\textbf{5$\times$}};

\def\pos{11}
\fill [rounded corners=3pt, yellow!50] (\posX-1, \posY-0.15+\pos*\deltaY) rectangle (\posX+1, \posY+0.15+\pos*\deltaY);    
\draw  [black] (\posX, \posY+\pos*\deltaY) node {\scriptsize{3 $\times$ 3, 256}};    
\draw [->] (\posX, \posY-0.2+\pos*\deltaY) -- (\posX, \posY-0.5+\pos*\deltaY);

%skip
\draw [->] (\posX, {\posY-0.2+(\pos-2.1)*\deltaY-0.2}) .. controls (\posX+2.25, {\posY-0.2+(\pos-1.5)*\deltaY})  and (\posX+2.25, {\posY-0.2+(\pos-0.0)*\deltaY})  ..  (\posX, \posY-0.2+\pos*\deltaY-0.2);

\def\pos{12}
\fill [rounded corners=3pt, yellow!50] (\posX-1, \posY-0.15+\pos*\deltaY) rectangle (\posX+1, \posY+0.15+\pos*\deltaY);    
\draw  [black] (\posX, \posY+\pos*\deltaY) node {\scriptsize{3 $\times$ 3, 512, /2}};    
\draw [->] (\posX, \posY-0.2+\pos*\deltaY) -- (\posX, \posY-0.5+\pos*\deltaY);

\def\pos{13}
\fill [rounded corners=3pt, yellow!50] (\posX-1, \posY-0.15+\pos*\deltaY) rectangle (\posX+1, \posY+0.15+\pos*\deltaY);    
\draw  [black] (\posX, \posY+\pos*\deltaY) node {\scriptsize{3 $\times$ 3, 512}};    
\draw [->] (\posX, \posY-0.2+\pos*\deltaY) -- (\posX, \posY-0.5+\pos*\deltaY);

%skip
\draw [->, dashed] (\posX, {\posY-0.2+(\pos-2.1)*\deltaY-0.2}) .. controls (\posX+2.25, {\posY-0.2+(\pos-1.5)*\deltaY})  and (\posX+2.25, {\posY-0.2+(\pos-0.0)*\deltaY})  ..  (\posX, \posY-0.2+\pos*\deltaY-0.2);

\def\pos{14}
\fill [rounded corners=3pt, yellow!50] (\posX-1, \posY-0.15+\pos*\deltaY) rectangle (\posX+1, \posY+0.15+\pos*\deltaY);    
\draw  [black] (\posX, \posY+\pos*\deltaY) node {\scriptsize{3 $\times$ 3, 512}};    
\draw [->] (\posX, \posY-0.2+\pos*\deltaY) -- (\posX, \posY-0.5+\pos*\deltaY);

%rectangle
\draw [rounded corners, black, dashed, thick] (\posX-1.75,\posY+0.35+\pos*\deltaY) rectangle (\posX+2, \posY-1.15+\pos*\deltaY);    
\draw (\posX+1.75, \posY+0.55+\pos*\deltaY) node {\textbf{2$\times$}};

\def\pos{15}
\fill [rounded corners=3pt, yellow!50] (\posX-1, \posY-0.15+\pos*\deltaY) rectangle (\posX+1, \posY+0.15+\pos*\deltaY);    
\draw  [black] (\posX, \posY+\pos*\deltaY) node {\scriptsize{3 $\times$ 3, 512}};    
\draw [->] (\posX, \posY-0.2+\pos*\deltaY) -- (\posX, \posY-0.5+\pos*\deltaY);

%skip
\draw [->] (\posX, {\posY-0.2+(\pos-2.1)*\deltaY-0.2}) .. controls (\posX+2.25, {\posY-0.2+(\pos-1.5)*\deltaY})  and (\posX+2.25, {\posY-0.2+(\pos-0.0)*\deltaY})  ..  (\posX, \posY-0.2+\pos*\deltaY-0.2);

\def\pos{16}
\fill [rounded corners=3pt, blue!50] (\posX-1, \posY-0.15+\pos*\deltaY) rectangle (\posX+1, \posY+0.15+\pos*\deltaY);    
\draw  [black] (\posX, \posY+\pos*\deltaY) node {\scriptsize{avg, 2 $\times$ 2}};    
\draw [->] (\posX, \posY-0.2+\pos*\deltaY) -- (\posX, \posY-0.5+\pos*\deltaY);

\def\pos{17}
\fill [rounded corners=3pt, green!50] (\posX-1, \posY-0.15+\pos*\deltaY) rectangle (\posX+1, \posY+0.15+\pos*\deltaY);    
\draw  [black] (\posX, \posY+\pos*\deltaY) node {\scriptsize{dropout, 0.75}};    
\draw [->] (\posX, \posY-0.2+\pos*\deltaY) -- (\posX, \posY-0.5+\pos*\deltaY);

\def\pos{18}
\fill [rounded corners=3pt, red!50] (\posX-1, \posY-0.15+\pos*\deltaY) rectangle (\posX+1, \posY+0.15+\pos*\deltaY);    
\draw  [black] (\posX, \posY+\pos*\deltaY) node {\scriptsize{4,096}};

\def\pos{19}
\fill [rounded corners=3pt, red!50] (\posX-1, \posY-0.15+\pos*\deltaY) rectangle (\posX+1, \posY+0.15+\pos*\deltaY);    
\draw  [black] (\posX, \posY+\pos*\deltaY) node {\scriptsize{$d$}};    

\end{tikzpicture}
    \caption{Architectures of the CNNs used in the empirical validation of our method. The yellow boxes represent convolutional layers, and the blue and green boxes represent pooling and dropout (keeping probability 0.75) layers. Finally, the red boxes denote fully connected layers. In the ResNet architecture, the dashed skip connections represent convolutions with stride 2 $\times$ 2, yielding outputs with half of the spatial input size. The ''/2'' symbol in the convolution layers denotes stride 2 $\times$ 2 (the remaining layers use stride 1 $\times$ 1).}
        \label{fig:CNN}
    \end{center}
\end{figure}
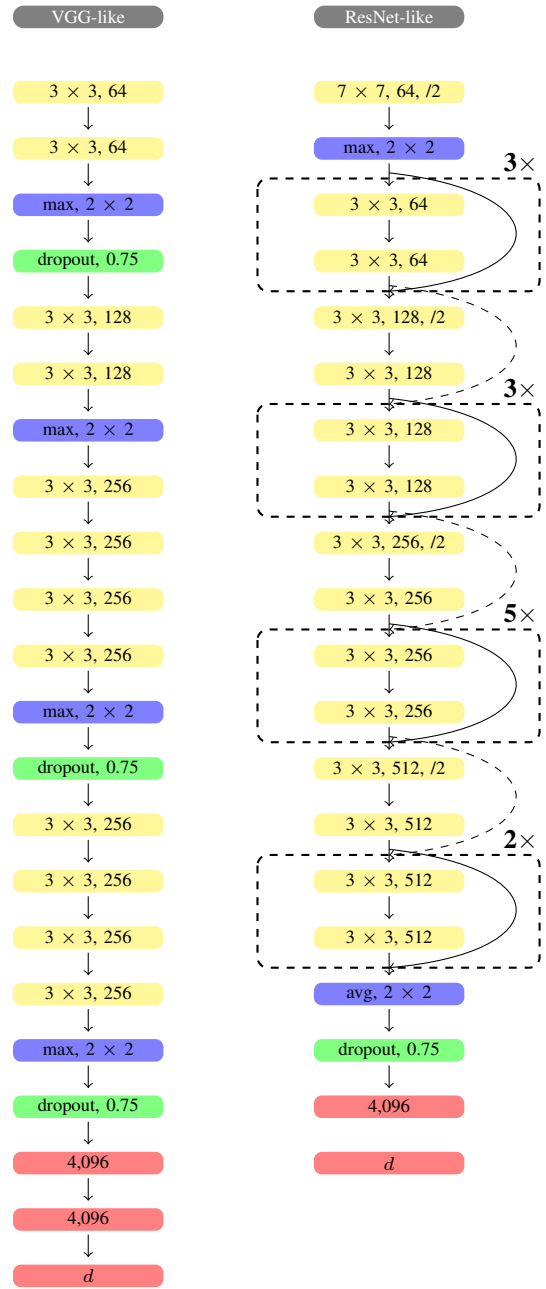

Two CNN architectures were considered, based in the \emph{VGG} and \emph{ResNet} models (Fig.~\ref{fig:CNN}). Here, our  goals were essentially to compare the performance of the quadruplet loss with respect to the baselines and also to perceive any variations in performance with respect to different learning architectures. A \emph{TensorFlow} implementation of the of both architectures is available at\footnote{\url{https://github.com/hugomcp/quadruplets}}.  

All the models were initialized with random weights, from zero-mean Gaussian distributions with standard deviation 0.01 and bias 0.5. Images were resized to 256 $\times$ 256, adding lateral white bands when needed to keep constant ratios. A batch size of 64 was defined, which obviously results in too many combinations of pairs for the triplet/quadruplet losses. At each iteration, we filtered out the invalid triplets/quadruplets instances and  randomly selected the mini-batch learning elements, composed of 64 instances in all cases. For all other baselines, 64 learning instances were also used as a batch. The learning rate started from 0.01, with momentum 0.9 and weight decay $5e^{-4}$. In the \emph{learning-from-scratch} paradigm, we stopped the learning process when the validation loss didn't decrease for 10 iterations (i.e., \emph{patience}=10).

 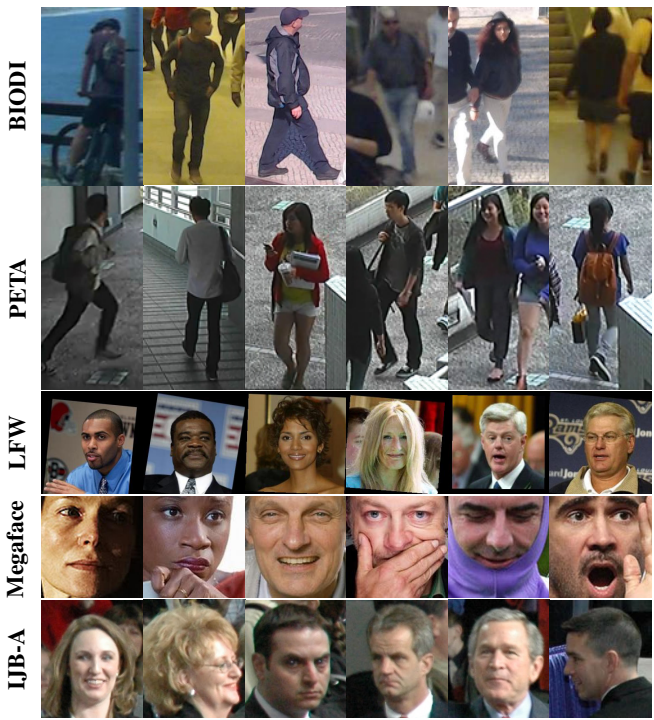
\begin{figure}[ht!]
\begin{center}
\begin{tikzpicture}

\def\sizeImg{1.35}
\def\deltaY{-2.4}

\draw (0*\sizeImg,0) node(segment_ok)  {\includegraphics[width=\sizeImg cm]{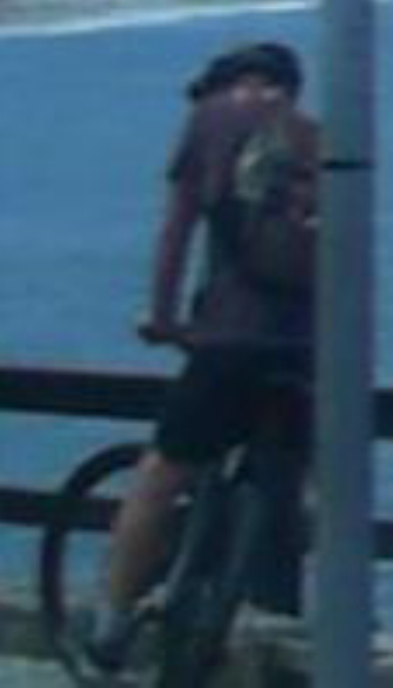}};  	
\draw (1*\sizeImg,0) node(segment_ok)  {\includegraphics[width=\sizeImg cm]{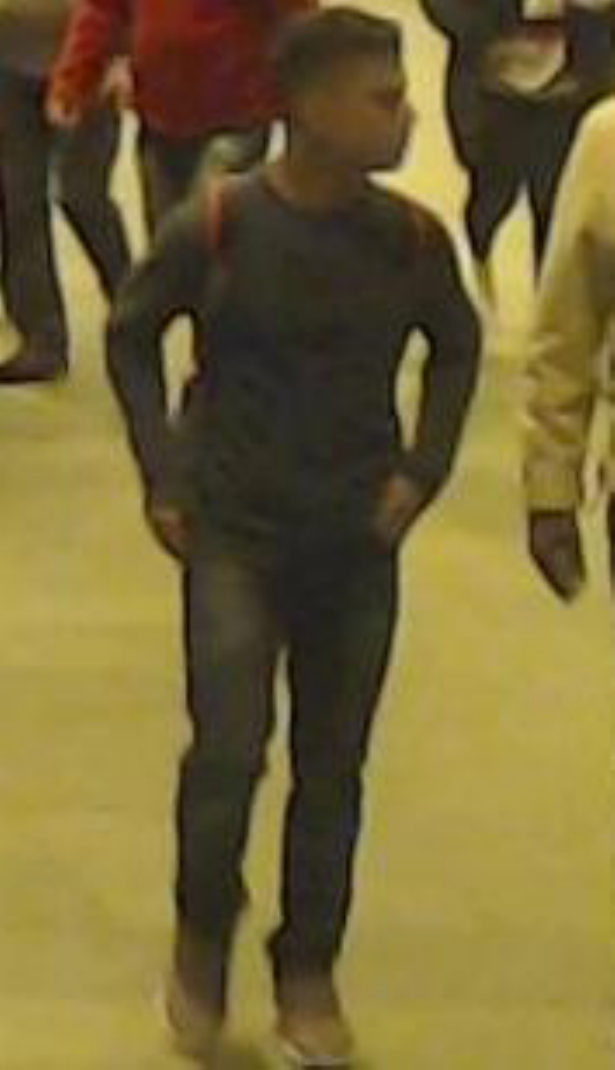}};  	
\draw (2*\sizeImg,0) node(segment_ok)  {\includegraphics[width=\sizeImg cm]{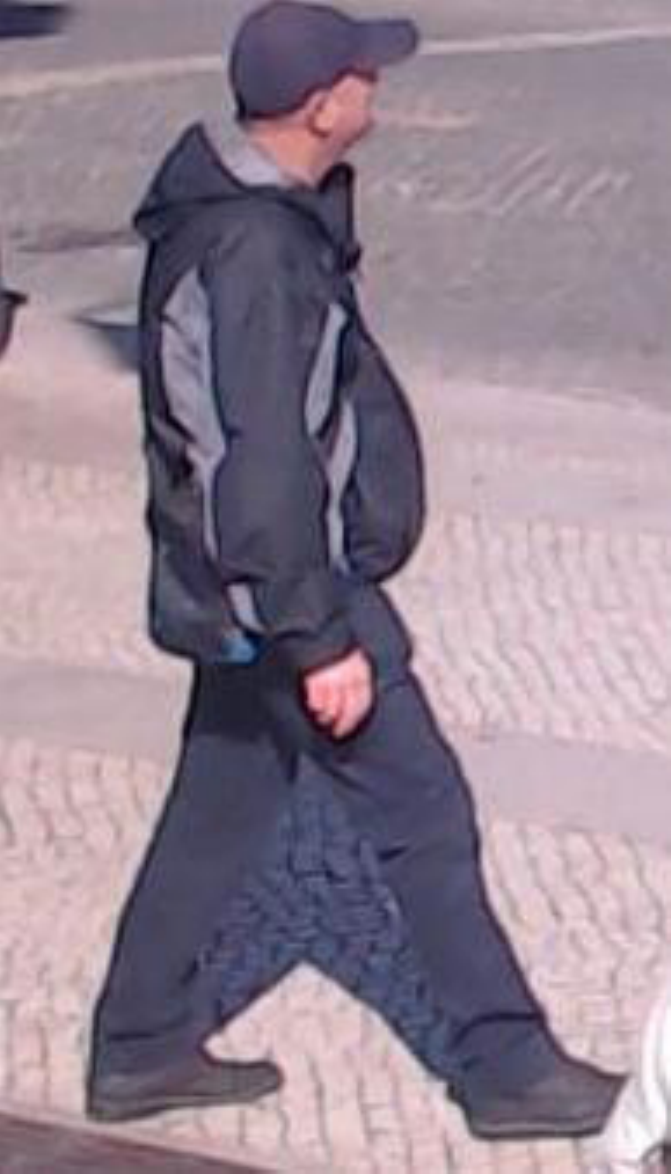}};  	
\draw (3*\sizeImg,0) node(segment_ok)  {\includegraphics[width=\sizeImg cm]{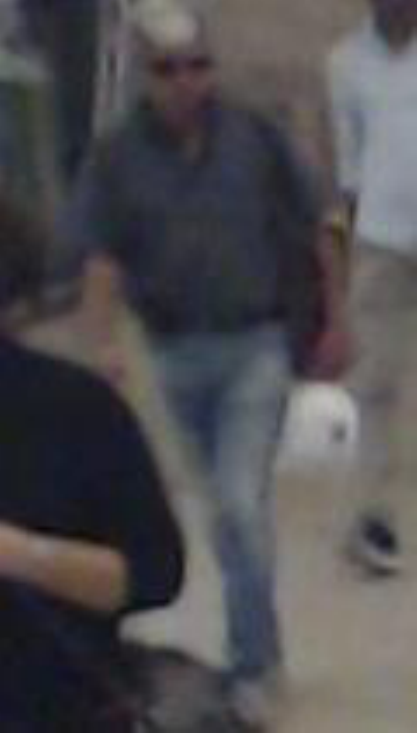}};  	
\draw (4*\sizeImg,0) node(segment_ok)  {\includegraphics[width=\sizeImg cm]{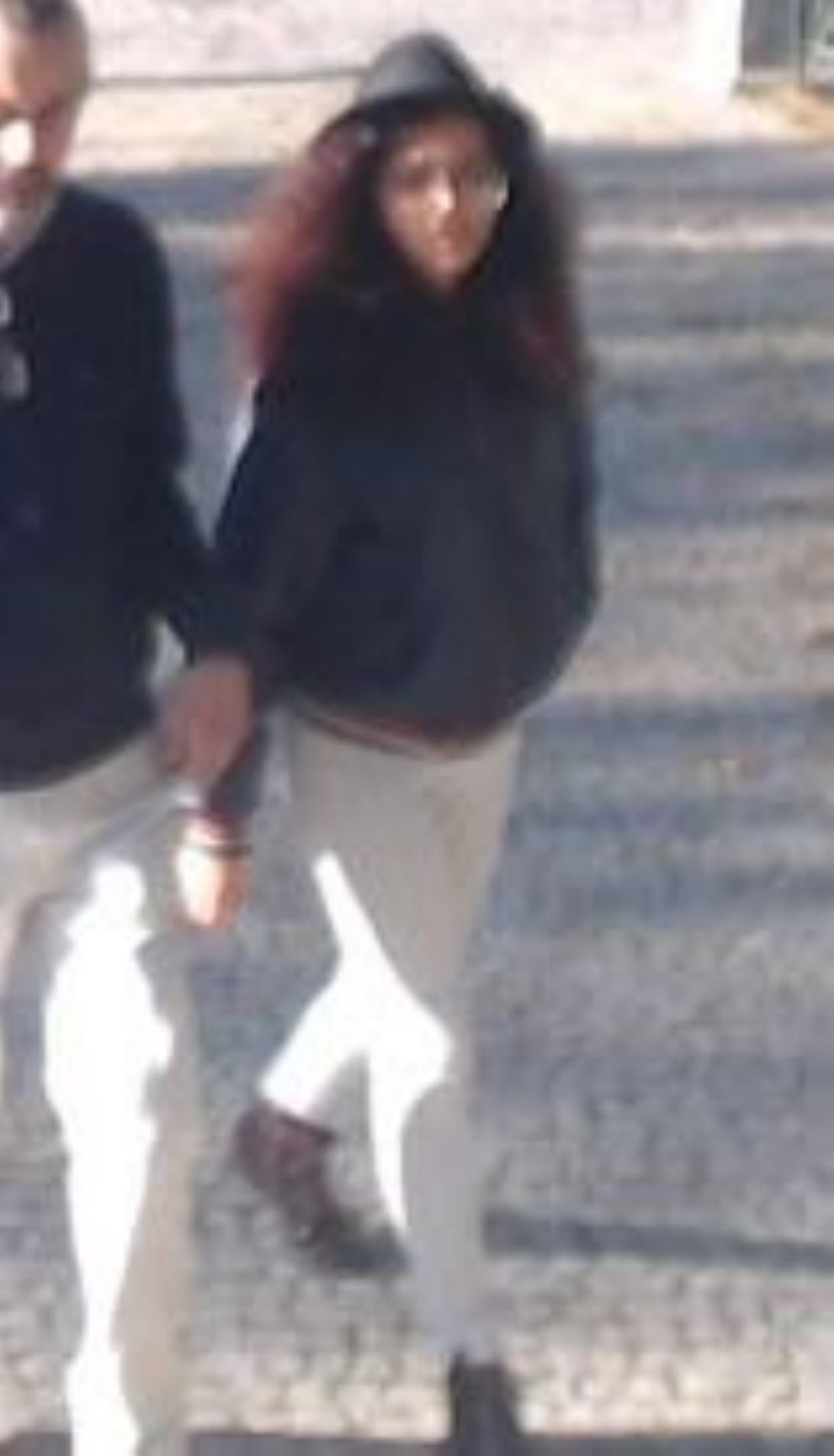}};  	
\draw (5*\sizeImg,0) node(segment_ok)  {\includegraphics[width=\sizeImg cm]{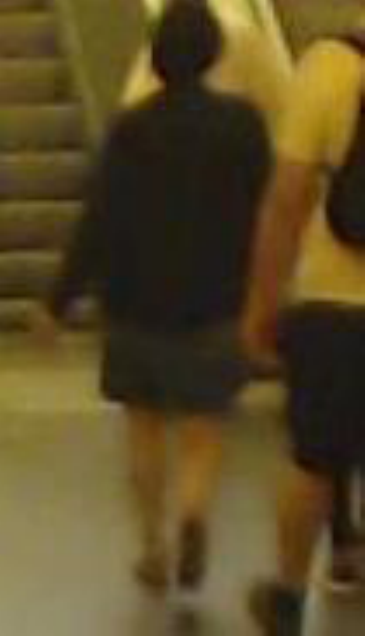}};  	
\draw (-1.0, 0) node[rectangle, rotate=90] {\small{\textbf{BIODI}}};

\draw (0*\sizeImg,0+1.065*\deltaY) node(segment_ok)  {\includegraphics[width=\sizeImg cm]{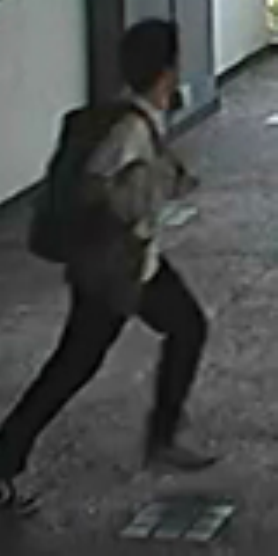}};  	
\draw (1*\sizeImg,0+1.065*\deltaY) node(segment_ok)  {\includegraphics[width=\sizeImg cm]{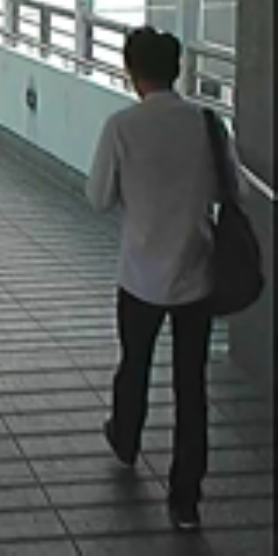}};  	
\draw (2*\sizeImg,0+1.065*\deltaY) node(segment_ok)  {\includegraphics[width=\sizeImg cm]{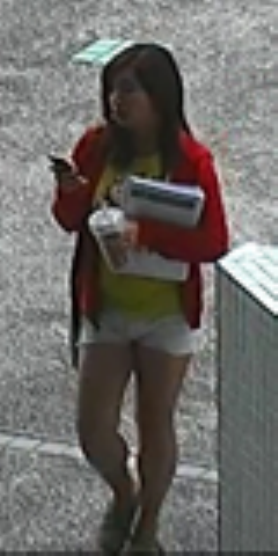}};  	
\draw (3*\sizeImg,0+1.065*\deltaY) node(segment_ok)  {\includegraphics[width=\sizeImg cm]{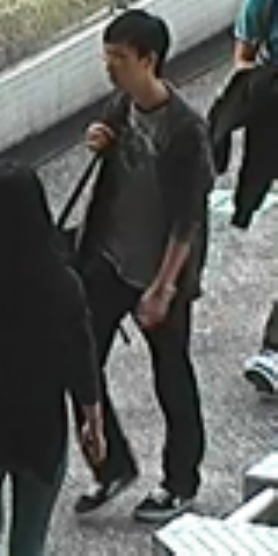}};  	
\draw (4*\sizeImg,0+1.065*\deltaY) node(segment_ok)  {\includegraphics[width=\sizeImg cm]{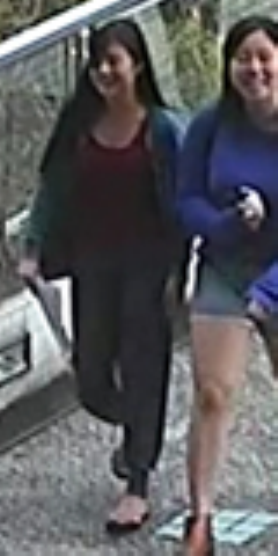}};  	
\draw (5*\sizeImg,0+1.065*\deltaY) node(segment_ok)  {\includegraphics[width=\sizeImg cm]{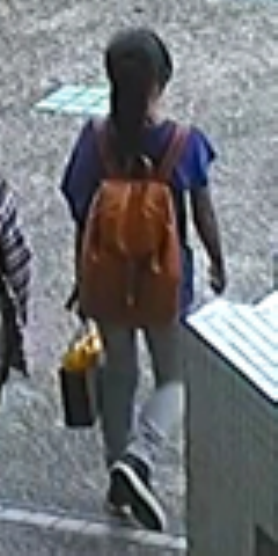}};  	
\draw (-1.0, 0+1.065*\deltaY) node[rectangle, rotate=90] {\small{\textbf{PETA}}};

\draw (0*\sizeImg,0+1.92*\deltaY) node(segment_ok)  {\includegraphics[width=\sizeImg cm]{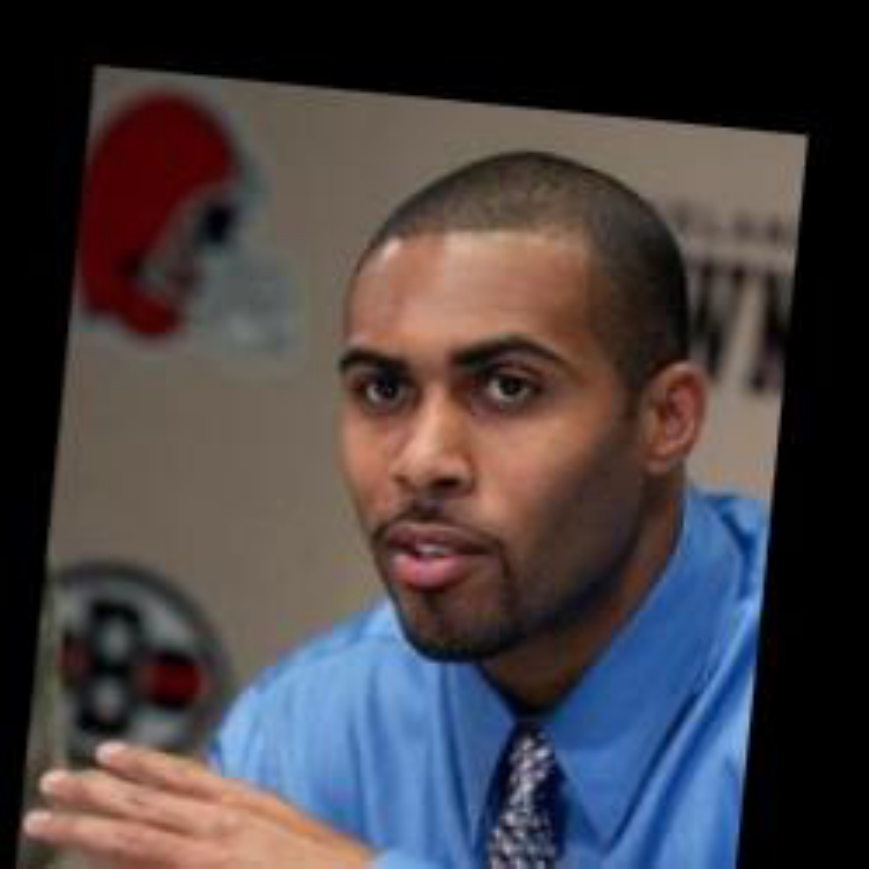}};  	
\draw (1*\sizeImg,0+1.92*\deltaY) node(segment_ok)  {\includegraphics[width=\sizeImg cm]{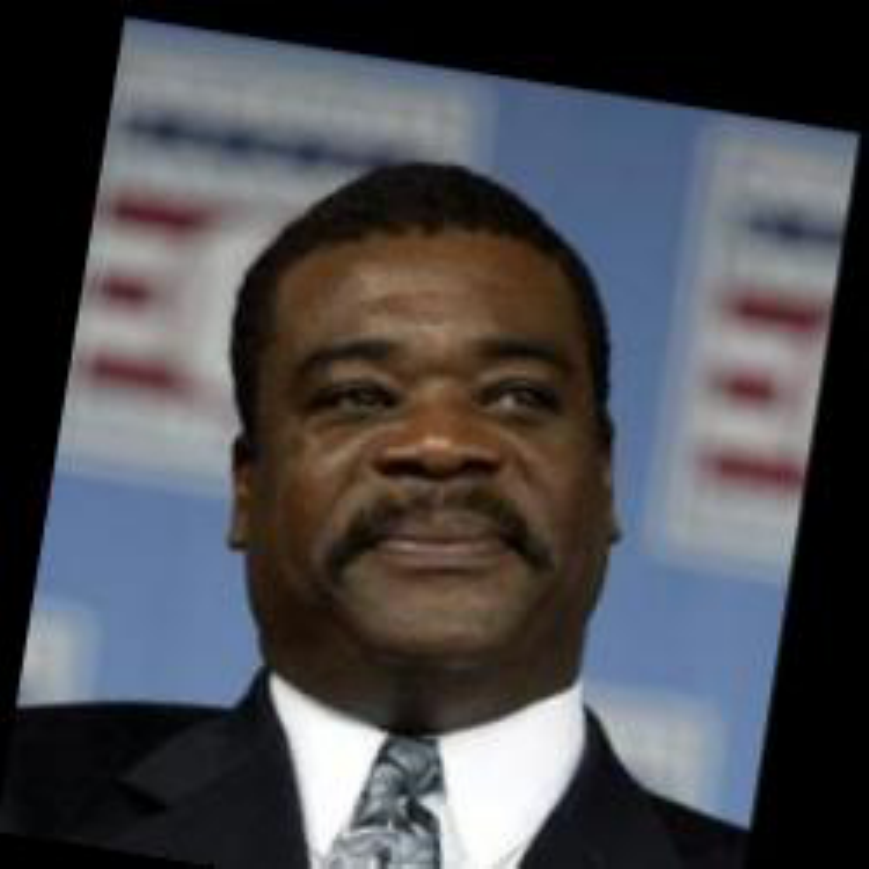}};  	
\draw (2*\sizeImg,0+1.92*\deltaY) node(segment_ok)  {\includegraphics[width=\sizeImg cm]{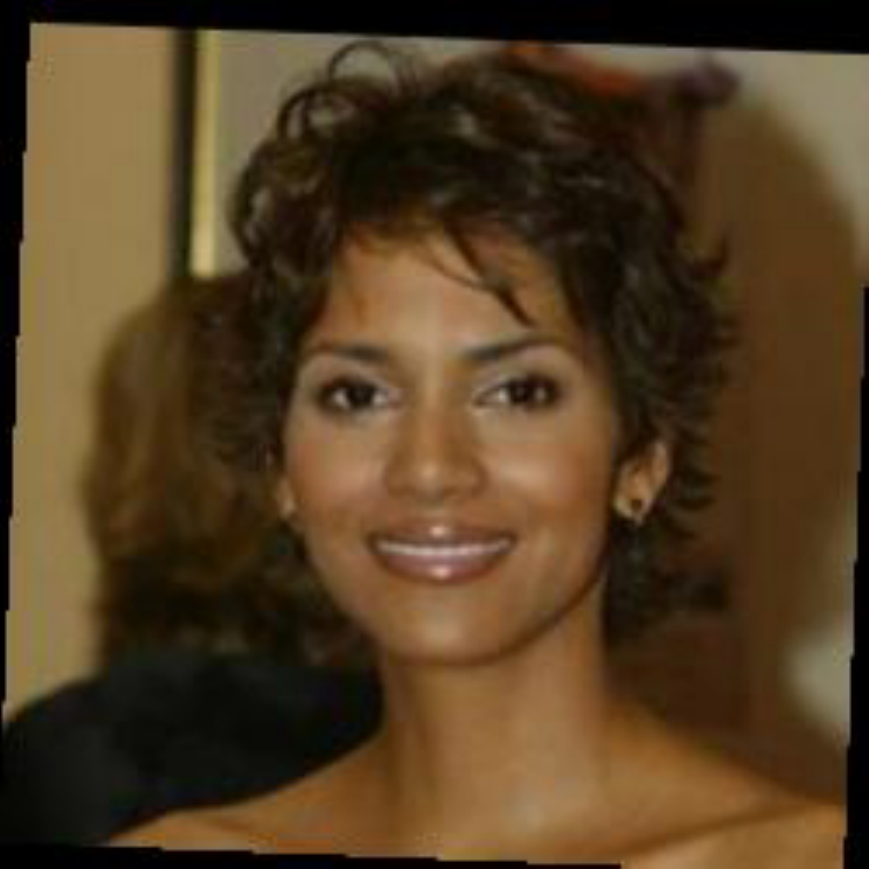}};  	
\draw (3*\sizeImg,0+1.92*\deltaY) node(segment_ok)  {\includegraphics[width=\sizeImg cm]{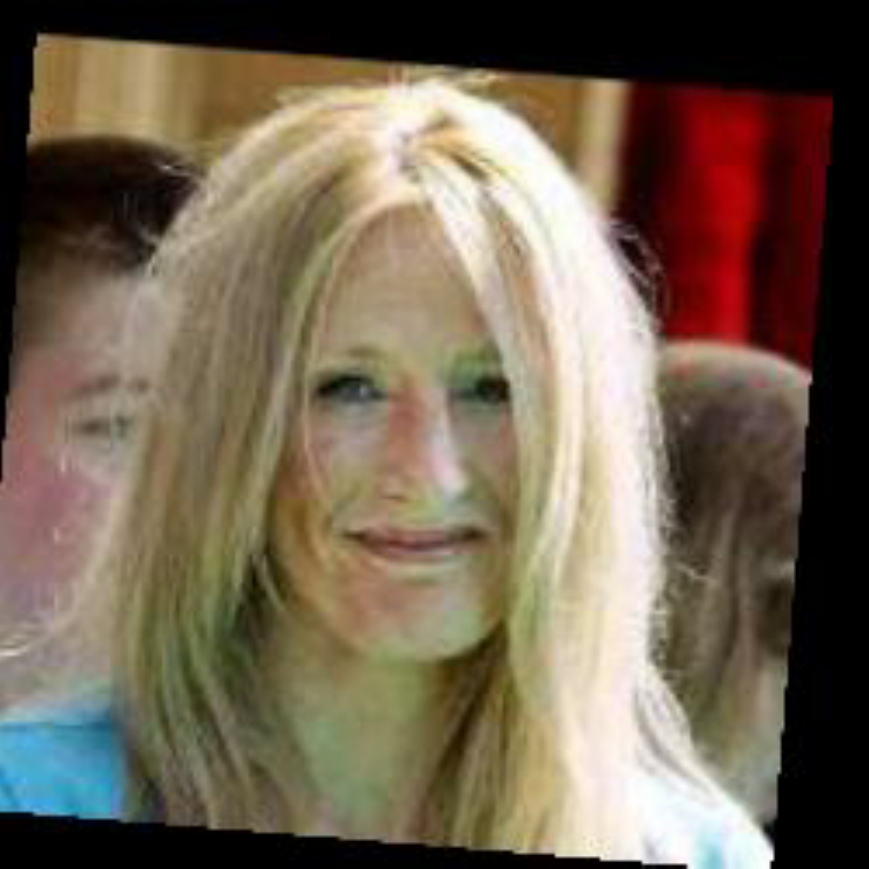}};  	
\draw (4*\sizeImg,0+1.92*\deltaY) node(segment_ok)  {\includegraphics[width=\sizeImg cm]{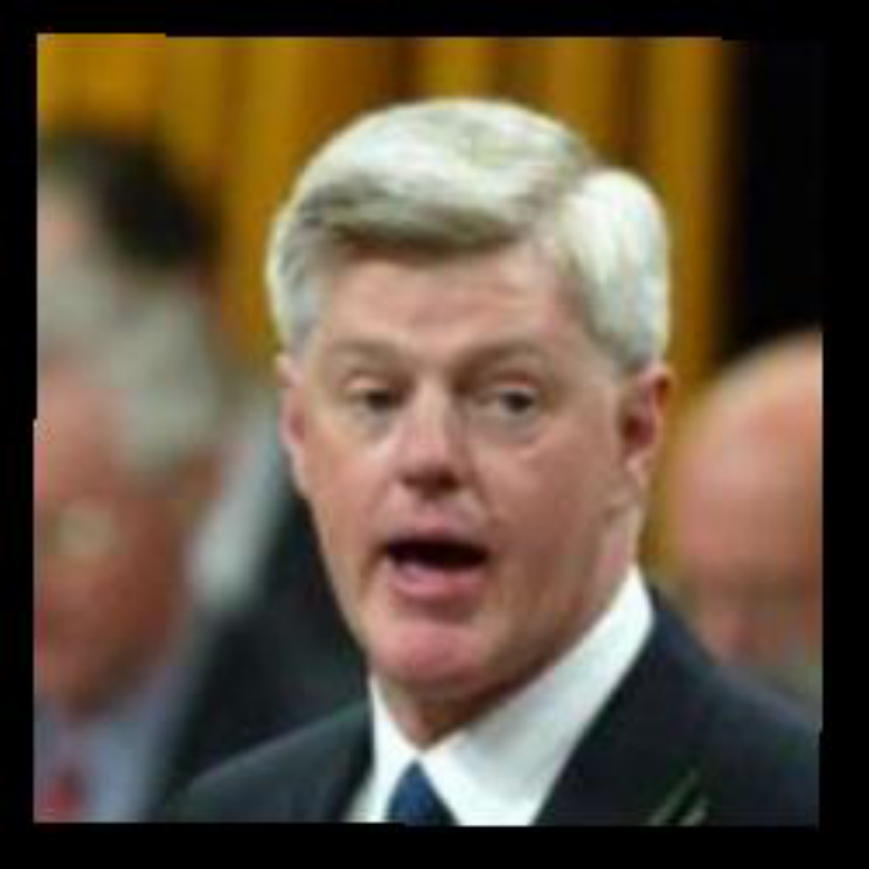}};  	
\draw (5*\sizeImg,0+1.92*\deltaY) node(segment_ok)  {\includegraphics[width=\sizeImg cm]{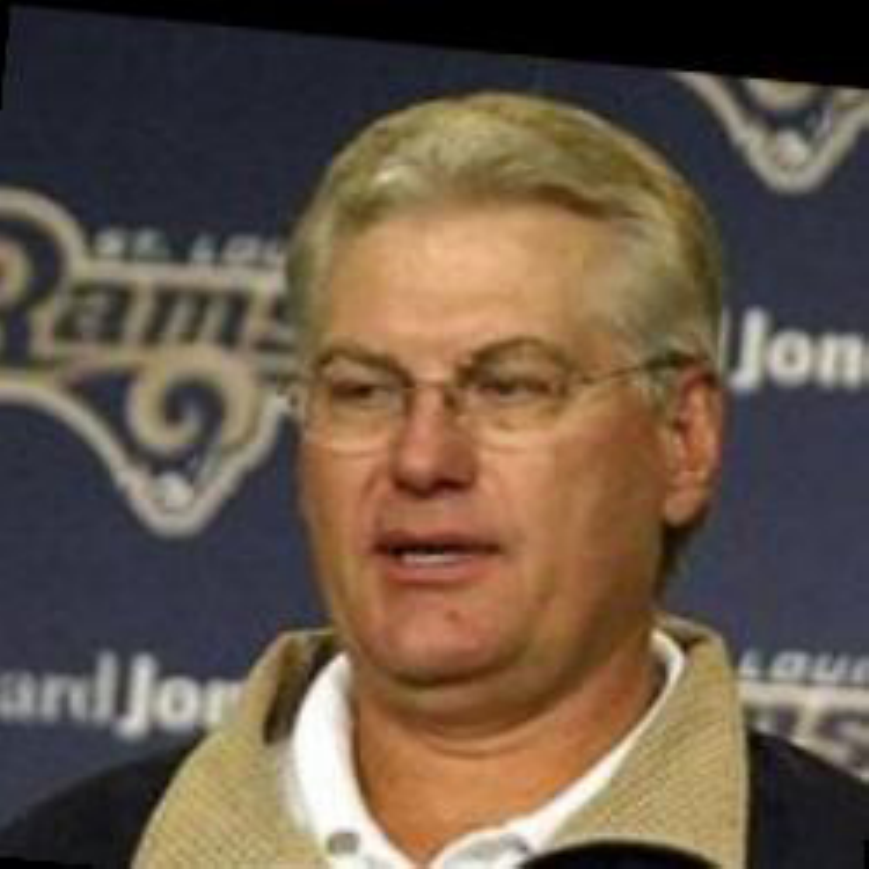}};  	
\draw (-1.0, 0+1.92*\deltaY) node[rectangle, rotate=90] {\small{\textbf{LFW}}};

\draw (0*\sizeImg,0+2.5*\deltaY) node(segment_ok)  {\includegraphics[width=\sizeImg cm]{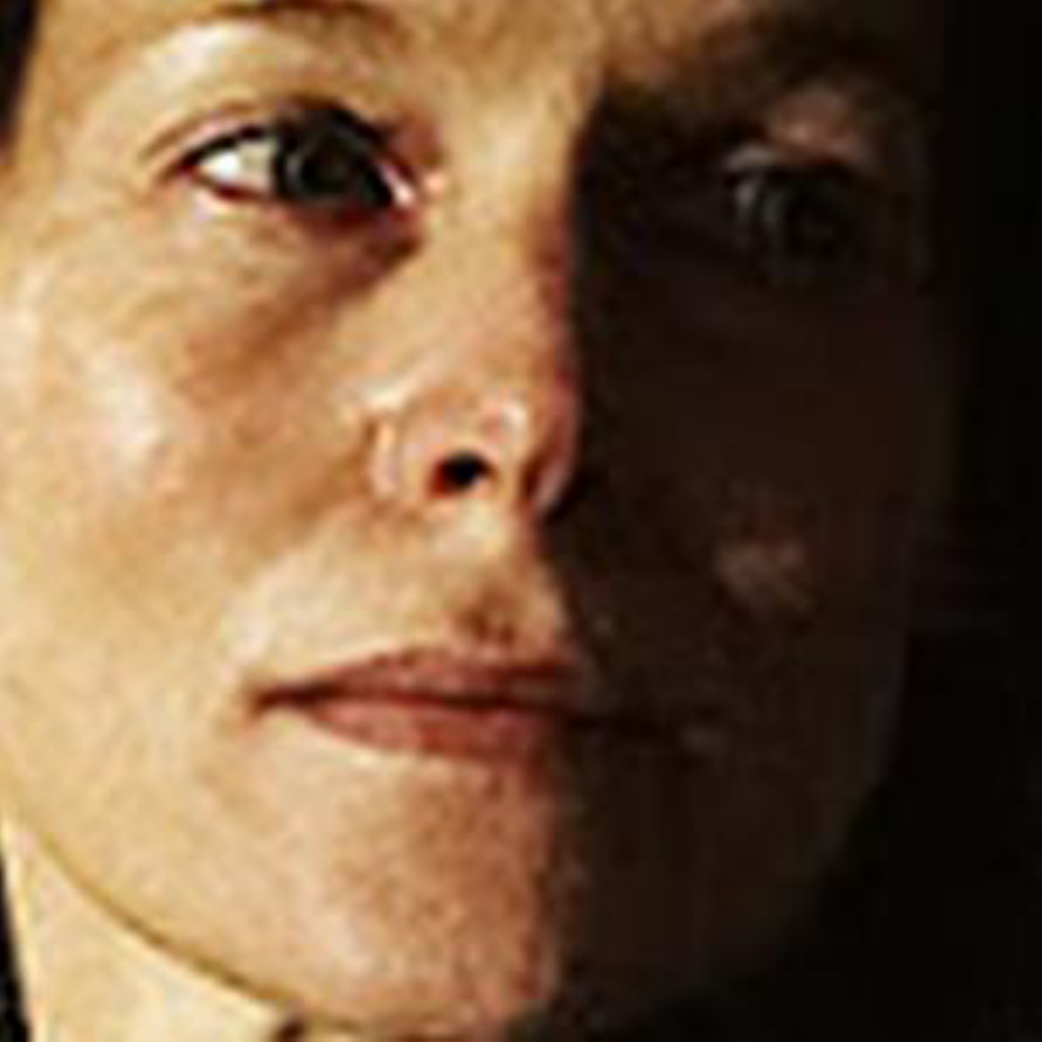}};  	
\draw (1*\sizeImg,0+2.5*\deltaY) node(segment_ok)  {\includegraphics[width=\sizeImg cm]{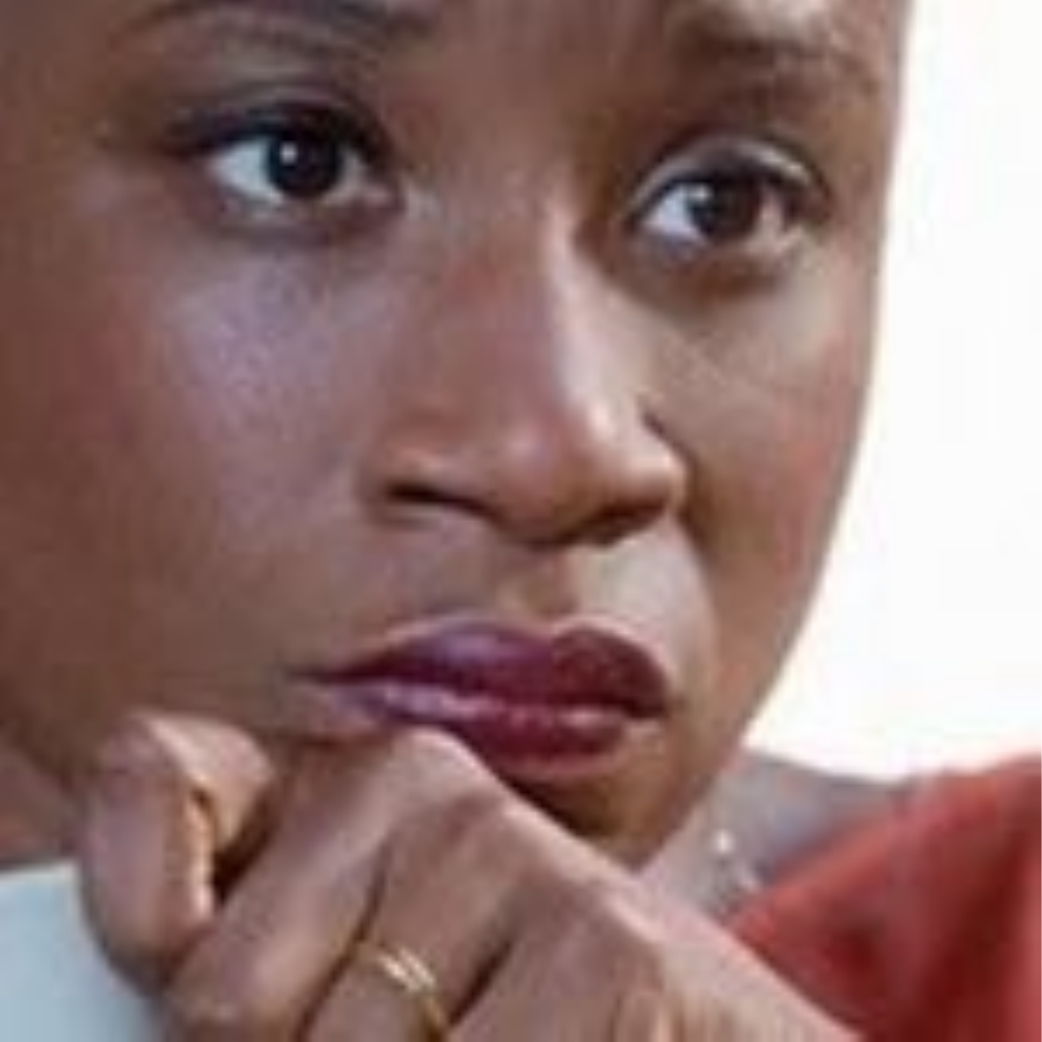}};  	
\draw (2*\sizeImg,0+2.5*\deltaY) node(segment_ok)  {\includegraphics[width=\sizeImg cm]{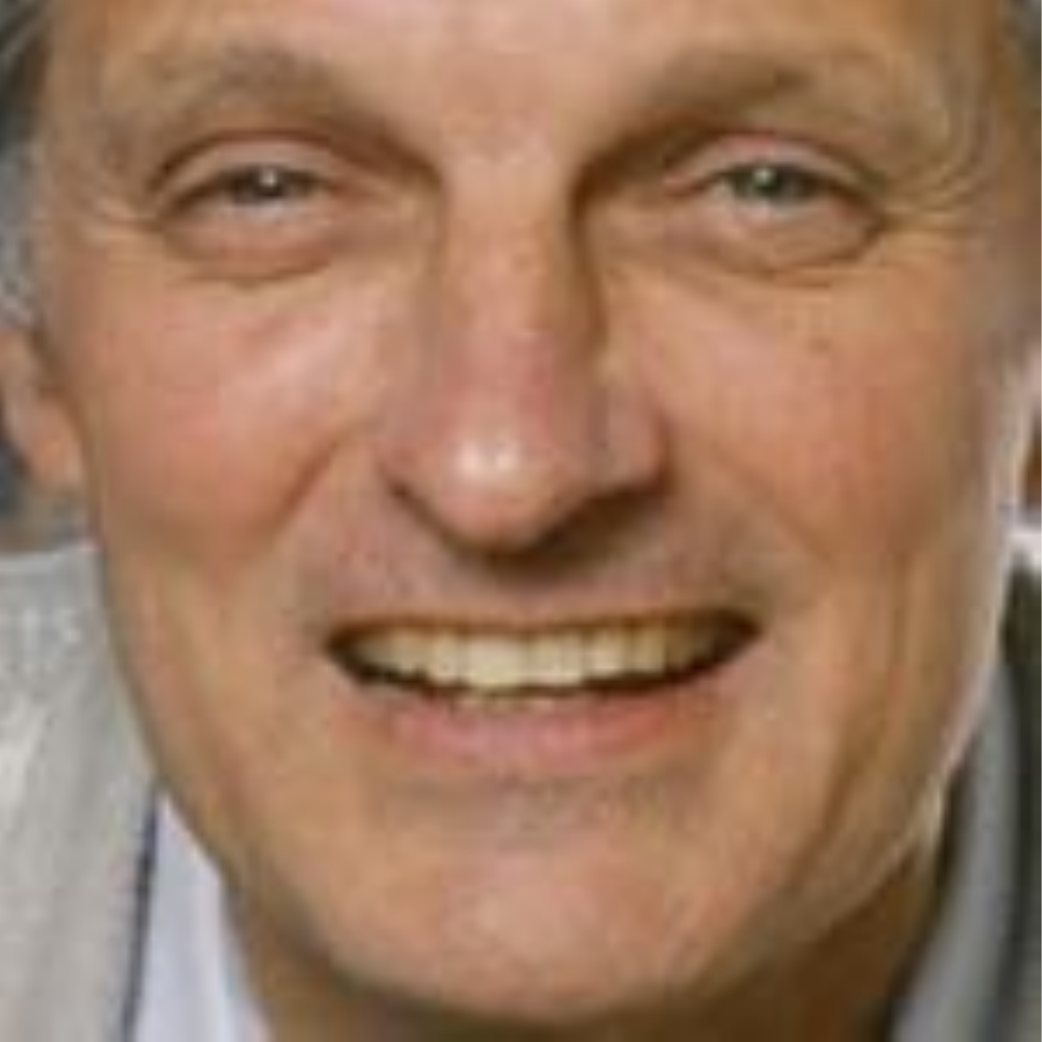}};  	
\draw (3*\sizeImg,0+2.5*\deltaY) node(segment_ok)  {\includegraphics[width=\sizeImg cm]{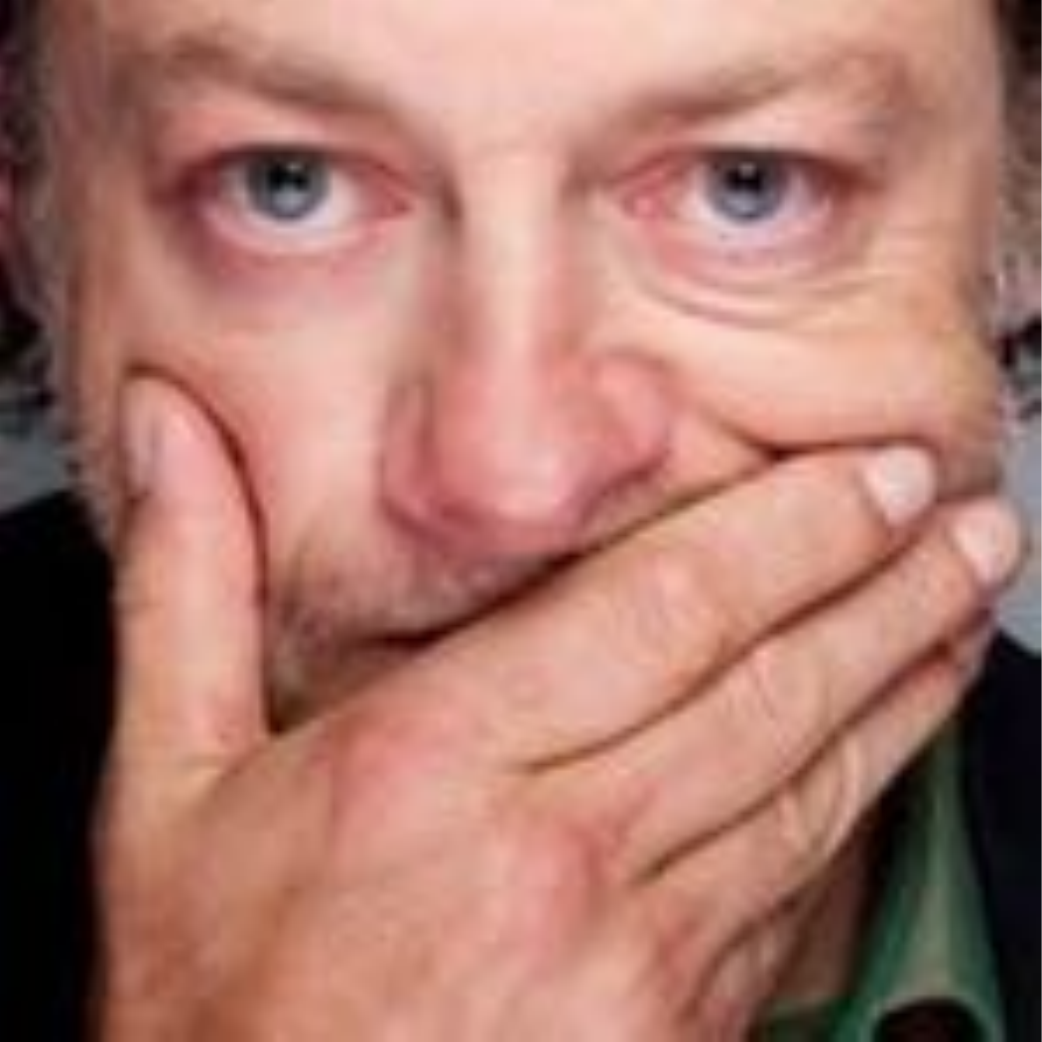}};  	
\draw (4*\sizeImg,0+2.5*\deltaY) node(segment_ok)  {\includegraphics[width=\sizeImg cm]{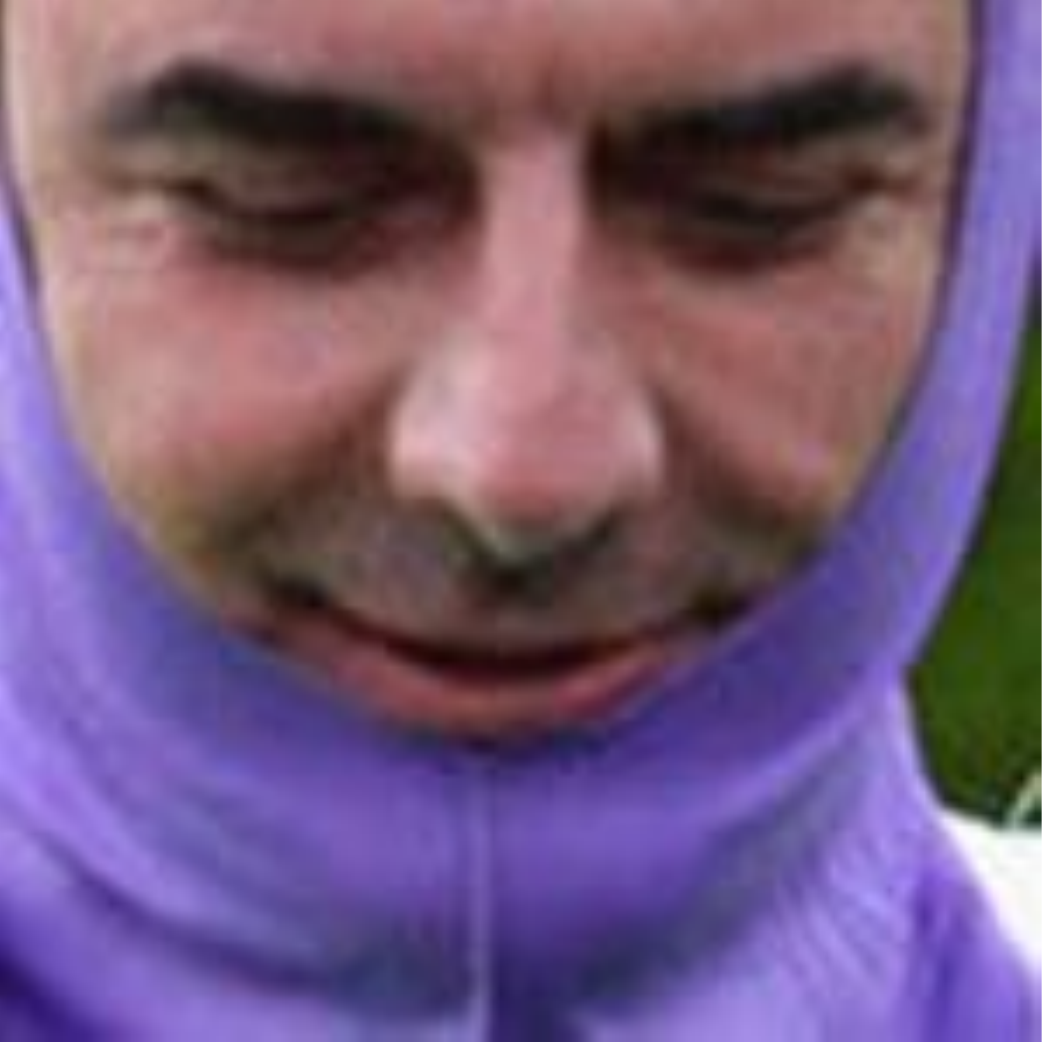}};  	
\draw (5*\sizeImg,0+2.5*\deltaY) node(segment_ok)  {\includegraphics[width=\sizeImg cm]{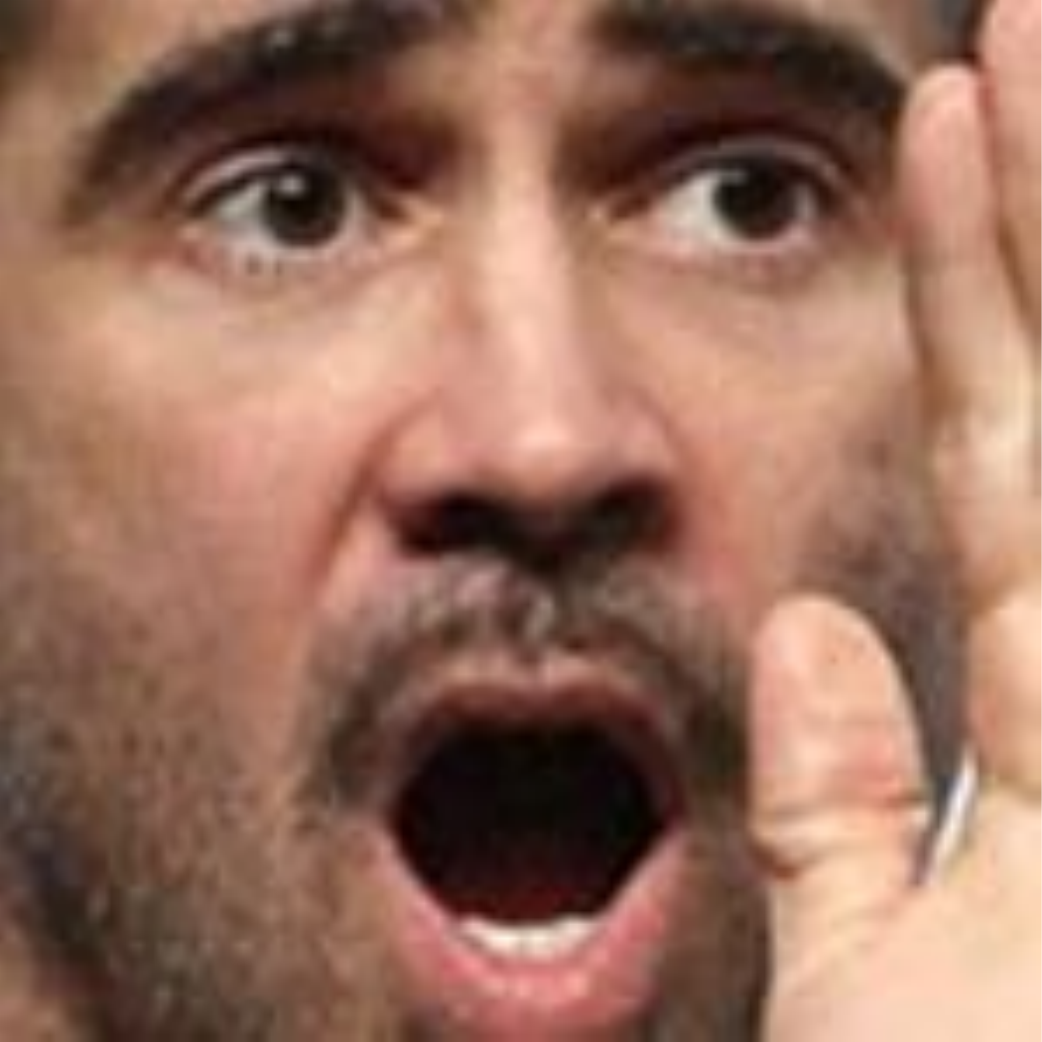}};  	
\draw (-1.0, 0+2.51*\deltaY) node[rectangle, rotate=90] {\small{\textbf{Megaface}}};

\draw (0*\sizeImg,0+3.12*\deltaY) node(segment_ok)  {\includegraphics[width=\sizeImg cm]{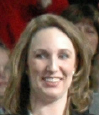}};  	
\draw (1*\sizeImg,0+3.12*\deltaY) node(segment_ok)  {\includegraphics[width=\sizeImg cm]{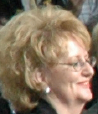}};  	
\draw (2*\sizeImg,0+3.12*\deltaY) node(segment_ok)  {\includegraphics[width=\sizeImg cm]{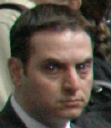}};  	
\draw (3*\sizeImg,0+3.12*\deltaY) node(segment_ok)  {\includegraphics[width=\sizeImg cm]{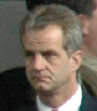}};  	
\draw (4*\sizeImg,0+3.12*\deltaY) node(segment_ok)  {\includegraphics[width=\sizeImg cm]{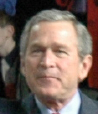}};  	
\draw (5*\sizeImg,0+3.12*\deltaY) node(segment_ok)  {\includegraphics[width=\sizeImg cm]{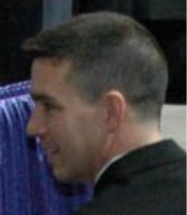}};  	
\draw (-1.0, 0+3.12*\deltaY) node[rectangle, rotate=90] {\small{\textbf{IJB-A}}};  

\end{tikzpicture}
    \caption{Datasets used in the empirical validation of the method proposed in this paper. From top to bottom rows, images of the BIODI, PETA, LFW, Megaface and IJB-A sets are shown.}
        \label{fig:Datasets}
    \end{center}
\end{figure}

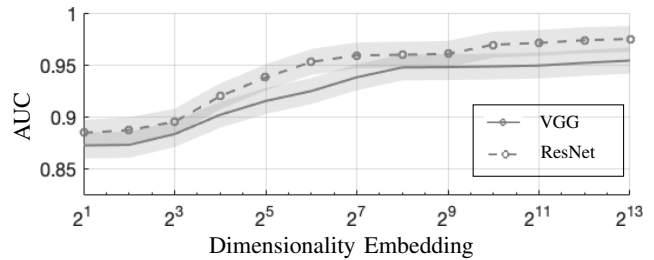
\begin{figure}[ht!]
\begin{center}
\begin{tikzpicture}

\draw (0,0) node(n1)  {\includegraphics[width=8.0 cm]{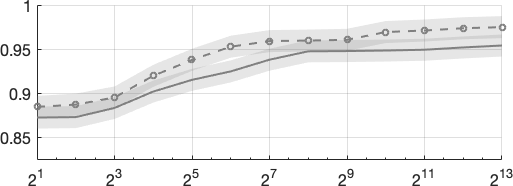}};     

\draw (-4.25, 0) node[rectangle, rotate=90] {\small{AUC}};  
\draw (-0, -1.75) node[rectangle] {\small{Dimensionality Embedding}};
 
\def\deltaY{-1.75}
\draw [black] (1.75,1.9+\deltaY) rectangle (3.75,0.95+\deltaY);    
\draw (2.15, 1.68+\deltaY) node(n1)  {\includegraphics[width=0.5 cm]{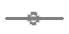}};     
\draw (2.9, 1.68+\deltaY) node[rectangle] {\scriptsize{VGG}}; 

\draw (2.15, 1.28+\deltaY) node(n1)  {\includegraphics[width=0.5 cm]{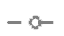}};     
\draw (3.01, 1.28+\deltaY) node[rectangle] {\scriptsize{ResNet}}; 

\end{tikzpicture}
    \caption{Variations in the mean AUC values ($\pm$ the standard deviations after 10 trials, given as shadowed regions) with respect to the dimensionality of the embedding. Results are given for the LFW validation set, when using the VGG-like (solid line) and ResNet-like (dashed line) CNN architectures.}
        \label{fig:Dimensionality}
    \end{center}
\end{figure}

Initially, we varied the dimensionality of the embedding ($d$) to perceive the sensitivity of the proposed loss with respect to this parameter. Considering the LFW set, the observed average AUC values with respect to $d$ are provided in Fig.~\ref{fig:Dimensionality} (the shadowed regions denote the $\pm$ standard deviation performance, after 10 trials). As expected, higher values are directly correlated to the levels of performance, even though results stabilised for dimensions larger than 128. In this regard, we assumed that using even larger dimensions would require much more training data, having resorted from this moment to dimension $d$=128 in all the subsequent experiments.

Interestingly, the absolute performance observed for very low $d$ values was not too far of the obtained for much larger dimensions, which raises the possibility of  using the position of the elements in the destiny space directly for classification and visualization, without the need of  any dimensionality reduction algorithm (MDS, LLE  or  PCA algorithms are frequently seen in the literature for this purpose). 

\subsection{Identity Retrieval}

We report the performance obtained by the quadruplet loss in the identity retrieval task, when compared to the baselines triplet loss, center loss, \emph{softmax} and also to Chen \etal~\cite{Chen2017}'s method.  Here, we considered the LFW, Megaface and IJB-A sets, and three labels: \{''ID'', ''Gender'', ''Ethnicity''\} ($t=3$), with the annotations for the IJB-A set provided by the Face++ algorithm and subjected to human validation. In this setting, it should be noted that all the baselines classify pairs into \emph{positive}/\emph{negative} depending exclusively of the ID, while the proposed loss uses all the labels to determine the relative similarity between any two classes.

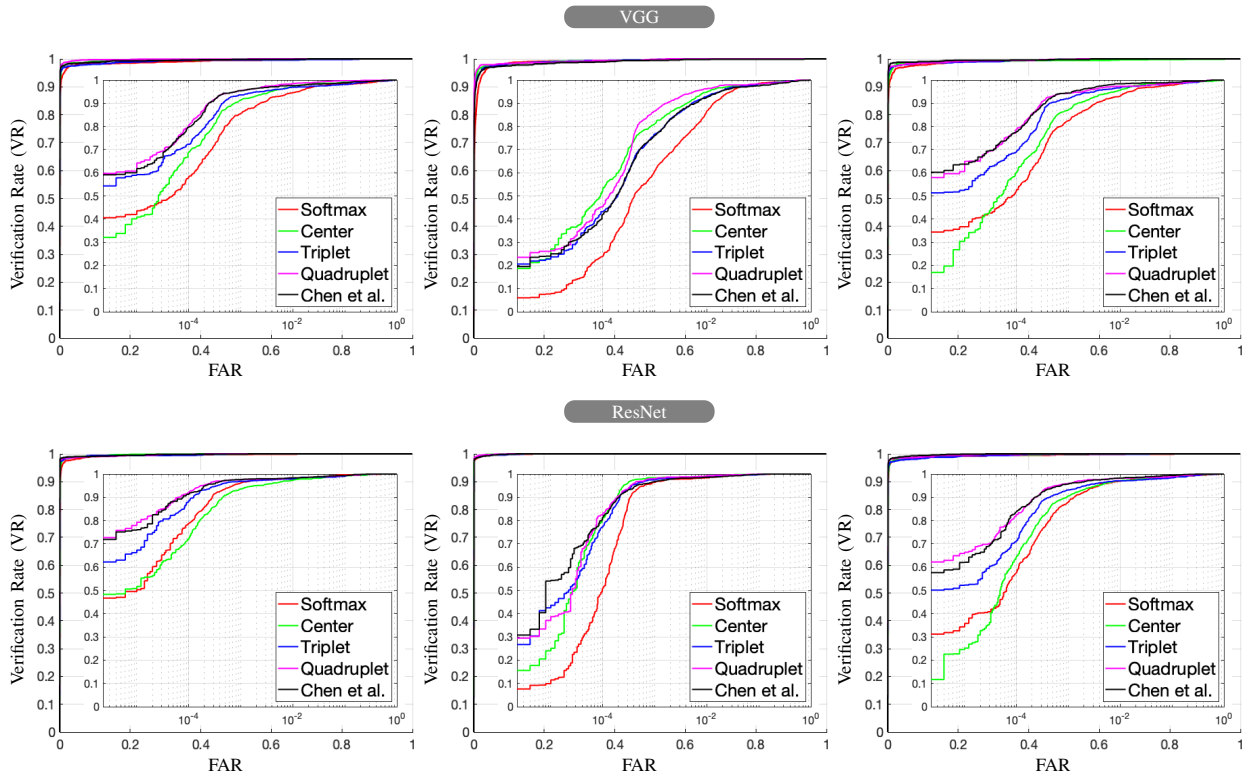
\begin{figure*}[ht!]
\begin{center}
\begin{tikzpicture}

\def\deltaY{0}

\def\posX{5.5}
\def\posY{2.5}
\fill [rounded corners, gray] (\posX-1, \posY-0.15+\deltaY) rectangle (\posX+1, \posY+0.15+\deltaY);    
\draw  [white] (\posX, \posY+\deltaY) node {\scriptsize{VGG}};    

\draw (0,1.8+\deltaY) node[rectangle] {\scriptsize{\textbf{LFW}}};  
\draw (0,0+\deltaY) node(n1)  {\includegraphics[width=5 cm]{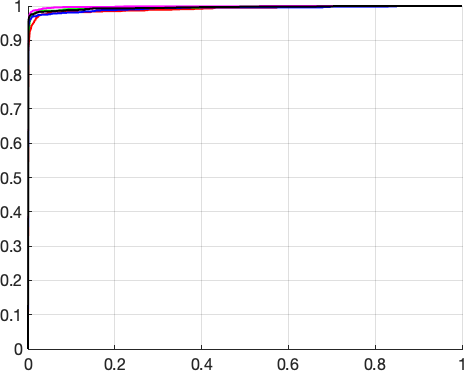}};     
\draw (0.25,0.025+\deltaY) node(n1)  {\includegraphics[width=4.25 cm]{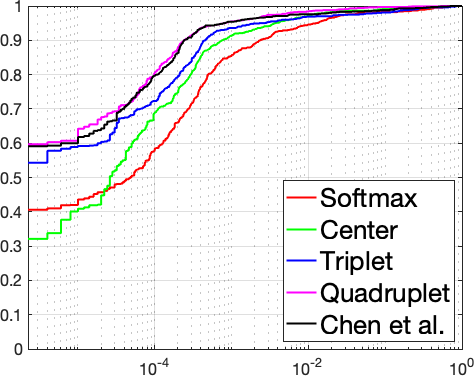}};     
\draw (-2.75, 0+\deltaY) node[rectangle, rotate=90] {\scriptsize{Verification Rate (VR)}};  
\draw (0,-2.2+\deltaY) node[rectangle] {\scriptsize{FAR}};

\def\deltaX{5.5}
 \draw (0+\deltaX,1.8+\deltaY) node[rectangle] {\scriptsize{\textbf{Megaface}}};  
\draw (0+\deltaX,0+\deltaY) node(n1)  {\includegraphics[width=5 cm]{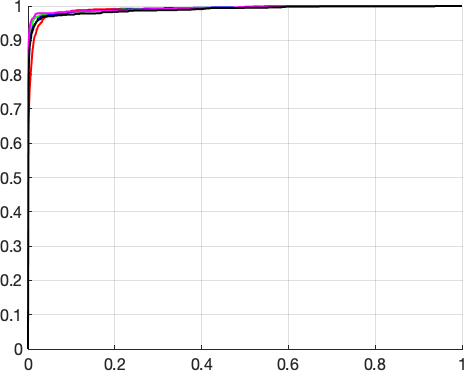}};     
\draw (0.25+\deltaX,0.025+\deltaY) node(n1)  {\includegraphics[width=4.25 cm]{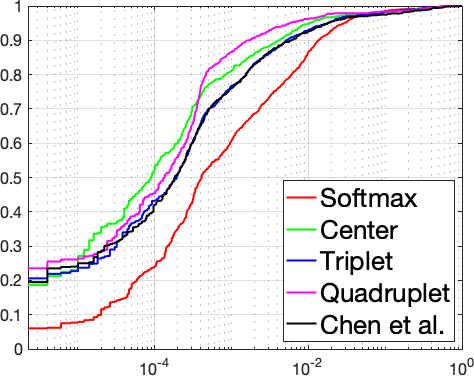}};     
\draw (-2.75+\deltaX, 0+\deltaY) node[rectangle, rotate=90] {\scriptsize{Verification Rate (VR)}};  
\draw (0+\deltaX,-2.2+\deltaY) node[rectangle] {\scriptsize{FAR}};  

\def\deltaX{11.0}
 \draw (0+\deltaX,1.8+\deltaY) node[rectangle] {\scriptsize{\textbf{IJB-A}}};  
\draw (0+\deltaX,0+\deltaY) node(n1)  {\includegraphics[width=5 cm]{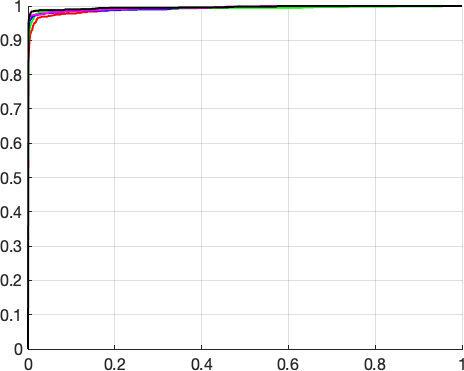}};     
\draw (0.25+\deltaX,0.025+\deltaY) node(n1)  {\includegraphics[width=4.25 cm]{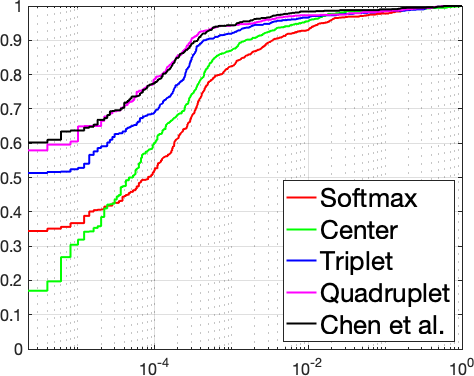}};     
\draw (-2.75+\deltaX, 0+\deltaY) node[rectangle, rotate=90] {\scriptsize{Verification Rate (VR)}};  
\draw (0+\deltaX,-2.2+\deltaY) node[rectangle] {\scriptsize{FAR}};  

\def\deltaY{-5.25}

\def\posX{5.5}
\def\posY{2.5}
\fill [rounded corners, gray] (\posX-1, \posY-0.15+\deltaY) rectangle (\posX+1, \posY+0.15+\deltaY);    
\draw  [white] (\posX, \posY+\deltaY) node {\scriptsize{ResNet}};    

\draw (0,1.8+\deltaY) node[rectangle] {\scriptsize{\textbf{LFW}}};  
\draw (0,0+\deltaY) node(n1)  {\includegraphics[width=5 cm]{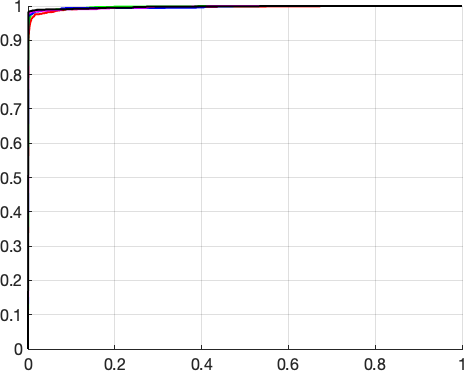}};     
\draw (0.25,0.025+\deltaY) node(n1)  {\includegraphics[width=4.25 cm]{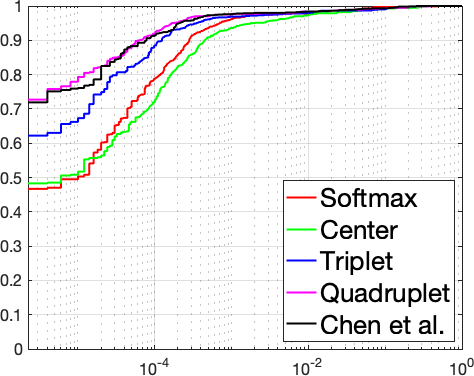}};     
\draw (-2.75, 0+\deltaY) node[rectangle, rotate=90] {\scriptsize{Verification Rate (VR)}};  
\draw (0,-2.2+\deltaY) node[rectangle] {\scriptsize{FAR}};

 \def\deltaX{5.5}
 \draw (0+\deltaX,1.8+\deltaY) node[rectangle] {\scriptsize{\textbf{Megaface}}};  
\draw (0+\deltaX,0+\deltaY) node(n1)  {\includegraphics[width=5 cm]{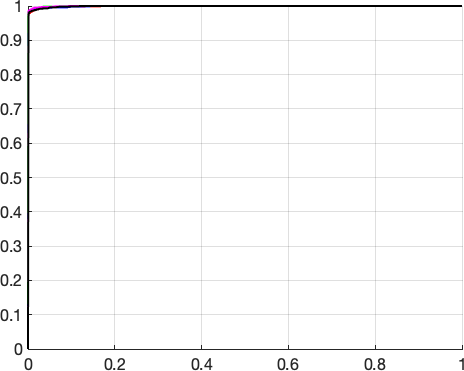}};     
\draw (0.25+\deltaX,0.025+\deltaY) node(n1)  {\includegraphics[width=4.25 cm]{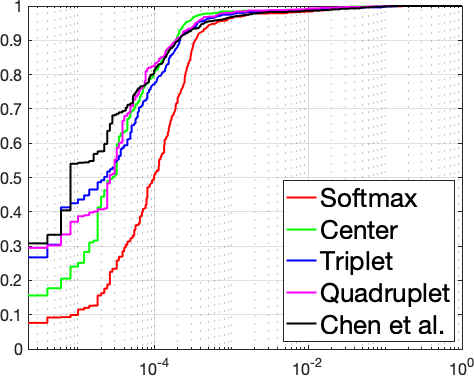}};     
\draw (-2.75+\deltaX, 0+\deltaY) node[rectangle, rotate=90] {\scriptsize{Verification Rate (VR)}};  
\draw (0+\deltaX,-2.2+\deltaY) node[rectangle] {\scriptsize{FAR}};  

\def\deltaX{11.0}
 \draw (0+\deltaX,1.8+\deltaY) node[rectangle] {\scriptsize{\textbf{IJB-A}}};  
\draw (0+\deltaX,0+\deltaY) node(n1)  {\includegraphics[width=5 cm]{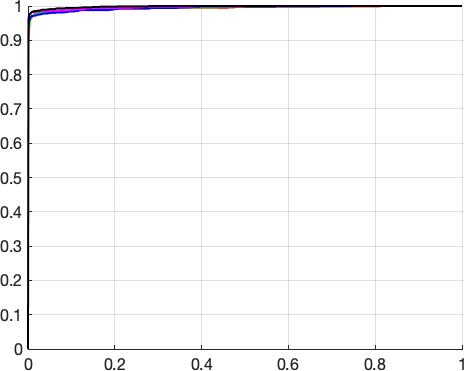}};     
\draw (0.25+\deltaX,0.025+\deltaY) node(n1)  {\includegraphics[width=4.25 cm]{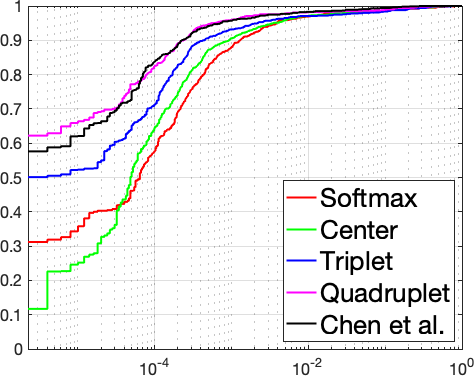}};     
\draw (-2.75+\deltaX, 0+\deltaY) node[rectangle, rotate=90] {\scriptsize{Verification Rate (VR)}};  
\draw (0+\deltaX,-2.2+\deltaY) node[rectangle] {\scriptsize{FAR}};

\end{tikzpicture}
    \caption{Comparison between the Receiver Operating Characteristic (ROC) curves observed for the LFW, Megaface and IJB-A sets, in linear and logarithmic (inner plot) scales, using the VGG and ResNet architectures. Results are shown for the quadruplet loss function (purple lines), and four baselines: \emph{softmax} (red lines), center loss (green lines), triplet loss (blue lines) and Chen \etal~\cite{Chen2017}'s method.}
        \label{fig:ROC}
    \end{center}
\end{figure*}

\begin{figure*}[ht!]
\begin{center}
\begin{tikzpicture}

\def\deltaY{0}

\def\posX{5.5}
\def\posY{2.5}
\fill [rounded corners, gray] (\posX-1, \posY-0.15+\deltaY) rectangle (\posX+1, \posY+0.15+\deltaY);    
\draw  [white] (\posX, \posY+\deltaY) node {\scriptsize{VGG}};    

\draw (0,1.8+\deltaY) node[rectangle] {\scriptsize{\textbf{LFW}}};  
\draw (0,0+\deltaY) node(n1)  {\includegraphics[width=5 cm]{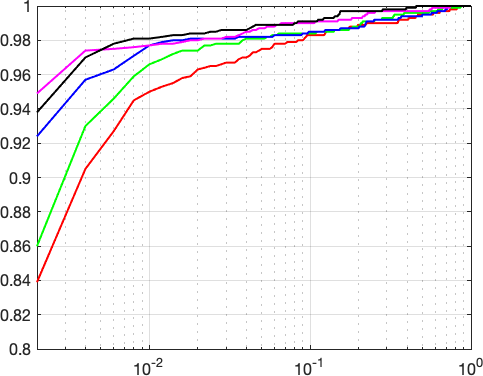}};     
\draw (0.45,-0.15+\deltaY) node(n1)  {\includegraphics[width=3.5 cm]{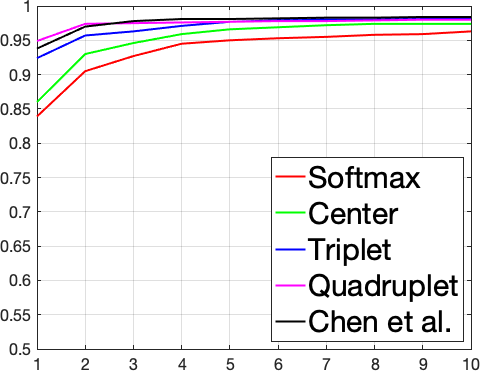}};     
\draw (-2.75, 0+\deltaY) node[rectangle, rotate=90] {\scriptsize{Identification Rate}};  
\draw (0,-2.2+\deltaY) node[rectangle] {\scriptsize{Rank}};

 \def\deltaX{5.5}
  \draw (0+\deltaX,1.8+\deltaY) node[rectangle] {\scriptsize{\textbf{Megaface}}};  
\draw (0+\deltaX,0+\deltaY) node(n1)  {\includegraphics[width=5 cm]{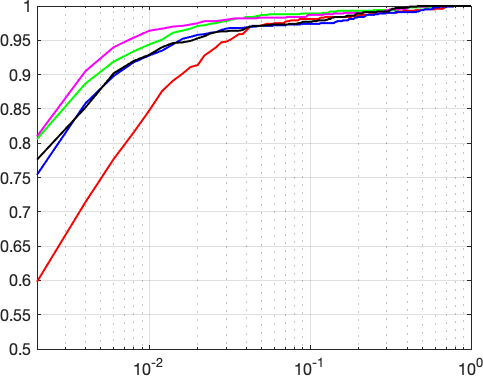}};     
\draw (0.45+\deltaX,-0.15+\deltaY) node(n1)  {\includegraphics[width=3.5 cm]{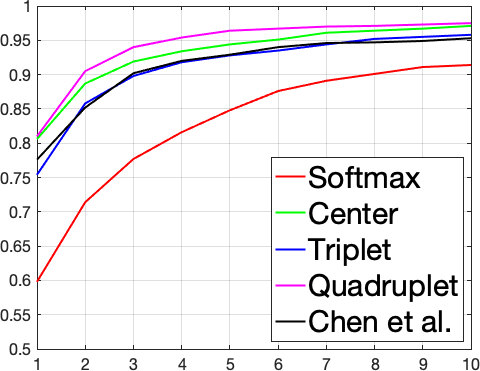}};     
\draw (-2.75+\deltaX, 0+\deltaY) node[rectangle, rotate=90] {\scriptsize{Identification Rate}};  
\draw (0+\deltaX,-2.2+\deltaY) node[rectangle] {\scriptsize{Rank}};  

 \def\deltaX{11.0}
  \draw (0+\deltaX,1.8+\deltaY) node[rectangle] {\scriptsize{\textbf{IJB-A}}};  
\draw (0+\deltaX,0+\deltaY) node(n1)  {\includegraphics[width=5 cm]{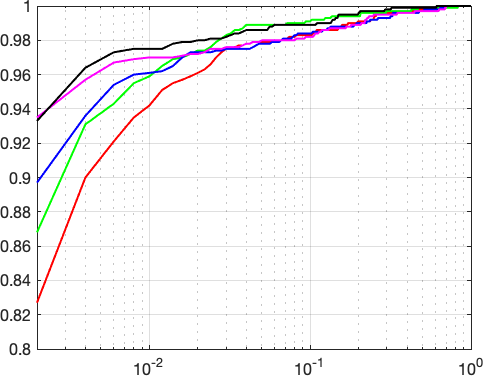}};     
\draw (0.45+\deltaX,-0.15+\deltaY) node(n1)  {\includegraphics[width=3.5 cm]{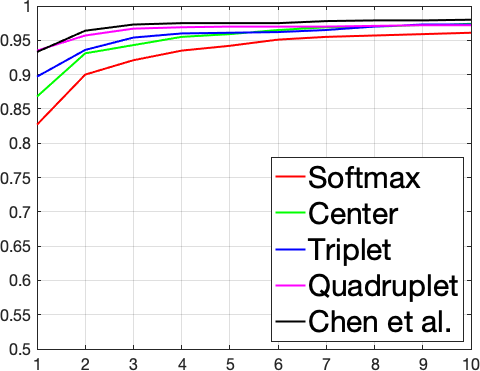}};     
\draw (-2.75+\deltaX, 0+\deltaY) node[rectangle, rotate=90] {\scriptsize{Identification Rate}};  
\draw (0+\deltaX,-2.2+\deltaY) node[rectangle] {\scriptsize{Rank}};

\def\deltaY{-5.25}

\def\posX{5.5}
\def\posY{2.5}
\fill [rounded corners, gray] (\posX-1, \posY-0.15+\deltaY) rectangle (\posX+1, \posY+0.15+\deltaY);    
\draw  [white] (\posX, \posY+\deltaY) node {\scriptsize{ResNet}};    

\draw (0,1.8+\deltaY) node[rectangle] {\scriptsize{\textbf{LFW}}};  
\draw (0,0+\deltaY) node(n1)  {\includegraphics[width=5 cm]{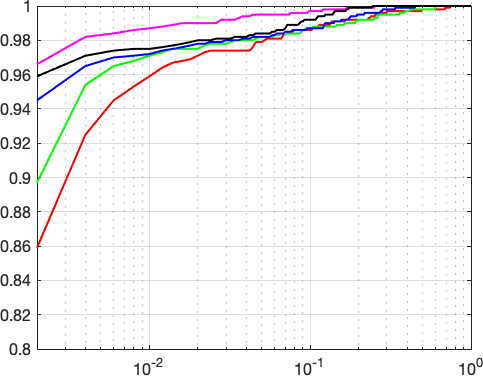}};     
\draw (0.45,-0.15+\deltaY) node(n1)  {\includegraphics[width=3.5 cm]{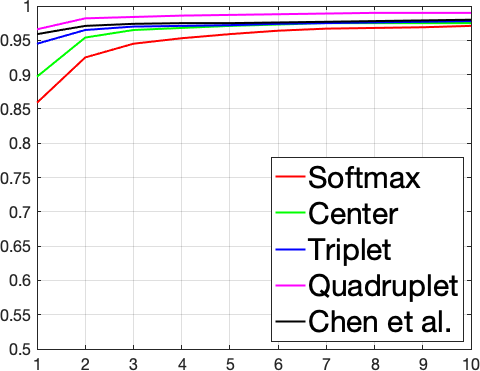}};     
\draw (-2.75, 0+\deltaY) node[rectangle, rotate=90] {\scriptsize{Identification Rate}};  
\draw (0,-2.2+\deltaY) node[rectangle] {\scriptsize{Rank}};

 \def\deltaX{5.5}
  \draw (0+\deltaX,1.8+\deltaY) node[rectangle] {\scriptsize{\textbf{Megaface}}};  
\draw (0+\deltaX,0+\deltaY) node(n1)  {\includegraphics[width=5 cm]{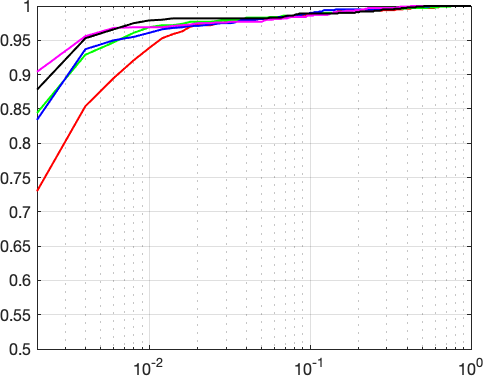}};     
\draw (0.45+\deltaX,-0.15+\deltaY) node(n1)  {\includegraphics[width=3.5 cm]{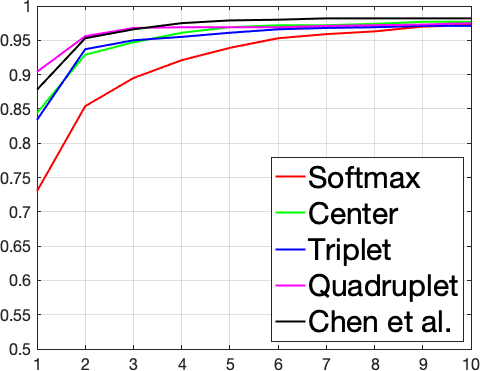}};     
\draw (-2.75+\deltaX, 0+\deltaY) node[rectangle, rotate=90] {\scriptsize{Identification Rate}};  
\draw (0+\deltaX,-2.2+\deltaY) node[rectangle] {\scriptsize{Rank}};  

 \def\deltaX{11.0}
  \draw (0+\deltaX,1.8+\deltaY) node[rectangle] {\scriptsize{\textbf{IJB-A}}};  
\draw (0+\deltaX,0+\deltaY) node(n1)  {\includegraphics[width=5 cm]{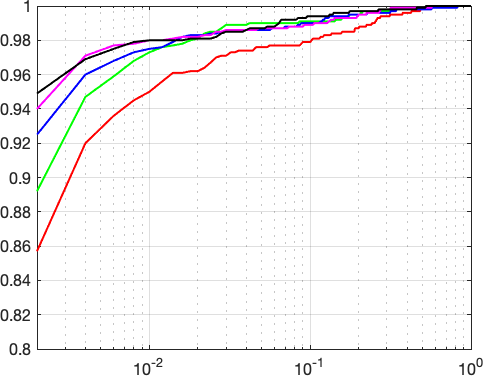}};     
\draw (0.45+\deltaX,-0.15+\deltaY) node(n1)  {\includegraphics[width=3.5 cm]{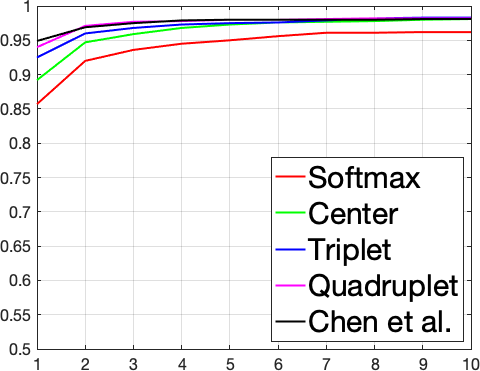}};     
\draw (-2.75+\deltaX, 0+\deltaY) node[rectangle, rotate=90] {\scriptsize{Identification Rate}};  
\draw (0+\deltaX,-2.2+\deltaY) node[rectangle] {\scriptsize{Rank}};

\end{tikzpicture}
    \caption{Closed-set identification (CMC) curves observed for the LFW, Megaface and IJB-A sets, using the VGG and ResNet architectures. A zoomed-in region with the top-10 results is shown in the inner plot. Results are shown for the quadruplet loss function (purple lines), and  four baselines: the \emph{softmax} (red lines), center loss (green lines), triplet loss (blue lines) and Chen \etal~\cite{Chen2017}'s method.}
        \label{fig:CMC}
    \end{center}
\end{figure*}

\begin{figure}[ht!]
\begin{center}
\begin{tikzpicture}[thick,scale=0.75, every node/.style={scale=0.685}]

\def\deltaY{0}

\def\posX{4.0}
\def\posY{2.15}
\fill [rounded corners, gray] (\posX-1, \posY-0.15+\deltaY) rectangle (\posX+1, \posY+0.15+\deltaY);    
\draw  [white] (\posX, \posY+\deltaY) node {\scriptsize{VGG}};    

\draw (0,1.6+\deltaY) node[rectangle] {\scriptsize{\textbf{LFW}}};  
\draw (0,0+\deltaY) node(n1)  {\includegraphics[width=4 cm]{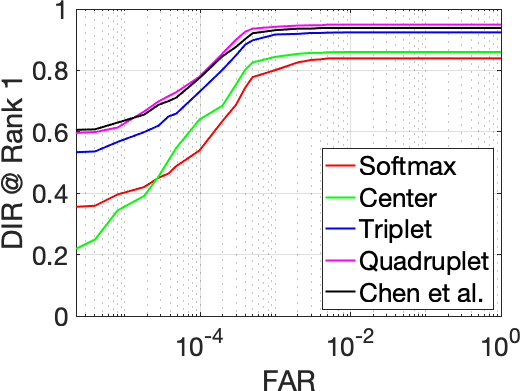}};     
%\draw (-2.25, 0+\deltaY) node[rectangle, rotate=90] {\scriptsize{DIR @ rank-1}};  
%\draw (0,-1.8+\deltaY) node[rectangle] {\scriptsize{FAR}};  

 \def\deltaX{4.0}
  \draw (0+\deltaX,1.6+\deltaY) node[rectangle] {\scriptsize{\textbf{Megaface}}};  
\draw (0+\deltaX,0+\deltaY) node(n1)  {\includegraphics[width=4 cm]{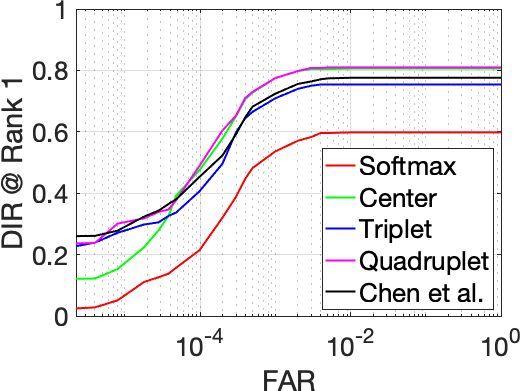}};     
%\draw (-2.25+\deltaX, 0+\deltaY) node[rectangle, rotate=90] {\scriptsize{DIR @ rank-1}};  
%\draw (0+\deltaX,-1.8+\deltaY) node[rectangle] {\scriptsize{FAR}};  

 \def\deltaX{8.0}
  \draw (0+\deltaX,1.6+\deltaY) node[rectangle] {\scriptsize{\textbf{IJB-A}}};  
\draw (0+\deltaX,0+\deltaY) node(n1)  {\includegraphics[width=4 cm]{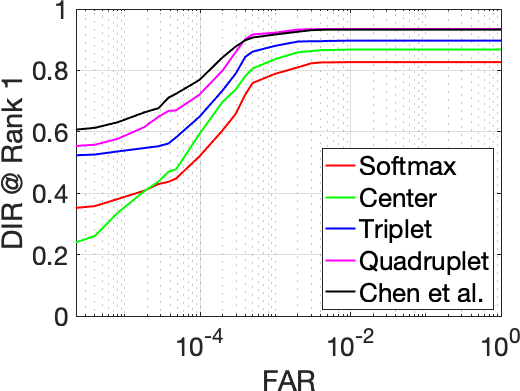}};     
%\draw (-2.25+\deltaX, 0+\deltaY) node[rectangle, rotate=90] {\scriptsize{DIR @ rank-1}};  
%\draw (0+\deltaX,-1.8+\deltaY) node[rectangle] {\scriptsize{FAR}};  

\def\deltaY{-4.25}

\def\posX{4.0}
\def\posY{2.15}
\fill [rounded corners, gray] (\posX-1, \posY-0.15+\deltaY) rectangle (\posX+1, \posY+0.15+\deltaY);    
\draw  [white] (\posX, \posY+\deltaY) node {\scriptsize{ResNet}};    

\draw (0,1.6+\deltaY) node[rectangle] {\scriptsize{\textbf{LFW}}};  
\draw (0,0+\deltaY) node(n1)  {\includegraphics[width=4 cm]{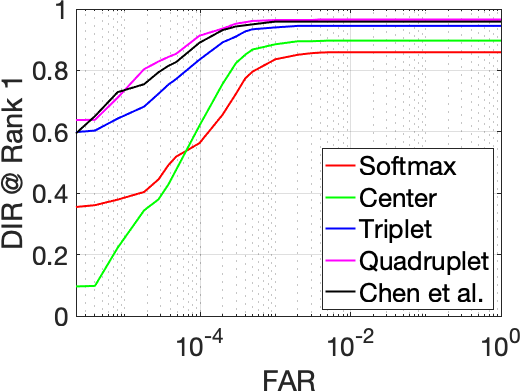}};     
%\draw (-2.25, 0+\deltaY) node[rectangle, rotate=90] {\scriptsize{DIR @ rank-1}};  
%\draw (0,-1.8+\deltaY) node[rectangle] {\scriptsize{FAR}};  

 \def\deltaX{4.0}
  \draw (0+\deltaX,1.6+\deltaY) node[rectangle] {\scriptsize{\textbf{Megaface}}};  
\draw (0+\deltaX,0+\deltaY) node(n1)  {\includegraphics[width=4 cm]{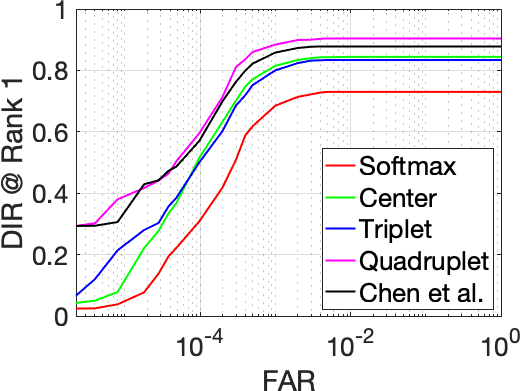}};     
%\draw (-2.25+\deltaX, 0+\deltaY) node[rectangle, rotate=90] {\scriptsize{DIR @ rank-1}};  
%\draw (0+\deltaX,-1.8+\deltaY) node[rectangle] {\scriptsize{FAR}};  

 \def\deltaX{8.0}
  \draw (0+\deltaX,1.6+\deltaY) node[rectangle] {\scriptsize{\textbf{IJB-A}}};  
\draw (0+\deltaX,0+\deltaY) node(n1)  {\includegraphics[width=4 cm]{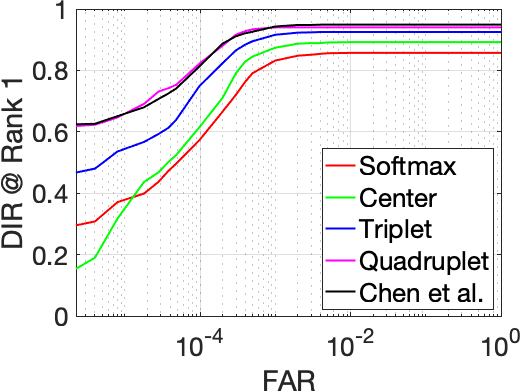}};     
%\draw (-2.25+\deltaX, 0+\deltaY) node[rectangle, rotate=90] {\scriptsize{DIR @ rank-1}};  
%\draw (0+\deltaX,-1.8+\deltaY) node[rectangle] {\scriptsize{FAR}};  

\end{tikzpicture}
    \caption{Comparison between the Detection and Identification Rate (DIR) at rank = 1 curves, for the LFW, Megaface and IJB-A sets, using the VGG-like (top row) and ResNet-like (bottom row) architectures. Results are shown in linear and logarithmic (inner plot) scales,  for the proposed loss (purple lines), and four baselines: \emph{softmax} (red lines), center loss (green lines), triplet loss (blue lines) and Chen \etal~\cite{Chen2017}'s work (black line).}
        \label{fig:DIR}
    \end{center}
\end{figure}
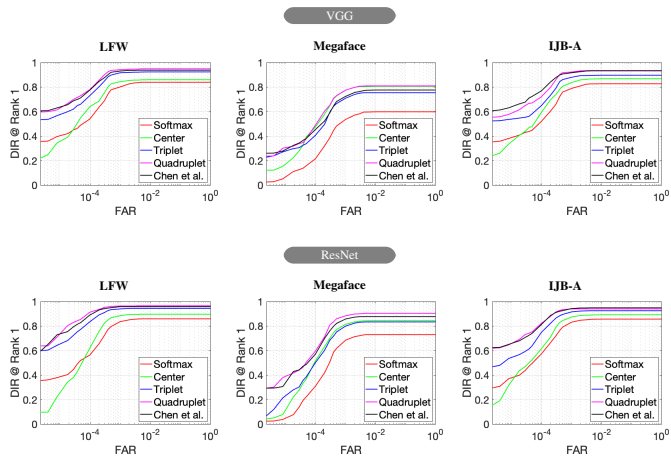

Considering the verification and identification scenarios, we provide the Receiver Operating Characteristic curves (ROC, Fig.~\ref{fig:ROC}), the Cumulative Match curves (CMC, Fig.~\ref{fig:CMC}) and the Detection and Identification rates at rank-1 (DIR, Fig.~\ref{fig:DIR}). The results are summarized in Table~\ref{tab:IdentityResults}, reporting the rank-1, top-10\% values and the mean average precision (mAP) scores:

\begin{align}
\text{mAP} = \frac{\sum_{q=1}^n \bar{P}(q)}{n}, 
 \label{eq:map}
 \end{align}
where $n$ is the number of queries, $\bar{P}(q) = \sum_{k=1}^n P(k) \Delta r(k)$,   $P(k)$ is the precision at cut-off $k$  and $\Delta r(k)$ is the change in recall from $k-1$ to $k$. 

For the LFW set experiment, the BLUFR\footnote{\url{http://www.cbsr.ia.ac.cn/users/scliao/projects/blufr/}} evaluation protocol was chosen. In the verification (1:1) setting, the test set contained 9,708 face images of 4,249 subjects, which yielded over 47 million matching scores. For the open-set identification problem,  the genuine probe set contained 4,350 face images of 1,000 subjects,  the impostor probe set had 4,357 images of 3,249 subjects, and the gallery set had 1,000 images. This evaluation protocol was the basis to design, for the other sets, as close as possible experiments, in terms of the number of matching scores,  gallery and probe sets.

Generally, we observed that the proposed quadruplet loss slightly outperforms the other loss functions, which might also be the result of using additional information for learning. These improvements in performance were observed in most cases by a consistent margin for both the verification and identification tasks, both for the VGG and ResNet architectures. 

In terms of the errors per CNN architecture, the ResNet-like error rates were roughly 0.9 $\times$ (90\%) of the observed for the VGG-like networks (higher margins were observed for the \emph{softmax} loss). Not surprisingly, the Chen \etal~\cite{Chen2017}' method outperformed the remaining competitors, followed by the triplet loss function, which is consistent with most of the results reported in the literature. The \emph{softmax} loss got repeatedly the worst performance among the five functions considered. 

Regarding the levels of performance per dataset, the values observed for the Megaface set were far worse for all objective functions than the corresponding values for the LFW and IJB-A sets. In the former dataset, we followed the protocol of the \emph{small} training set, using 490,000 images from 17,189 subjects for learning (images overlapping with Facescrub dataset were discarded).  Also, it is important to note that the relative performance between the loss functions was roughly the same in all sets. Degradations in performance were slight from the LFW to the IJB-A set and much more visible in case of the Megaface set. In this context, the \emph{softmax} loss produced the most evident degradations, followed by the center loss.
    
\begin{table}
\caption{Identity retrieval performance of the proposed loss with respect to the baselines: \emph{softmax}, center and triplet losses, and also Chen \etal~\cite{Chen2017}'s method. The average performance $\pm$ standard deviation values are given, after 10 trials. Inside each cell, values regard (from top to bottom) the LFW, Megaface and IJB-A datasets. The bold font highlights the best results per dataset among all methods tested.}
\label{tab:IdentityResults}
\begin{center}
\begin{tabular}{|l|c|c|c|}
\hline
\textbf{\scriptsize{Method}} & \textbf{\scriptsize{mAP}} & \textbf{\scriptsize{rank-1}} &  \textbf{\scriptsize{top-10\%}} \\
\hline\hline

%%%%%%%%%%%%%%%%%%%%%%%%%%%%%%%%%%%%%%%%%%%%%%%%%%%%%%%%%%%%%%%%%%%%%%%%%%%
\multicolumn{4}{|c|}{ \cellcolor{gray!25}  \small{VGG}} \\
\multirow{3}{*}{\scriptsize{Quadruplet loss}} & \scriptsize{0.958} \scriptsize{$\pm$ 3$e^{-3}$} &    \scriptsize{\textbf{0.951}} \scriptsize{$\pm$ 0.020} &   \scriptsize{\textbf{0.979}} \scriptsize{$\pm$ 6$e^{-3}$}\\ 
& \scriptsize{\textbf{0.877}} \scriptsize{$\pm$ 0.011} & \scriptsize{\textbf{0.812}} \scriptsize{$\pm$ 0.053}  & \scriptsize{\textbf{0.960}} \scriptsize{$\pm$ 9$e^{-3}$}\\
&  \scriptsize{\textbf{0.953}} \scriptsize{$\pm$ 5$e^{-3}$} &   \scriptsize{\textbf{0.939}} \scriptsize{$\pm$ 0.037} & \scriptsize{0.958} \scriptsize{$\pm$ 6$e^{-3}$} \\ \hline
\multirow{3}{*}{\scriptsize{Softmax loss}} & \scriptsize{0.897} \scriptsize{$\pm$ 4$e^{-3}$} & \scriptsize{0.842} \scriptsize{$\pm$ 0.034}  & \scriptsize{0.953} \scriptsize{$\pm$ 0.011}\\
& \scriptsize{0.727} \scriptsize{$\pm$ 0.014} & \scriptsize{0.615} \scriptsize{$\pm$ 0.060}  & \scriptsize{0.863} \scriptsize{$\pm$ 0.017}\\
& \scriptsize{0.849} \scriptsize{$\pm$ 0.010}  & \scriptsize{0.823} \scriptsize{$\pm$ 0.039}   & \scriptsize{0.941} \scriptsize{$\pm$ 0.014} \\ \hline

\multirow{3}{*}{\scriptsize{Triplet loss~\cite{Schroff2015}}} & \scriptsize{0.934} \scriptsize{$\pm$ 4$e^{-3}$} & \scriptsize{0.929} \scriptsize{$\pm$ 0.033}  & \scriptsize{0.964} \scriptsize{$\pm$ 8$e^{-3}$}\\
& \scriptsize{0.854} \scriptsize{$\pm$ 9$e^{-3}$} & \scriptsize{0.758} \scriptsize{$\pm$ 0.059}  & \scriptsize{0.946} \scriptsize{$\pm$ 0.017}\\
& \scriptsize{0.917} \scriptsize{$\pm$ 5$e^{-3}$} &   \scriptsize{0.901} \scriptsize{$\pm$ 0.040} &  \scriptsize{0.950} \scriptsize{$\pm$ 0.011}\\ \hline

\multirow{3}{*}{\scriptsize{Center loss~\cite{Wen2019}}} & \scriptsize{0.918} \scriptsize{$\pm$ 3$e^{-3}$} & \scriptsize{0.863} \scriptsize{$\pm$ 0.020}  & \scriptsize{0.962} \scriptsize{$\pm$ 6$e^{-3} $}\\
& \scriptsize{0.850} \scriptsize{$\pm$ 0.013} & \scriptsize{0.773} \scriptsize{$\pm$ 0.052}  & \scriptsize{0.939} \scriptsize{$\pm$ 0.012}\\
&  \scriptsize{0.862} \scriptsize{$\pm$ 0.010} &  \scriptsize{0.867} \scriptsize{$\pm$ 0.041}  & \scriptsize{0.944} \scriptsize{$\pm$ 0.012} \\ \hline

\multirow{3}{*}{\scriptsize{Chen \etal~\cite{Chen2017}}} & \scriptsize{\textbf{0.961}} \scriptsize{$\pm$ 2$e^{-3}$} & \scriptsize{0.945} \scriptsize{$\pm$ 0.022}  & \scriptsize{0.976} \scriptsize{$\pm$ 6$e^{-3} $}\\
& \scriptsize{0.864} \scriptsize{$\pm$ 0.012} & \scriptsize{0.772} \scriptsize{$\pm$ 0.061}  & \scriptsize{0.947} \scriptsize{$\pm$ 9$e^{-3} $}\\
&  \scriptsize{0.948} \scriptsize{$\pm$ 6$e^{-3}$} &  \scriptsize{0.936} \scriptsize{$\pm$ 0.055}  &  \scriptsize{\textbf{0.970}} \scriptsize{$\pm$ 4$e^{-3}$}\\ \hline

%%%%%%%%%%%%%%%%%%%%%%%%%%%%%%%%%%%%%%%%%%%%%%%%%%%%%%%%%%%%%%%%%%%%%%%%%%%
\multicolumn{4}{|c|}{ \cellcolor{gray!25}  \small{ResNet}} \\

\multirow{3}{*}{\scriptsize{Quadruplet loss}} &  \scriptsize{\textbf{0.968}} \scriptsize{$\pm$ 2$e^{-3}$} &  \scriptsize{\textbf{0.966}} \scriptsize{$\pm$ 0.012}  & \scriptsize{0.981} \scriptsize{$\pm$ 4$e^{-3}$} \\ 
&  \scriptsize{0.902} \scriptsize{$\pm$ 9$e^{-3}$} &  \scriptsize{\textbf{0.906}} \scriptsize{$\pm$ 0.048}  & \scriptsize{\textbf{0.972}} \scriptsize{$\pm$ 8$e^{-3}$} \\ 
&  \scriptsize{\textbf{0.959}} \scriptsize{$\pm$ 3$e^{-3}$} &  \scriptsize{0.947} \scriptsize{$\pm$ 0.021}  & \scriptsize{0.980} \scriptsize{$\pm$ 4$e^{-3}$} \\ \hline

\multirow{3}{*}{\scriptsize{Softmax loss}} &  \scriptsize{0.912} \scriptsize{$\pm$ 4$e^{-3}$} &  \scriptsize{0.861} \scriptsize{$\pm$ 0.029}  & \scriptsize{0.960} \scriptsize{$\pm$ 8$e^{-3}$} \\ 
&  \scriptsize{0.730} \scriptsize{$\pm$ 0.010} &  \scriptsize{0.745} \scriptsize{$\pm$ 0.051}  & \scriptsize{0.899} \scriptsize{$\pm$ 0.011} \\
&  \scriptsize{0.841} \scriptsize{$\pm$ 9$e^{-3}$}& \scriptsize{0.860} \scriptsize{$\pm$ 0.030}   & \scriptsize{0.958} \scriptsize{$\pm$ 8$e^{-3}$}\\ \hline

\multirow{3}{*}{\scriptsize{Triplet loss~\cite{Schroff2015}}} &  \scriptsize{0.947} \scriptsize{$\pm$ 4$e^{-3}$} &  \scriptsize{0.948} \scriptsize{$\pm$ 0.026}  & \scriptsize{0.968} \scriptsize{$\pm$ 9$e^{-3}$} \\
&  \scriptsize{0.872} \scriptsize{$\pm$ 8$e^{-3}$} &  \scriptsize{0.839} \scriptsize{$\pm$ 0.052}  & \scriptsize{0.957} \scriptsize{$\pm$ 9$e^{-3}$} \\
&  \scriptsize{0.919} \scriptsize{$\pm$ 5$e^{-3}$}&  \scriptsize{0.937} \scriptsize{$\pm$ 0.031}  & \scriptsize{0.961} \scriptsize{$\pm$ 0.011}\\ \hline

\multirow{3}{*}{\scriptsize{Center loss~\cite{Wen2019}}} &  \scriptsize{0.939} \scriptsize{$\pm$ 3$e^{-3}$} &  \scriptsize{0.898} \scriptsize{$\pm$ 0.016}  & \scriptsize{0.967} \scriptsize{$\pm$ 6$e^{-3}$} \\
&  \scriptsize{0.847} \scriptsize{$\pm$ 9$e^{-3}$} &  \scriptsize{0.845} \scriptsize{$\pm$ 0.048}  & \scriptsize{0.945} \scriptsize{$\pm$ 9$e^{-3}$} \\ 
&  \scriptsize{0.877} \scriptsize{$\pm$ 7$e^{-3}$} &   \scriptsize{0.893} \scriptsize{$\pm$ 0.035} &  \scriptsize{0.963} \scriptsize{$\pm$ 9$e^{-3}$} \\ \hline

\multirow{3}{*}{\scriptsize{Chen \etal~\cite{Chen2017}}} &  \scriptsize{0.966} \scriptsize{$\pm$ 2$e^{-3}$} &  \scriptsize{0.959} \scriptsize{$\pm$ 0.015}  & \scriptsize{\textbf{0.983}} \scriptsize{$\pm$ 4$e^{-3}$} \\
&  \scriptsize{\textbf{0.916}} \scriptsize{$\pm$ 8$e^{-2}$} &  \scriptsize{0.880} \scriptsize{$\pm$ 0.050}  & \scriptsize{0.975} \scriptsize{$\pm$ 8$e^{-3}$} \\
&  \scriptsize{0.952} \scriptsize{$\pm$ 4$e^{-3}$} &  \scriptsize{\textbf{0.960}} \scriptsize{$\pm$ 0.022}  &  \scriptsize{\textbf{0.986}} \scriptsize{$\pm$ 6$e^{-3}$}\\

\hline
\end{tabular}
\end{center}

\end{table}

\subsection{Soft Biometrics Inference}

As stated above, the proposed loss can also be used for learning a soft biometrics estimator. Then, in test time the position to where one element is projected to can be used to infer the soft labels, in a simple nearest neighbour fashion. In these experiments, we considered only 1-NN, i.e., the label inferred for each query was given by the closest gallery element. Better results would be possibly attained if a larger number of neighbours had been considered, even though the computational cost of classification will also increase. All experiments were conducted according to a bootstrapping-like strategy: having $n$ test images available, the bootstrap randomly selected (with replacement) $0.9 \times n$ images, obtaining samples composed of 90\% of  the whole data. Ten test samples were created and the experiments were conducted independently on each trail, which enabled to obtain the mean and the standard deviation at each performance value.
   
As baselines, we used two off-the-shelf techniques that represent the state-of-the-art: the Matlab SDK  for \emph{Face++}\footnote{\url{http://www.faceplusplus.com/}} and the \emph{Microsoft Cognitive Toolkit Commercial}\footnote{\url{https://www.microsoft.com/cognitive-services/}}. Face++ is a commercial face recognition system, with good performance  reported for the LFW face recognition competition (second best rate). Microsoft Cognitive Toolkit is a deep learning framework that provides useful information based on vision, speech and language.

We considered exclusively the ''Gender'', ''Ethnicity'' and ''Age'' labels ($t=3$), quantised respectively into two classes for Gender (''\emph{male}'', ''\emph{female}''), three classes for Age (''\emph{young}'', ''\emph{adult}'', ''\emph{senior}''), and three classes for Ethnicity (''\emph{white}'', ''\emph{black}'', ''\emph{asian}''). The average and standard deviation performance values are reported in Table~\ref{tab:SoftBiometricsResults} for the BIODI, PETA and LFW sets. 

The results achieved by the quadruplet loss can be favourably compared  to the COTS techniques for most labels, particularly for the BIODI and LFW datasets. Regarding the PETA set, Face++ invariably outperformed the other techniques, even if at a reduced margin in most cases. This was justified by the extreme heterogeneity of image features in this set, in result of being the concatenation of different databases. This probably had reduced the representativity of the learning data with respect the test set, being the Face++ model apparently the least sensitive to this covariate. Note that the ''Ethnicity'' label is only provided by the Face++ framework.

Globally, these experiments supported the possibility of using such the proposed method to estimate soft labels in a \emph{single-shot} paradigm, which can be particularly interesting to reduce the computational cost of using specialized soft labelling tools.

\begin{table}
\caption{Soft biometrics labelling performance (mAP) attained by the proposed method, with respect  to two commercial-off-the-shelf systems (Face++ and Microsoft Cognitive). The average performance $\pm$ standard deviation values are given, after 10 trials. Inside each cell, the top values regard the VGG-like model, and the bottom values regard the ResNet-like architecture}
\label{tab:SoftBiometricsResults}
\begin{center}
\begin{tabular}{|l|c|c|c|}
\hline
\textbf{\scriptsize{Method}} & \textbf{\scriptsize{Gender}} & \textbf{\scriptsize{Age}} &  \textbf{\scriptsize{Ethnicity}} \\
\hline\hline
\multicolumn{4}{|c|}{ \cellcolor{gray!25}  \small{BIODI}} \\
\multirow{2}{*}{\scriptsize{Quadruplet loss}} & \scriptsize{0.816} \scriptsize{$\pm$ 6$e^{-3}$} & \scriptsize{0.603} \scriptsize{$\pm$ 0.014}  & \scriptsize{0.777} \scriptsize{$\pm$ 0.011}\\
&  \scriptsize{\textbf{0.834}} \scriptsize{$\pm$ 5$e^{-3}$} &  \scriptsize{\textbf{0.649}} \scriptsize{$\pm$ 0.011}  & \scriptsize{0.786} \scriptsize{$\pm$ 9$e^{-3}$} \\ \hline

\scriptsize{Face++} & \scriptsize{0.760} \scriptsize{$\pm$ 8$e^{-3}$} & \scriptsize{0.588} \scriptsize{$\pm$ 0.019}  & \scriptsize{\textbf{0.788}} \scriptsize{$\pm$ 0.017}\\ \hline
\scriptsize{Microsoft Cognitive} & \scriptsize{0.738} \scriptsize{$\pm$ 7$e^{-3}$} & \scriptsize{0.552} \scriptsize{$\pm$ 0.026}  & \scriptsize{-}\\\hline

\multicolumn{4}{|c|}{ \cellcolor{gray!25}  \small{PETA}} \\
\multirow{2}{*}{\scriptsize{Quadruplet loss}} & \scriptsize{0.862} \scriptsize{$\pm$ 0.024} & \scriptsize{0.649} \scriptsize{$\pm$ 0.061}  & \scriptsize{0.797} \scriptsize{$\pm$ 0.053}\\
&  \scriptsize{0.882} \scriptsize{$\pm$ 0.018} &  \scriptsize{0.658} \scriptsize{$\pm$ 0.057}  & \scriptsize{0.810} \scriptsize{$\pm$ 0.036} \\ \hline

\scriptsize{Face++} & \scriptsize{0.870} \scriptsize{$\pm$ 0.028} & \scriptsize{0.653} \scriptsize{$\pm$ 0.062}  & \scriptsize{\textbf{0.812}} \scriptsize{$\pm$ 0.054}\\  \hline

\scriptsize{Microsoft Cognitive} & \scriptsize{\textbf{0.885}} \scriptsize{$\pm$ 0.020} & \scriptsize{\textbf{0.660}} \scriptsize{$\pm$ 0.057}  & \scriptsize{-}\\ \hline

\multicolumn{4}{|c|}{ \cellcolor{gray!25}  \small{LFW}} \\
\multirow{2}{*}{\scriptsize{Quadruplet loss}} & \scriptsize{0.939} \scriptsize{$\pm$ 0.021} & \scriptsize{0.702} \scriptsize{$\pm$ 0.059}  & \scriptsize{0.801} \scriptsize{$\pm$ 0.044}\\
&  \scriptsize{\textbf{0.944}} \scriptsize{$\pm$ 0.017} &  \scriptsize{0.709} \scriptsize{$\pm$ 0.049}  & \scriptsize{0.817} \scriptsize{$\pm$ 0.041} \\ \hline

\scriptsize{Face++} & \scriptsize{0.928} \scriptsize{$\pm$ 0.041} & \scriptsize{0.527} \scriptsize{$\pm$ 0.063}  & \scriptsize{\textbf{0.842}} \scriptsize{$\pm$ 0.061}\\\hline

\scriptsize{Microsoft Cognitive} & \scriptsize{0.931} \scriptsize{$\pm$ 0.037} & \scriptsize{\textbf{0.710}} \scriptsize{$\pm$ 0.051}  & \scriptsize{-}\\ \hline

\hline
\end{tabular}
\end{center}

\end{table}

\begin{figure}[ht!]
\begin{center}
\begin{tikzpicture}

\draw (0,0) node(n1)  {\includegraphics[width=4.0 cm]{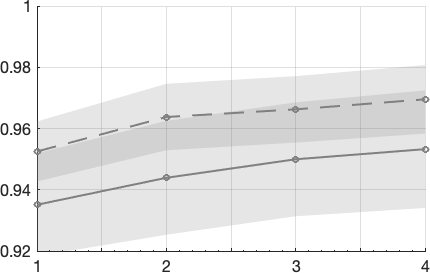}};     
\draw (-2.15, 0) node[rectangle, rotate=90] {\small{rank-1}};  
\draw (-0, -1.5) node[rectangle] {\small{$t$}}; 

%\draw [black] (1.75,1.9) rectangle (3.75,0.95);    
%\draw (2.15, 1.68) node(n1)  {\includegraphics[width=0.5 cm]{imgs/label_1}};     
%\draw (2.9, 1.68) node[rectangle] {\scriptsize{VGG}}; 
%
%\draw (2.15, 1.28) node(n1)  {\includegraphics[width=0.5 cm]{imgs/label_2}};     
%\draw (3.01, 1.28) node[rectangle] {\scriptsize{ResNet}}; 

\draw (4.4,0) node(n1)  {\includegraphics[width=4.0 cm]{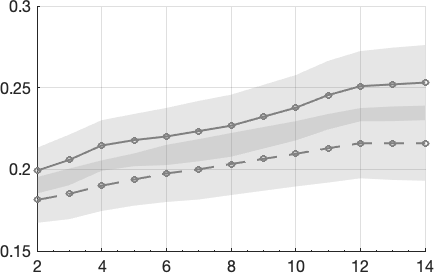}};     
\draw (4.4-2.15, 0) node[rectangle, rotate=90] {\small{e()}};  
\draw (4.4, -1.5) node[rectangle] {\small{$t$}}; 

\end{tikzpicture}
    \caption{Soft biometrics labelling and rank-1 effectiveness with respect to the value of $t$. At left: rank-1 identification accuracy in the LFW dataset, for $1 \leq t \leq 4$. At right: soft biometrics performance in the BIODI test dataset, for $2 \leq t \leq 14$. The solid lines represent the VGG CNN, and the dashed lines regard the ResNet architecture.}
        \label{fig:variations_t}
    \end{center}
\end{figure}
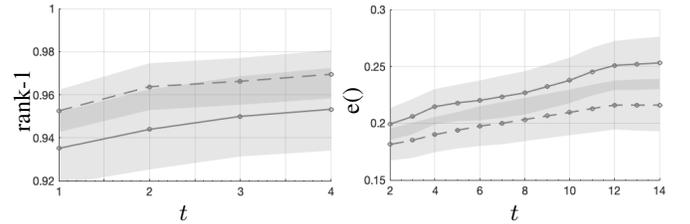

Finally, we analysed the variations in performance with respect to the number of labels considered, i.e., the value of the $t$ parameter. At first, to perceive how the identity retrieval performance depends of the number of soft labels, we used the annotations provided by the ATVS group~\cite{Sosa2018} for the LFW set, and measured the rank-1 variations for $1 \leq t \leq 4$, starting by the ''ID'' label alone and then adding iteratively the ''Gender'' $\rightarrow$ ''Ethnicity'' $\rightarrow$ ''Age'' labels. The results are shown in the left part of Fig.~\ref{fig:variations_t}. In a complementar way, to perceive the overall labelling effectiveness for large values of $t$, the BIODI dataset was used (the one with the largest number of annotated labels), and the values obtained for $t \in \{1, 2,\ldots, 14\}$. In all cases, $d=128$ was kept, with the average labelling error in the test set $\bm{X}$ given by: 

\begin{align}
e(\bm{X})=\frac{1}{n.t} \sum_{i=1}^n || \bm{p}_{i}- \bm{g}_{i}||_0, 
 \label{eq:variation_t}
 \end{align}
with $\bm{p}_{i}$ denoting the  $t$ labels predicted for the $i^{th}$ image and $\bm{g}_{i}$ being the ground-truth. $||~||_0$ denotes the $\ell_0$-norm. 

It is interesting to observe the apparently contradictory results in both plots: at first, a positive correlation between the labelling errors and the values of $t$ is evident, which was justified by the difficulty of inferring some of the hardest labels in the BIODI set  (e.g., the \emph{type of shoes}). However, the average rank-1 identification accuracy also increased when more soft labels were used, even if the results were obtained only for small values of $t$ (i.e., not considering the particularly hard labels, in result of no available ground truth). Overall, we concluded that the proposed loss obtain \emph{acceptable} performance (i.e., close to the state-of-the-art) when a small number of soft labels is available ($\geq$ 2), but also when a few more labels should be inferred (up to $t \approx 8$). In this regard, we presume that even larger values for $t$ ($t \gg 8$) would require substantially more amounts of learning data and also larger values for $d$ (dimension of the embedding).

 \subsection{Semantic Identity Retrieval}
 \label{sec:SemanticSearch}
 
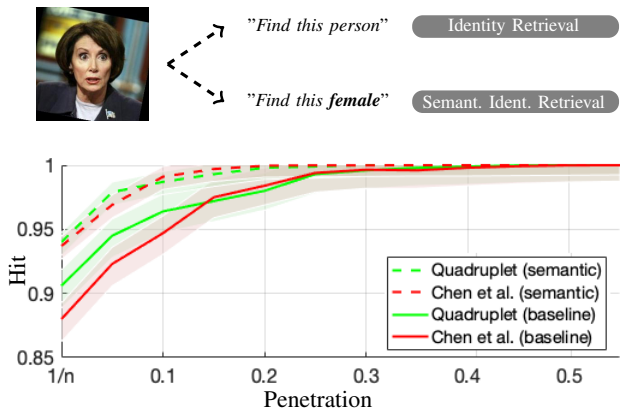
\begin{figure}[ht!]
\begin{center}
\begin{tikzpicture}

\draw (-3,2.75) node(n1)  {\includegraphics[width=1.5 cm]{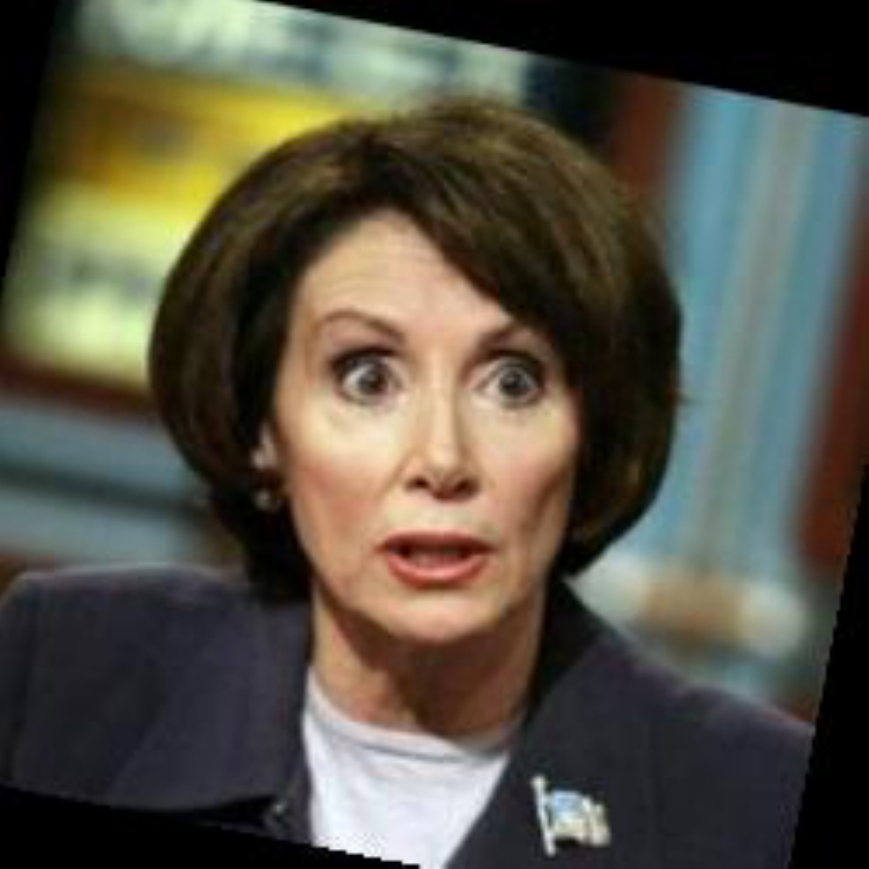}};     

\draw (-0, 3.25) node[rectangle] {\scriptsize{"\emph{Find this person}"}}; 
\draw (-0, 2.25) node[rectangle] {\scriptsize{"\emph{Find this \textbf{female}}"}}; 

\draw [dashed, very thick,->] (-2, 2.75) -- (-1.25, 3.25); 
\draw [dashed, very thick,->] (-2, 2.75) -- (-1.25, 2.25);

\fill [rounded corners, gray] (1.25, 3.1) rectangle (4, 3.4);    
\draw  [white] (2.6, 3.25) node {\scriptsize{Identity Retrieval}};    

\fill [rounded corners, gray] (1.25, 2.1) rectangle (4, 2.4);    
\draw  [white] (2.6, 2.25) node {\scriptsize{Semant. Ident. Retrieval}};    

\draw (0,0) node(n1)  {\includegraphics[width=8.0 cm]{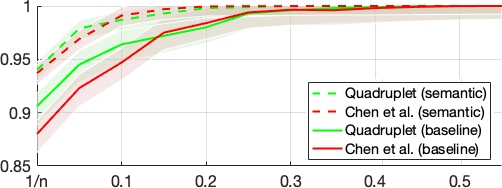}};     
\draw (-4.0, 0) node[rectangle, rotate=90] {\small{Hit}};  
\draw (-0, -1.7) node[rectangle] {\small{Penetration}};

%\draw (4.4,0) node(n1)  {\includegraphics[width=4.0 cm]{imgs/errors_t}};     
%\draw (4.4-2.15, 0) node[rectangle, rotate=90] {\small{e()}};  
%\draw (4.4, -1.5) node[rectangle] {\small{$t$}}; 

\end{tikzpicture}
    \caption{Comparison between the hit/penetration rates attained by the proposed loss and Chen \etal~\cite{Chen2017}'s work, when not considering (baseline) or considering semantic additional information. Results are given for the \emph{ResNet} architecture and Megaface dataset. The ''\emph{Gender}'' was used as the semantic criterium in each query and "n" is the number of enrolled identities.}
        \label{fig:semantic}
    \end{center}
\end{figure}

Finally, we considered the problem of \emph{semantic identity retrieval}, in which, along with the query image, it is also specified some semantic criteria that filter out the retrieval elements (i.e., ''\emph{Find this person}'' $\rightarrow$  ''\emph{Find this female}'', as illustrated in the top part of Fig.~\ref{fig:semantic}). In this setting, it is assumed that the ground truth labels of the gallery identities is known, even though the same guarantee does not exist for query elements. We used the dataset that produced the poorest identity retrieval performance (Megaface) and compared our performance to Chen \etal's method (the most frequent runner up in previous experiments). The soft label ''\emph{Gender}'' (automatically provided by the Microsoft Cognitive Toolkit for all the queries) was used as additional semantic data, to filter out the retrieved identities that do not match the desired gender. The bottom plot in Fig.~\ref{fig:semantic} provides the results in terms of the hit/penetration rates, being particularly evident that both methods attained practically the same performance (''semantic'' data series). Essentially, this means that if coarse labels are available, our method and Chen \etal's attain embeddings of similar quality in terms of identity compactness and discriminability. However, it is noteworthy to observe that the proposed loss  - the baseline version, without using auxiliary semantic information in the query - is a way to approximate the results that are attained by the state-of-the-art when using semantic information to filter out the retrieved identities. This accords the insight that have led to its proposal, as it can be seen as implicitly inferring in a joint way the coarse (soft biometrics) and fine (identity) labels.
 
 \section{Conclusions and Further Work}
 \label{sec:Conclusions}
 
In this paper we proposed a loss function designed to work in multi-output classification problems, where the response variables have dimension larger than one.  Our function can be seen as a generalization of the well known triplet loss, that replaces the \emph{positive}/\emph{negative} division of pairs and the notion of \emph{anchor}, by:  i) a metric that measures the relative similarity between any two classes; and ii) a quadruplet term that - based in the semantic similarity -  imposes the margins between pairs of projections. 

In particular, having focused in the identity retrieval and soft biometrics problems, the proposed loss uses the ''ID'' and available soft labels (e.g., ''Gender'', ''Age'' and ''Ethnicity'') to produce feature embeddings that are semantically coherent, i.e., in which not only the intra-class compactness is guaranteed, but also the broad families of classes appear in adjacent regions. This enables a direct correspondence between the ID centroids and their semantic descriptions, i.e., where a "\emph{young, black, male}'' is surely closer to an  "\emph{adult, black, male}'' than to a ''\emph{young, white, female}''. This property is in opposition to the way previous loss functions work, where elements are projected based uniquely in their appearance, being assumed that the semantical coherence naturally yields from the similarities in appearance between classes  (i.e., upon the similarity of image features).

\section*{Acknowledgements}

This work is funded by FCT/MCTES through national funds and when applicable co-funded by EU funds under the project UIDB/EEA/50008/2020. Also, this research was funded  by  the  FEDER,  Fundo  de  Coes\~ao  e  Fundo Social  Europeu, under  the  PT2020 program, in the scope of the POCI-01-0247-FEDER-033395 project. 
We acknowledge the support of NVIDIA Corporation$^{\scriptsize{\textregistered}}$, with the donation of one \emph{Titan X} GPU board.

{\small
\bibliographystyle{ieee}

}

\end{document}